%% file: main.tex
\pdfoutput=1

\documentclass[10pt,twocolumn,letterpaper]{article}

\usepackage[pagenumbers]{cvpr} %

\input{preamble}

\definecolor{cvprblue}{rgb}{0.21,0.49,0.74}
\usepackage[pagebackref,breaklinks,colorlinks,allcolors=cvprblue]{hyperref}

\def\confName{CVPR}
\def\confYear{2027}

\input{notation}

\title{AvatarAnyone: Agentic Reconstruction of Mesh-Based Layered Avatars from YouTube Video}

\author{Aymen Mir$^{1}$ \qquad Peter Wonka$^{2}$ \qquad Stefanos Zafeiriou$^{1}$ \qquad R{\i}za Alp G\"{u}ler$^{1}$\\
{\normalsize $^{1}$Imperial College London \qquad $^{2}$KAUST}}

\makeatletter
\let\@oldmaketitle\@maketitle
\renewcommand{\@maketitle}{
	\@oldmaketitle
	\begin{center}
		\vspace*{-13.95pt}%
		\includegraphics[width=\linewidth]{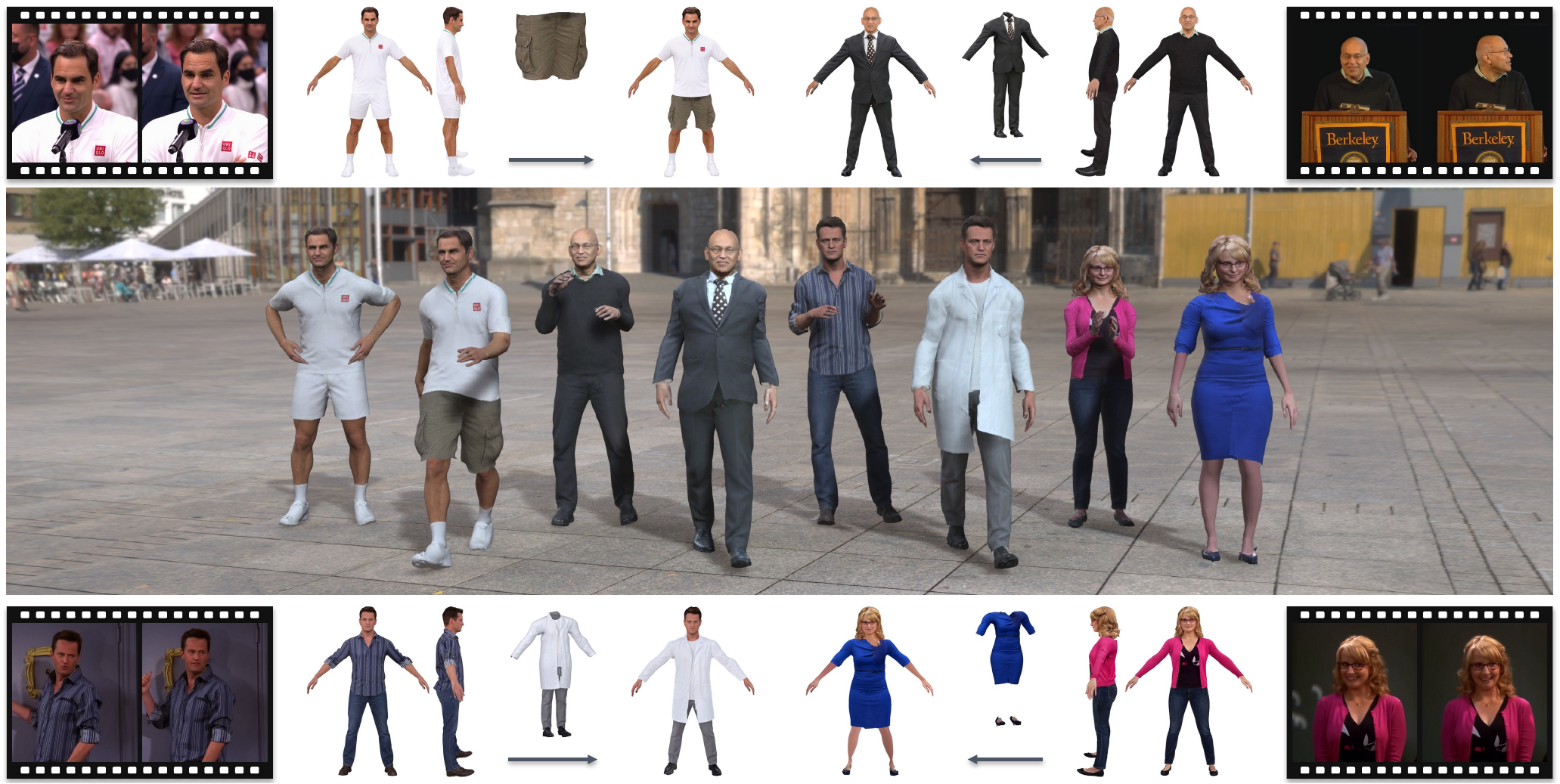}
	\end{center}
	\vspace{-4mm}
	\captionsetup{hypcap=false}
	\captionof{figure}{\textbf{Layered avatars from YouTube videos.} From a YouTube video of a person (film strips), AvatarAnyone reconstructs a layered, textured mesh avatar that can be animated. Each garment is a separate layer, so garments can be replaced. \textbf{Top and bottom:} the avatar from the video, new clothes from another avatar, and the avatar in these clothes. \textbf{Middle:} the four people in their own and in new clothes, together in an Unreal Engine scene.}
	\label{fig:teaser}
	\vspace{1mm}
}
\makeatother

\begin{document}
\maketitle
\input{sec/0_abstract}

\input{sec/1_intro}

\input{sec/2_related}

\input{sec/3_method}

\input{sec/4_exp}
\input{sec/5_conc}

{
    \small
    \bibliographystyle{ieeenat_fullname}
    \bibliography{main}
}

\end{document}

%% file: preamble.tex
\ifdefined\pdfinclusioncopyfonts \pdfinclusioncopyfonts=1 \fi

\usepackage{multirow}
\usepackage{colortbl}
\usepackage{overpic}
\usepackage{comment}
\definecolor{best}{RGB}{220,250,220}
\definecolor{second}{RGB}{240,245,220}
\definecolor{headerbg}{RGB}{200,220,240}
\newcommand{\fst}[1]{\cellcolor{best}\textbf{#1}}
\newcommand{\snd}[1]{\cellcolor{second}#1}

\newif\iftablerow
\tablerowtrue
\newenvironment{rowtable}[1]%
	{\iftablerow\minipage[t]{#1}\else\table[t]\fi}
	{\iftablerow\endminipage\else\vspace{-3mm}\endtable\fi}
\newcommand{\rowtablefont}{\iftablerow\fontsize{6}{7.2}\selectfont\setlength{\tabcolsep}{2pt}\else\fontsize{7}{8.4}\selectfont\fi}

%% file: notation.tex
\input{shared_notation}

%% file: shared_notation.tex
\usepackage{bm}

\renewcommand{\vec}[1]{\mathbf{#1}}

\newcommand{\mat}[1]{\mathbf{#1}}

\newcommand{\set}[1]{\mathcal{#1}}

\newcommand{\real}{\mathbb{R}}

%% file: sec/0_abstract.tex
\begin{abstract}
We introduce AvatarAnyone, a method for reconstructing avatars from casual YouTube videos as layered meshes, with each garment and footwear as a separate mesh layered on top of the body mesh. In such videos, people differ in garment classes, clothing types, and layers, and each combination needs a different reconstruction. A single fixed layered reconstruction pipeline therefore fails. AvatarAnyone solves this problem with AI agents that plan a custom optimization for each person's reconstruction. The agents predict the garment classes, the clothing type, and the layering. They use these predictions to choose an initialization and stage the optimization. Because the avatars are layered meshes, they support applications such as virtual try-on and cloth simulation. We change the clothing layers of an avatar and render it in different garments and poses. We can also simulate a clothing layer as cloth over the animated body with a standard cloth simulator. Most existing avatars are one fused surface or a set of Gaussians and cannot do this. We load the avatars directly into Blender, Unreal Engine, and robotics simulators such as MuJoCo. As the avatars are meshes, they support contact modeling with other objects. We show results from around 150 YouTube videos. On four datasets, AvatarAnyone outperforms existing mesh-based methods for animatable monocular reconstruction. Our project page is available at \href{https://miraymen.github.io/avataranyone}{miraymen.github.io/avataranyone}.
\end{abstract}
\vspace{\dimexpr-6mm+11.9pt\relax}%

%% file: sec/1_intro.tex
\section{Introduction}
\label{sec:intro}

Animatable 3D humans are essential for applications such as AR/VR, games, virtual try-on, and embodied AI.
Digital humans are usually built using multi-camera scanning~\cite{RenderPeople} or by artists with specialized tools~\cite{epic2026metahuman, clo2026marvelous}.
Both routes are expensive and difficult to scale.
A long-standing goal of computer vision and graphics is therefore to reconstruct 3D humans automatically from ordinary video~\cite{alldieck2018video,saito2019pifu}.

Many of these applications need more than a textured surface.
Virtual try-on and cloth simulation need a layered avatar (Fig.~\ref{fig:teaser}): the body, the hair, and each garment as separate surfaces, with the body complete underneath.
Most reconstruction methods do not provide such layers.
Most recent methods represent the person with 3D Gaussian Splatting (3DGS)~\cite{kerbl2023gaussians,kocabas2023hugs,qian20233dgsavatar,hu2024gauhuman,moon2024exavatar,qiu2025lhm,zhuang2025idol,dong2025moga,mir2026ahoy}.
These avatars render well, but graphics engines~\cite{akenine2018realtime,gregory2018gameengine} cannot use them directly, and a set of Gaussians has no surface for contact or simulation.
Mesh-based methods provide a surface, but they deform one body template~\cite{alldieck2018video,alldieck19cvpr,jiang2022selfrecon} or output one fused surface~\cite{li2025pshuman,chen2025synchuman,hu2026humannova,xiang2026trellis2,li2026pixal3d}.
The body, the hair, and the clothing are then one layer, so no garment can be removed, replaced, or simulated.

Layers are hard for a reason: a layered reconstruction depends on the person.
A fused surface can come from one fixed pipeline, because the pipeline does not need to know what the person wears.
A layered avatar needs this knowledge.
A loose jacket needs a different initialization than a skin-tight top, and a skirt needs a different one than trousers.
A tucked shirt needs a different layer order than an untucked one.
No single fixed optimization handles all these cases, so a human expert usually makes these choices.

Large vision-language models (VLMs), the core of modern agentic AI systems~\cite{anthropic2026claudeopus55,google2026gemini4argon,openai2026gpt6astra}, are a natural candidate for these decisions.
From a few images, a VLM can tell if a top is loose, if a shirt is tucked in, or what class of clothing a person wears.
On its own, however, a VLM cannot recover body shape or garment geometry precisely.
Test-time optimization has the opposite strengths: it fits geometry precisely, but only from a good initialization and with the right objective.
The two are therefore complementary.

Our key insight is that the structure of a person is a set of visual questions, and an agent can answer them for each person.
AvatarAnyone takes as input a generated 360$^\circ$ video of the person, which it synthesizes from a casual YouTube video with image~\cite{openai2026gptimage25} and video diffusion~\cite{yan2026scail2}.
Agents first analyze the structure of the person.
They then select the models and adapt the optimization to the structure.
A hair model fits and regularizes the hair, and a clothing model fits and regularizes each garment.
The agents also pick the initialization for each layer, and numerical optimization fits the layers to the 360$^\circ$ video.

This design gives AvatarAnyone three advantages.
(a) The avatar carries semantics, not only geometry.
(b) The reconstruction uses the semantics to select different models and priors for different parts.
(c) The avatar is layered, so a garment can be replaced or simulated on its own.

We show results on around 150 YouTube videos.
On 100 scans from four datasets, AvatarAnyone lowers the Chamfer distance of the best mesh-based baseline from 1.92\,cm to 1.36\,cm.
In a perceptual study against mesh-based methods on YouTube videos, it receives 79.81\% of the votes.
The layers support virtual try-on and cloth simulation.
The avatars load directly into Blender, Unreal Engine, and the robotics simulator MuJoCo~\cite{todorov2012mujoco}, and support contact modeling with other objects.

To the best of our knowledge, ours is the first work to apply VLM agents to animatable, clothed 3D human reconstruction from casual monocular video.

In summary, our contributions are:

\noindent 1) AvatarAnyone, a method that reconstructs a layered, animatable mesh avatar from a casual YouTube video.

\noindent 2) An agentic formulation of human reconstruction: agents analyze the structure and semantics of a person and select matching models and initializations for the optimization.

\noindent 3) Agent-checked initializations, and a joint optimization that combines all parts into one layered avatar.

%% file: sec/2_related.tex
\begin{figure*}[t]
	\centering
	\includegraphics[width=\linewidth]{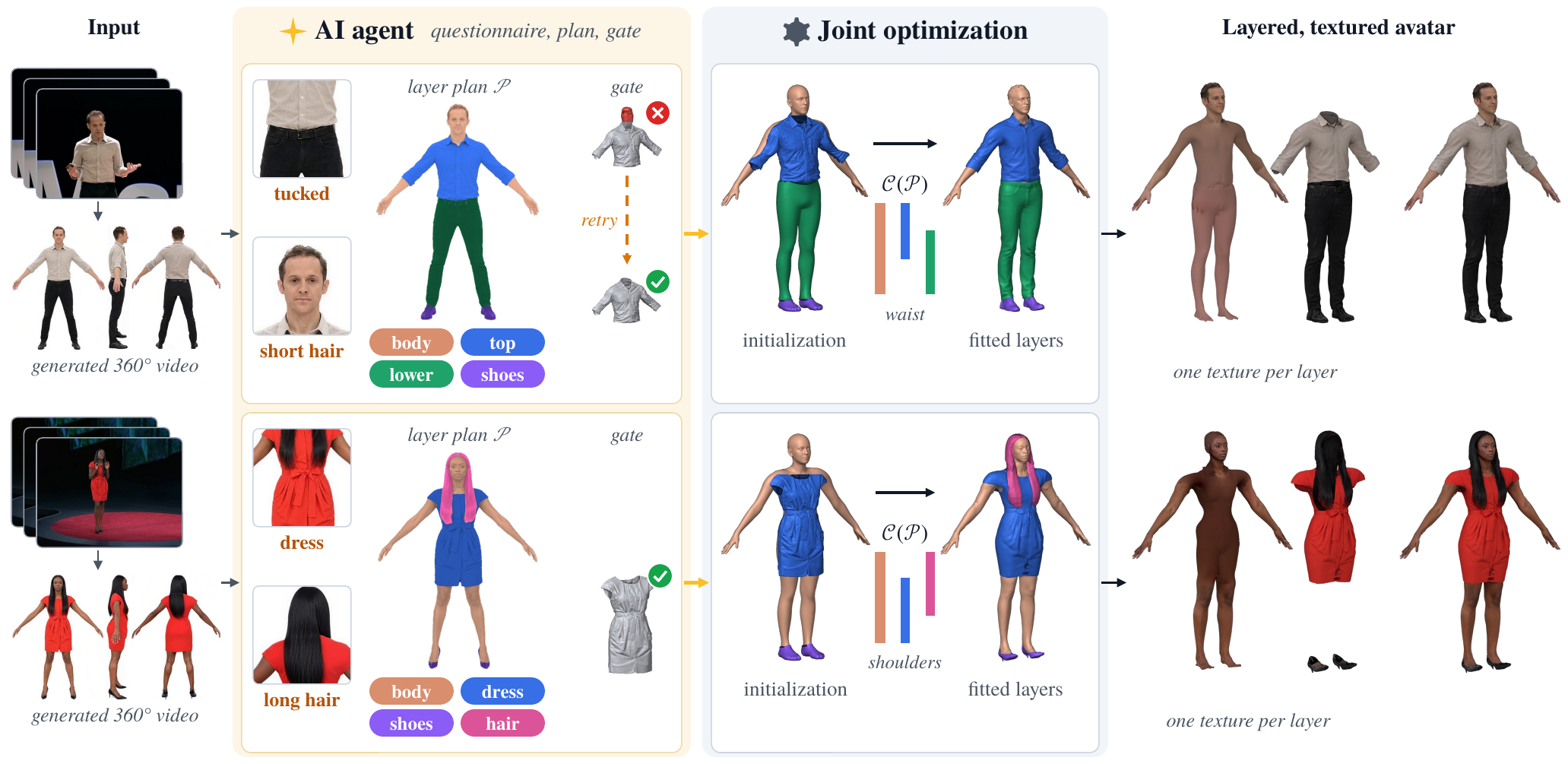}
	\caption{\textbf{Method overview.} From a YouTube video, the method generates a 360$^\circ$ video of the person. An AI agent answers a questionnaire about the clothing and the hair. The answers form the layer plan $\set{P}$, which builds the optimization for that person. \textbf{Top:} the agent finds a tucked shirt, trousers, and short hair. \textbf{Bottom:} the agent finds one garment, a dress, and long hair. The plan selects the layers, the initialization of each layer, and the order of the layers. The agent gate rejects a generated initialization that carries a head (red) and retries until an initialization is clean. The joint optimization fits all layers. The bars $\set{C}(\set{P})$ show the order of the layers at the waist and at the shoulders, from the body outwards. Each layer gets its own texture. Colors mark the layers: the body (salmon), the upper garment (blue), the lower garment (green), the shoes (violet), and the hair (pink).}
	\label{fig:method}
	\vspace{-4mm}
\end{figure*}

\section{Related work}
\label{sec:related}

\noindent\textbf{Human reconstruction and neural rendering.}
Human reconstruction methods differ in how they represent the person.
Mesh-based methods fit, clothe, and animate a parametric body template~\cite{smpl2015loper, SMPL-X:2019, Bogo2016keepitsmpl, kanazawaHMR18, mir20hps, alldieck2018video, alldieck19cvpr, mir2020pix2surf, mir2024origin}.
DensePose~\cite{guler2018densepose, neverova2018denseposetransfer, neverova2020cse} links each image pixel to a point on this template.
Implicit functions~\cite{mescheder2019occupancy, park2019deepsdf, chen2021snarf, alldieck2021imghum, saito2020pifuhd, he2021arch++, huang2020arch, deng2020nasa} replace the mesh with a continuous field of arbitrary topology.
Both representations recover geometry but lack photorealism or reposability.
NeRF~\cite{mildenhall2020nerf} and its extensions~\cite{mueller2022instant, barron2022mip, barron2023zipnerf, nerfstudio, nerf_review} render photorealistic novel views, and 3DGS~\cite{kerbl2023gaussians} reduces their cost with explicit Gaussian primitives~\cite{lassner2021pulsar, westover1991phdsplatting}.
Avatars inherit these strengths: NeRF-based methods~\cite{peng2021animatable, guo2023vid2avatar, weng_humannerf_2022_cvpr, jian2022neuman, haberman2023hdhumans, heminggaberman2024trihuman, li2022tava, liu2021neural, xu2021hnerf} produce photorealistic renderings from video, and 3DGS-based avatars~\cite{guo2025vid2avatarpro, kocabas2023hugs, moreau2024human, abdal2023gaussian, zielonka2023drivable, moon2024exavatar, li2024animatablegaussians, pang2024ash, lei2023gart, hu2024gaussianavatar, li2024relightableandanimatable, zheng2024gpsgaussian, jiang2024robust, dhamo2023headgas, qian20233dgsavatar, xu2024gaussian, junkawitsch2025eva} render in real time.
However, a 3DGS avatar is a set of Gaussians, not a surface, so graphics, game, and physics engines cannot use it directly.
AvatarAnyone therefore returns to meshes, the native representation of these engines, with one layer per garment.

\noindent\textbf{Humans and garments.} Some methods model each garment as a separate layer, but only for garment classes learned from 3D scans or simulation~\cite{bhatnagar2019mgn, ponsmoll2017clothcap, patel2020tailornet, tiwari2020sizer, jiang2020bcnet, corona2021smplicit}.
Others represent the clothing, and sometimes the hair, as NeRFs~\cite{feng2022scarf, feng2023delta}, implicit fields~\cite{guo2024reloo, chen2025d3human}, or Gaussians~\cite{chen2024ggavatar, lin2024layga, chen2026singlecamera} over a body layer.
Further methods output garment meshes from monocular video~\cite{qiu2023recmv}, a 3D scan~\cite{kim2024gala}, or a single image~\cite{li2025dress1to3}.
None of them reconstructs the body, the hair, and each garment as separate textured meshes from occluded video.

\noindent\textbf{Humans from monocular input.}
Early implicit methods~\cite{saito2019pifu, zheng2020pamir, xiu2023econ, zhang2024sifu, huang2024tech} and optimization-based approaches~\cite{jiang2022selfrecon, hu2024gauhuman, sim2025persona} reconstruct humans from single images or monocular video.
Feed-forward models~\cite{qiu2025lhm, qiu2025pflhm, hong2023lrm, zhuang2025idol} predict animatable avatars in seconds.
Many methods use diffusion to synthesize multi-view images as an intermediate step~\cite{pan2025humansplat, peng2024charactergen, he2024magicman, qiu2025anigs, lu2025gas, chen2025synchuman, li2025pshuman, huang2025adahuman, buehler2026dreamliftanimate, dong2025moga, taubner2025cap4d}, and others extend to 4D or unconstrained settings~\cite{tan2025dressrecon, pang2024disco4d, shao2024human4dit, yang2024havefun, lin2025omnihuman}.
All these methods assume a fully visible subject as input.
Diffusion models~\cite{Ho2020DDPM, Rombach2022LDM} now generate realistic video~\cite{blattmann2023svd, yang2024cogvideox, kong2025hunyuanvideo, wan2025wan}, and pose-driven variants animate a person from one image~\cite{hu2024animateanyone, xu2024magicanimate, zhu2024champ, zhang2024mimicmotion, tu2025stableanimator, wang2025unianimate_dit, yan2026scail2}.
AHOY~\cite{mir2026ahoy} reconstructs 3DGS avatars from occluded YouTube videos with a video diffusion prior.
AvatarAnyone also completes the occluded person with generative models, but it reconstructs layered meshes instead of Gaussians.

\noindent\textbf{Agentic 3D reconstruction.} Agentic 3D reconstruction is a fast-evolving area of research.
Language models solve visual tasks with programs that call vision tools~\cite{gupta2023visprog, suris2023vipergpt}.
In 3D, a language model writes or edits graphics code that produces the shape or the scene~\cite{sun20253dgpt, hu2024scenecraft, huang2024blenderalchemy, kulits2024rethinking, shen2026fourdcodebench}.
Other systems express single images~\cite{he2026thinking, yin2026viga, li2026legoanything}, articulated objects~\cite{mandi2025real2code, le2025articulateanything, zhou2026articraft}, and indoor scenes~\cite{huang2026litereality} as executable programs.
Agents also steer reconstruction from images~\cite{sun2026harmony, luo2026scenerig, wu2026gator, xie2026codimrecon} and monocular video: AgentSTAR~\cite{mazur2026agentstar} pairs a VLM agent with an optimizer to reconstruct and track rigid and articulated objects.
\mbox{AGILE}~\cite{shi2026agile} and HAT-4D~\cite{li2026hat4d} use VLMs to guide 3D generation for hand--object and multi-object interactions.

For humans, language models estimate body pose~\cite{feng2024chatpose}, call tools for 3D human analysis~\cite{lin2025chathuman}, predict sewing patterns~\cite{korosteleva2023garmentcode, bian2025chatgarment, nakayama2025aipparel, jiang2026dressagent}, and score body meshes~\cite{shen2026vlmguided}.
SmartAvatar~\cite{huangmenders2025smartavatar} uses VLM agents to generate a rigged avatar from a photo or a text prompt.
We note that SmartAvatar is a preprint that, to our knowledge, has not been peer reviewed.
Its agents set the parameters and choose the preset garments of an off-the-shelf parametric human generator.
The garments are therefore library assets, not reconstructions.
None of these methods reconstructs a layered, textured avatar of a person from video.
Clothed humans differ from objects, because the structure of the optimization changes with the garments of each person.
In AvatarAnyone, agents answer visual questions about the garments, and the answers build the optimization for that person.

%% file: sec/3_method.tex
\section{Method}
\label{sec:method}

Our goal is to reconstruct a layered, animatable mesh avatar of a person, with a separate mesh for the body, each garment, footwear, and the hair, from a YouTube video.
The difficulty is that every person needs a different optimization.
The layers and their order depend on what the person wears.
AvatarAnyone therefore divides the work (Fig.~\ref{fig:method}).
Agents read the structure of the person and build the optimization from it (Secs.~\ref{subsec:questionnaire} and~\ref{subsec:initialization}).
A custom joint optimization then fits all layers (Sec.~\ref{subsec:arrangement}), and each layer gets its own texture (Sec.~\ref{subsec:checks}).

\subsection{Layered avatar and input}
\label{subsec:inputs}

The avatar is a set of mesh layers $\set{A} = \{\set{M}_k\}_{k \in \set{K}}$.
The index set $\set{K} \subseteq \{0, \dots, 4\}$ holds the layers of the person: the body ($k{=}0$), the hair ($1$), the upper garment ($2$), the lower garment ($3$), and the shoes ($4$).
Each layer has its own vertices, texture, and skinning weights.
SMPL-X~\cite{SMPL-X:2019} initializes the body, and linear blend skinning poses every layer with the SMPL-X skeleton.

The input is a generated 360$^\circ$ video of the person in one pose.
A casual video rarely shows the person from all sides, so we generate this video.
An image model~\cite{openai2026gptimage25} completes the person in a canonical front view, and a video model~\cite{yan2026scail2} turns this view into a 360$^\circ$ video of the person in canonical pose.
Existing models then give every view per-pixel labels from the classes in $\set{K}$~\cite{carion2025sam3}, normals~\cite{khirodkar2026sapiens2}, and a body pose~\cite{sarandi2024nlf}. The generated 360$^\circ$ video, the labels, and the normals are the targets of the optimization.
For details, see the supplementary material.

\subsection{Agentic structure analysis}
\label{subsec:questionnaire}

Our key idea is to treat the structure of a person as a set of visual questions.
Does the person wear a dress or two garments?
Is the shirt tucked in?
Does long hair lie over the clothes?
A fixed pipeline assumes one answer to each question.
An agent answers them for each person.

\noindent\textbf{Questionnaire.}
The agent is a vision-language model~\cite{anthropic2026claudeopus55}.
Before any optimization, it sees the canonical view and the generated video from four sides, with close-ups of the waist and the feet.
It answers a fixed questionnaire.
The questions cover the garments, the order of the layers at the waist, the way the trousers meet the shoes, and the hair.
The answers change the optimization.
The supplementary material lists all questions and the effect of each answer.

\noindent\textbf{Layer plan.}
The answers form a layer plan $\set{P}$.
The plan fixes the layers $\set{K}$ and their semantics.
For example, a dress removes the lower garment and long hair adds a hair layer.
The plan also selects a model for each layer: an initialization (Sec.~\ref{subsec:initialization}) and the terms that fit the layer (Sec.~\ref{subsec:arrangement}).

\subsection{Agent-checked initialization}
\label{subsec:initialization}

The joint optimization never changes the topology of a layer, so it cannot repair a wrong initialization.
A skirt initialized as trousers keeps two legs, and a loose top initialized from the body surface has no free hem.
We therefore select an initialization that matches the class of each layer. 
Tight garments such as trousers and shoes are initialized adaptively from cutouts of the initial SMPL-X mesh. These initializations also vary depending on the agent's plan. See supp mat. for details.
Loose garments and long hair are initialized from a mesh that an image-to-3D model~\cite{li2026pixal3d} builds from eight views of the generated video. For alignment of the initialization mesh to the video and more details, please see supp mat.

\noindent\textbf{Agent gate.}
Generative models fail silently, and a wrong initialization yields incorrect avatars.
For example, a generated top can carry the head of the person (Fig.~\ref{fig:ablation}). The agent therefore compares renders of each generated initialization with the input views, and it rejects a faulty one.
After a rejection, the agent asks the image-to-3D model for new candidates from changed inputs: other seeds, gray-scale views, or fewer views.
The agent picks the best candidate among at most 15.
It also picks the prompts to use for SAM 3 segmentation ~\cite{carion2025sam3}. It then checks the labels: where two segmentation modes of SAM~3 (image vs.\ video) disagree, it selects the mode closer to the image.

\noindent\textbf{Garments under hair.}
Long hair hides a part of the upper garment in the generated video.
When the agent detects a person with a hair layer, it asks the image model to first remove the hair from the canonical view.
The video model generates a second 360$^\circ$ video from this view.
The initialization of the upper garment comes from this second video, so the garment is complete under the hair.

\noindent\textbf{Openings.}
A generated mesh often has improper holes.
The agent checks and closes every boundary of a garment, except the neckline, the cuffs, and the hem.
It also cuts the initial hair mesh to match its silhouettes.

\subsection{Structure-aware joint optimization}
\label{subsec:arrangement}

A custom joint optimization fits all layers, using the initialization from Sec. \ref{subsec:initialization}.
Its variables $\Theta$ are the pose and the shape of the body, its vertex offsets, and the vertices of other layers.
The energy $E$ has terms that the layer plan sets:
\begin{equation}
	\begin{aligned}
		E(\Theta) ={}& \textstyle\sum_{k \in \set{K}} \big( E^{k}_{\mathrm{mask}} + \lambda_{\mathrm{n}} E^{k}_{\mathrm{n}} \big) + E_{\mathrm{reg}} \\
		& + \lambda_{\mathrm{c}} \!\!\textstyle\sum\limits_{(o,i) \in \set{C}(\set{P})}\!\! E_{\mathrm{c}}(\set{M}_o, \set{M}_i) + \!\!\textstyle\sum\limits_{h \in \set{H}(\set{P})}\!\! \lambda_{h} E_{h}.
	\end{aligned}
	\label{eq:fit}
\end{equation}
The mask term $E^{k}_{\mathrm{mask}}$ fits the silhouette of layer $k$ and the normal term $E^{k}_{\mathrm{n}}$ fits its normals, only to the pixels with its label, in all views.
$E_{\mathrm{reg}}$ collects the landmark and smoothness terms, and $\lambda$ are weights.
The plan sets the pairs $\set{C}(\set{P})$, in which layer $o$ contains layer $i$, and the hem terms $E_h$, $h \in \set{H}(\set{P})$: these terms layer the body and garments correctly and the hem terms constrain the vertices where two separate layers meet.
We optimize Eq.~\ref{eq:fit} from coarse to fine, raising the image resolution and the number of mesh vertices at each stage. See supp mat. for details.

\noindent\textbf{Normal term.}
The estimated normals carry the folds and the wrinkles of the clothing.
The normal term moves the vertices of layer $k$ until its rendered normals match them:
\begin{equation}
	E^{k}_{\mathrm{n}} = \frac{1}{|\set{V}|} \sum_{t \in \set{V}} \sum_{\vec{p} \in \Omega^{k}_{t}} \bar{w}_t(\vec{p}) \Big( 1 - \vec{n}_t(\vec{p})^{\top} \hat{\vec{n}}_t(\vec{p}) \Big).
	\label{eq:normal}
\end{equation}
Here $\set{V}$ is the set of views of the generated video (Sec.~\ref{subsec:inputs}), and $\Omega^{k}_{t}$ holds the pixels of view $t$ that carry label $k$ and show layer $k$ in front.
At pixel $\vec{p}$, $\hat{\vec{n}}_t(\vec{p})$ is the estimated normal, and $\vec{n}_t(\vec{p})$ is the normal of the rendered mesh.
A differentiable rasterizer interpolates $\vec{n}_t(\vec{p})$ from the vertex normals of the layer, so the gradient of $E^{k}_{\mathrm{n}}$ moves its vertices.
The weights $\bar{w}_t(\vec{p})$ sum to one over $\Omega^{k}_{t}$ and come from the view selection below. 

\noindent\textbf{Layer order.}
The containment term $E_{\mathrm{c}}(\set{M}_o, \set{M}_i)$ keeps layer $i$ at least 3\,mm inside layer $o$ where the two overlap.
The pairs $\set{C}(\set{P})$ encode the order of the layers.
Every garment contains the body.
For an untucked top, the top also contains the lower garment, so $(2,3) \in \set{C}(\set{P})$.
The lower garment then extends under the top, where no view sees it.
For a tucked top, the order inverts near the waist.
The hair is always placed over the body and the upper garment.

\noindent\textbf{Hems.}
The hem terms $E_{h}$, with $h \in \set{H}(\set{P})$, act where garments end.
Trousers over shoes stay outside the shoes, and their hem stays below the top of the shoe.
For trousers tucked into boots, the constraint is reversed.

\noindent\textbf{View selection.}
The views of the generated video are not exactly consistent, because video diffusion models do not enforce multi-view consistency.
A mean over all views therefore produces blurry results.
Following texture mapping~\cite{waechter2014color}, we assign each face of the mesh to one view that sees it frontally.
A smoothness prior keeps neighboring faces in the same view.
The face, the scalp, and the neck are rigid and empirically the most consistent across views, so they average all views that see them frontally.
 In the finer stages, the frontal view takes most of the weight $\bar{w}_t$ in Eq.~\ref{eq:normal}.

\subsection{Layer-order check and layered texture}
\label{subsec:checks}

\noindent\textbf{Layer-order check.}
After the optimization, skin can still show through a garment in a few views.
A layered avatar makes this error measurable.
We render the avatar in every view and find the pixels where the labels show a garment but another layer lies in front.
Where the garment lies at most 1\,cm behind, we push the body back under it.

\noindent\textbf{Ears and hands.} The optimization has no term for the ears, so we complete them afterward. An ear fitted to views in which hair covers it lands on the hair. The agent therefore decides if hair covers the ears, and we fit each ear only where it is visible. We also refit the fingers of each hand.

\noindent\textbf{Layered texture.}
Each layer takes its color only from the pixels with its own label.
Garment colors therefore do not bleed onto the skin, and the body has a skin texture under the clothes.
A garment can then be removed or replaced (Sec.~\ref{subsec:tryon}).
As in the optimization, each face takes its color from one view~\cite{waechter2014color}.
The face and the front of the garments come from the most frontal view.
We also separate the color from the light of the generated video.
The camera and the light stay fixed while the person turns.
We thus fit one light, a gain per view, and an albedo per surface point to the colors:
\begin{equation}
	\min_{\mat{L},\, \{\vec{g}_t\},\, \{\vec{a}_{\vec{x}}\}} \sum_{\vec{x}} \sum_{t \in \set{V}_{\vec{x}}} \big\| \vec{c}_{\vec{x},t} - \vec{g}_t \odot \vec{a}_{\vec{x}} \odot \mat{L}^{\top} \vec{y}(\vec{n}_{\vec{x},t}) \big\|^2.
	\label{eq:light}
\end{equation}
Here $\vec{c}_{\vec{x},t}$ is the color of the surface point $\vec{x}$ in view $t$, and $\set{V}_{\vec{x}}$ holds the views that see $\vec{x}$.
The vector $\vec{y}(\vec{n}_{\vec{x},t}) \in \real^{9}$ holds the spherical harmonics up to second order of the normal of $\vec{x}$ in the camera frame of view $t$.
$\mat{L} \in \real^{9 \times 3}$ is the light, $\vec{g}_t \in \real^{3}$ is the gain of view $t$, $\vec{a}_{\vec{x}} \in \real^{3}$ is the albedo, and $\odot$ is the element-wise product.
Alternating least squares solves Eq.~\ref{eq:light} on points that at least three views see.
The texture is the albedo in the selected view, $\vec{a}^{*}_{\vec{x}} = \vec{c}_{\vec{x},t^{*}} \oslash \big(\vec{g}_{t^{*}} \odot \mat{L}^{\top} \vec{y}(\vec{n}_{\vec{x},t^{*}})\big)$, where $t^{*}$ is the view selected for $\vec{x}$ and $\oslash$ divides element-wise.

%% file: sec/4_exp.tex
\begin{figure*}[t]
	\centering
	\setlength{\tabcolsep}{1pt}
	\ifdefined\comph\else\newlength{\comph}\fi
	\setlength{\comph}{3.2cm}
	\ifdefined\compsh\else\newlength{\compsh}\fi
	\setlength{\compsh}{3.05cm}%
	\newcommand{\compcell}[1]{\raisebox{-0.5\height}{\includegraphics[width=0.105\linewidth,height=\comph]{#1}}}
	\newcommand{\compcellT}[1]{\raisebox{-0.5\height}{\includegraphics[height=\compsh]{#1}}}%
	\newcommand{\complabel}[1]{\raisebox{-0.5\height}{\rotatebox{90}{\scriptsize #1}}}
	\begin{tabular}{@{}ccccccccc@{}}
		\scriptsize Input & & \scriptsize PSHuman & \scriptsize SyncHuman & \scriptsize HumanNOVA & \scriptsize TRELLIS.2 & \scriptsize Pixal3D & \scriptsize Ours & \scriptsize GT \\
		\compcellT{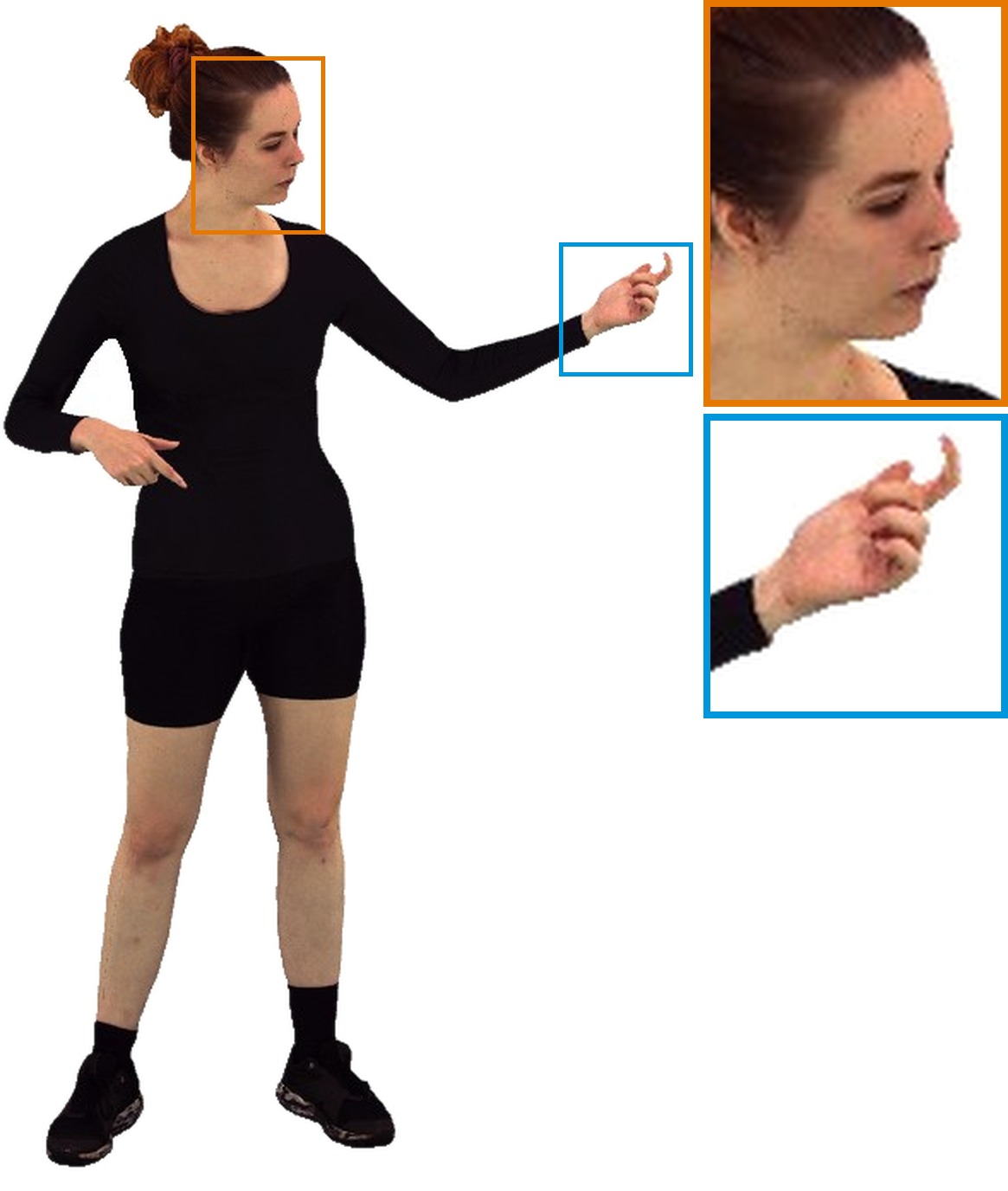} & \complabel{Static} & \compcellT{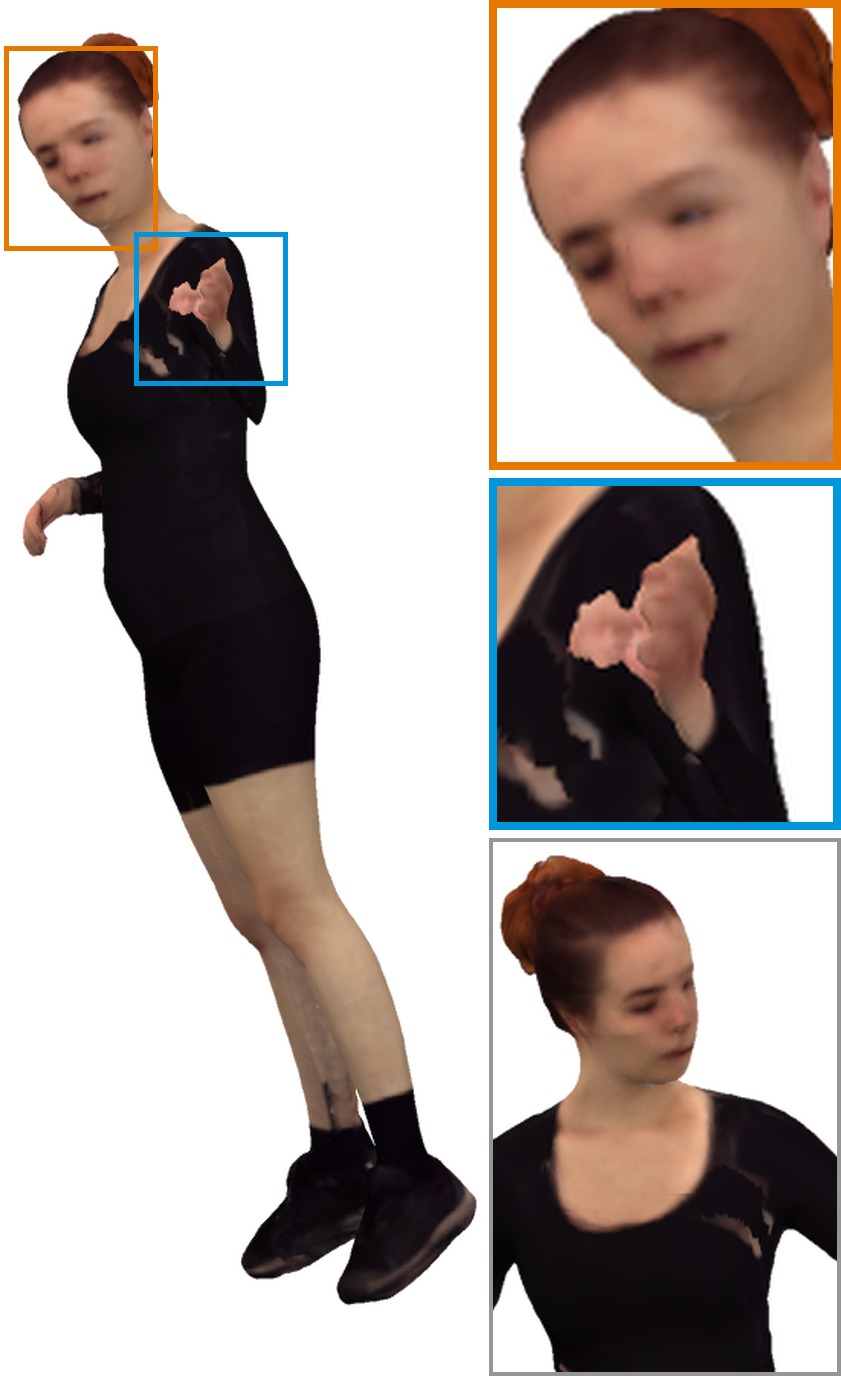} & \compcellT{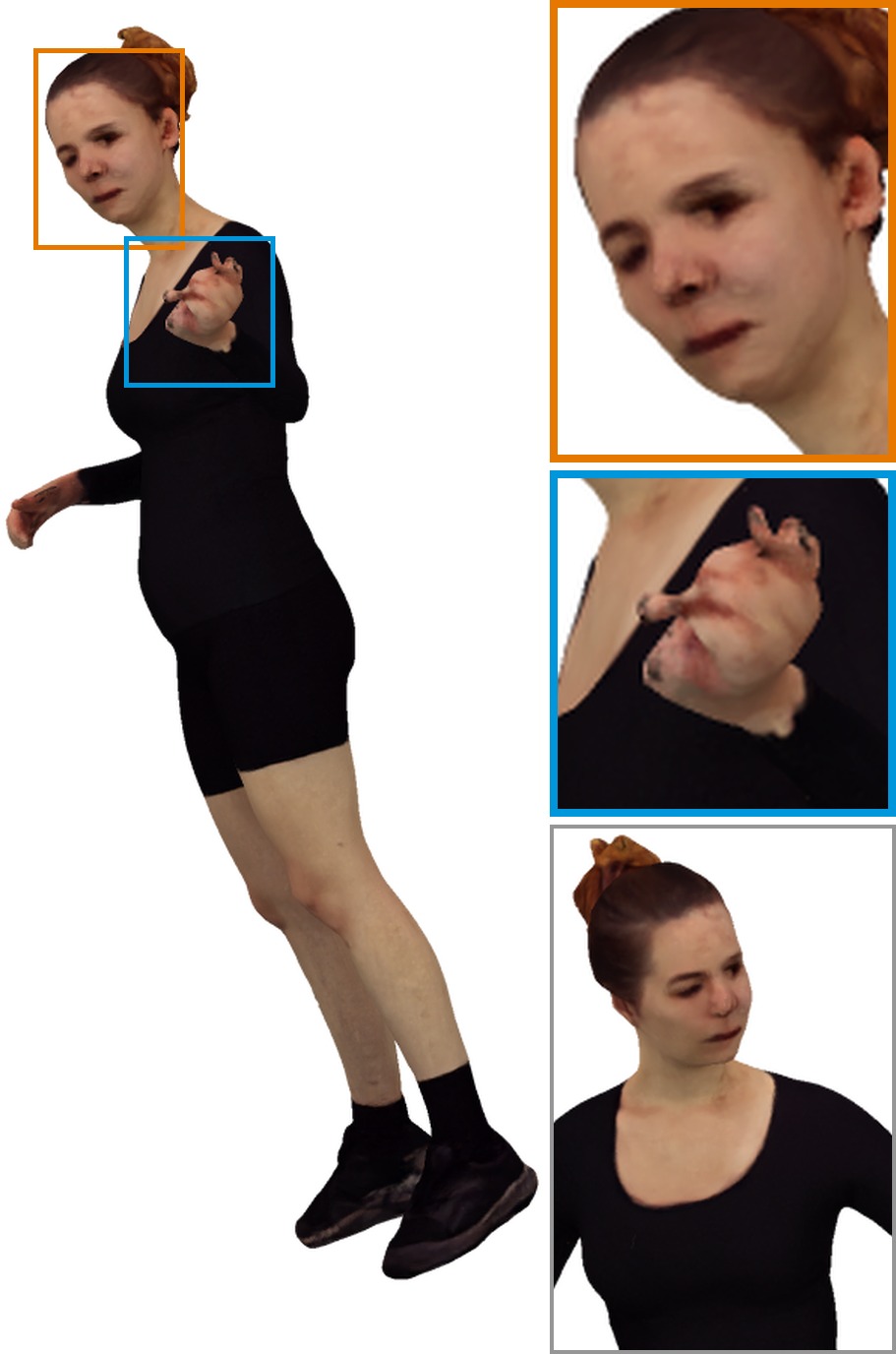} & \compcellT{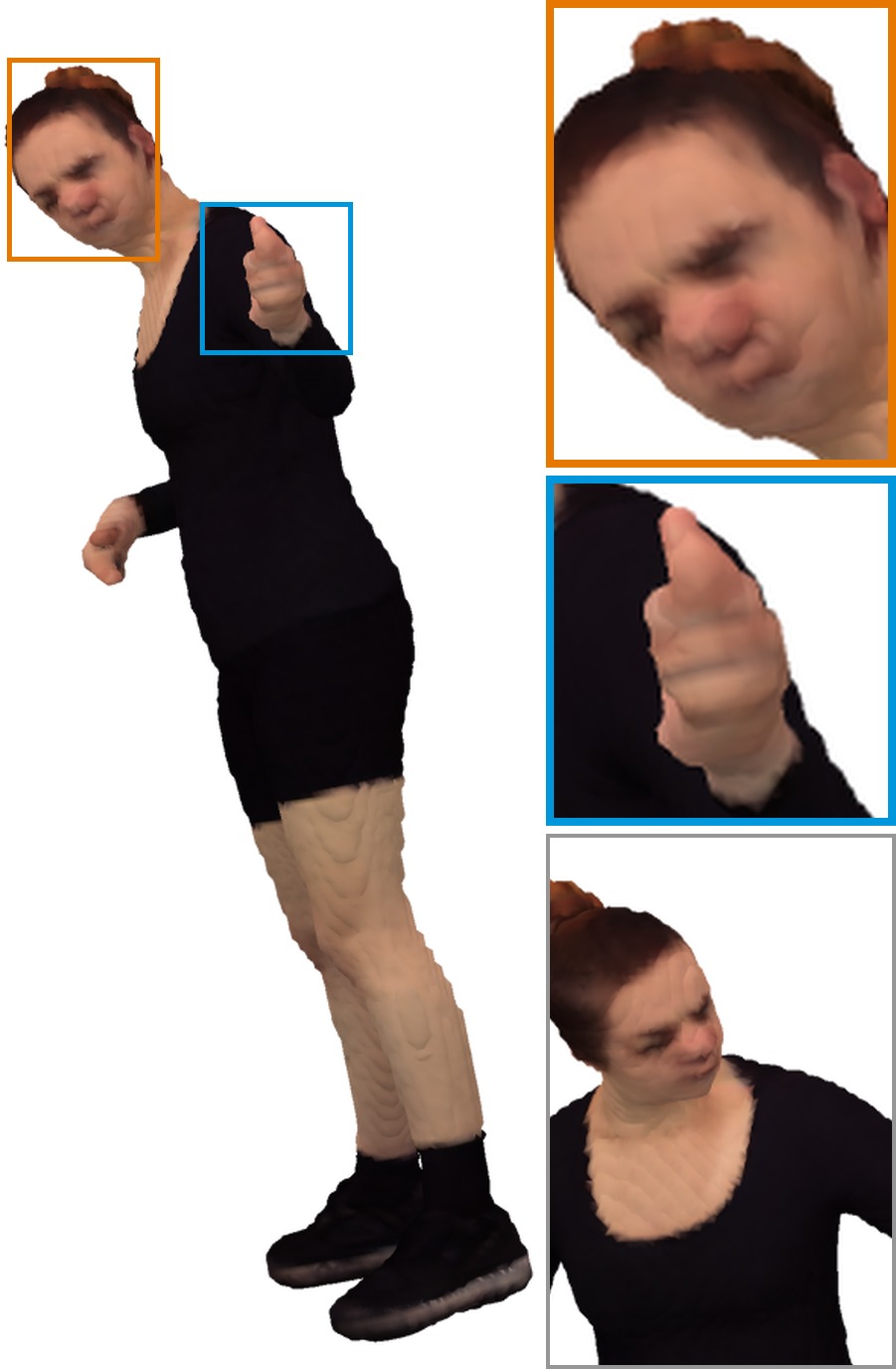} & \compcellT{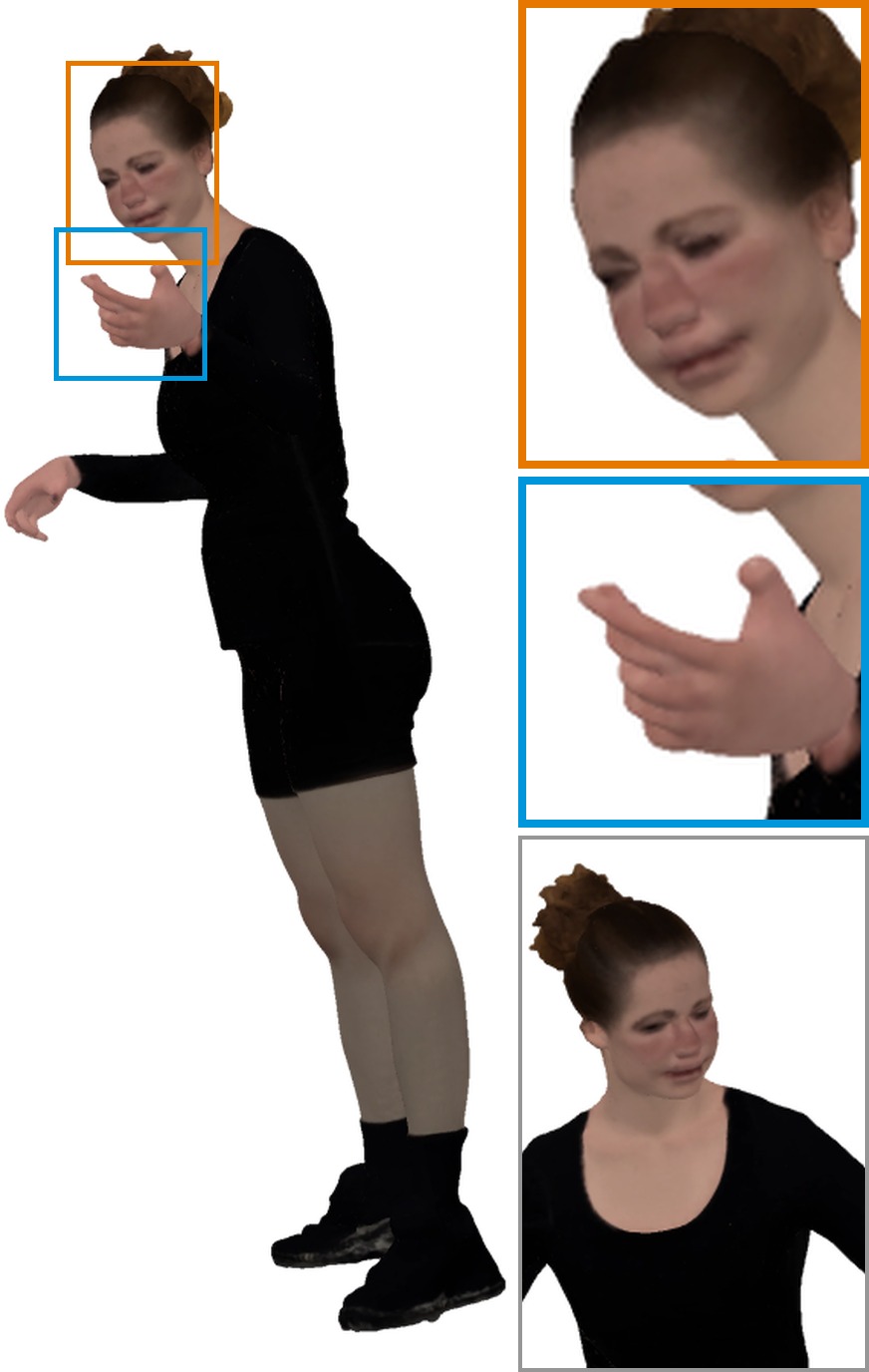} & \compcellT{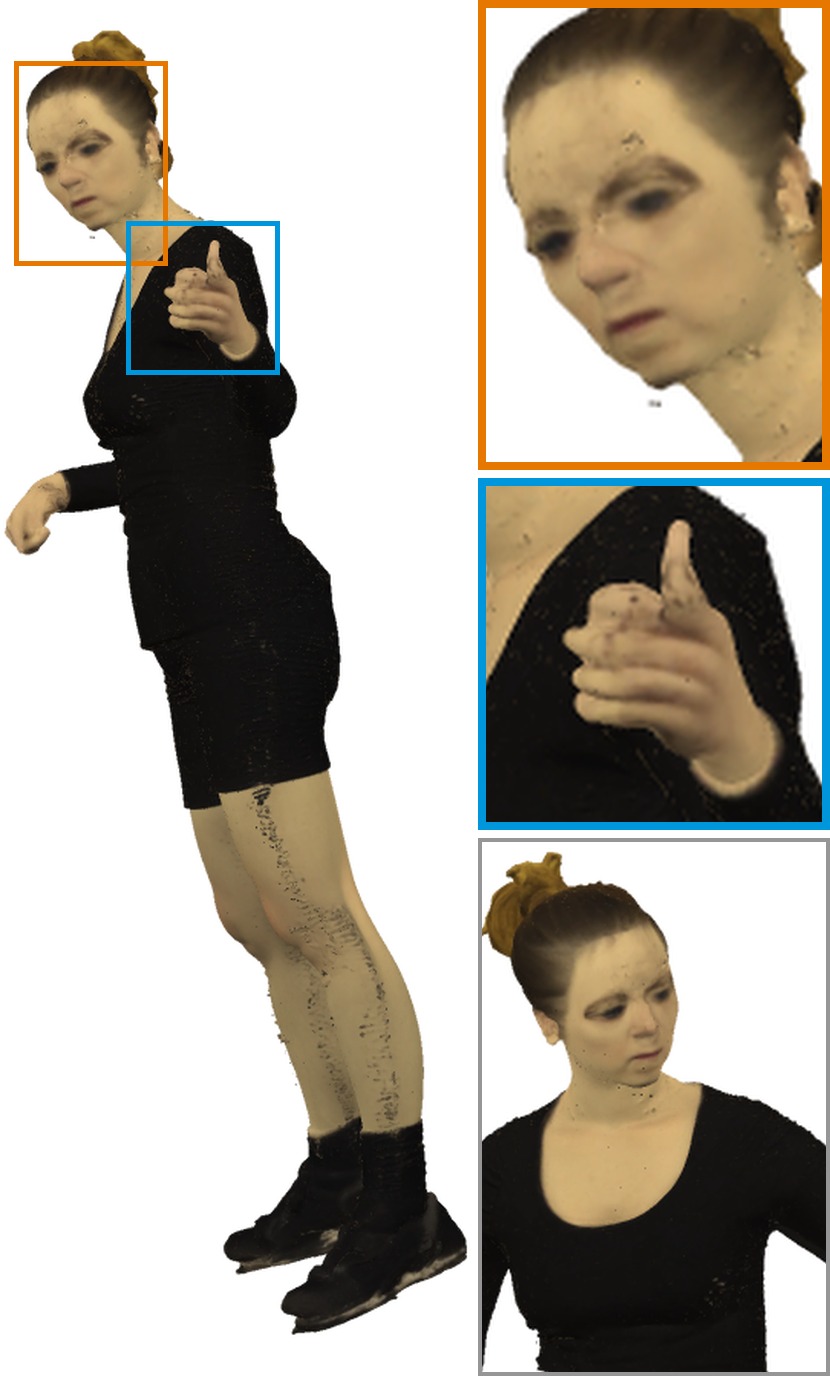} & \compcellT{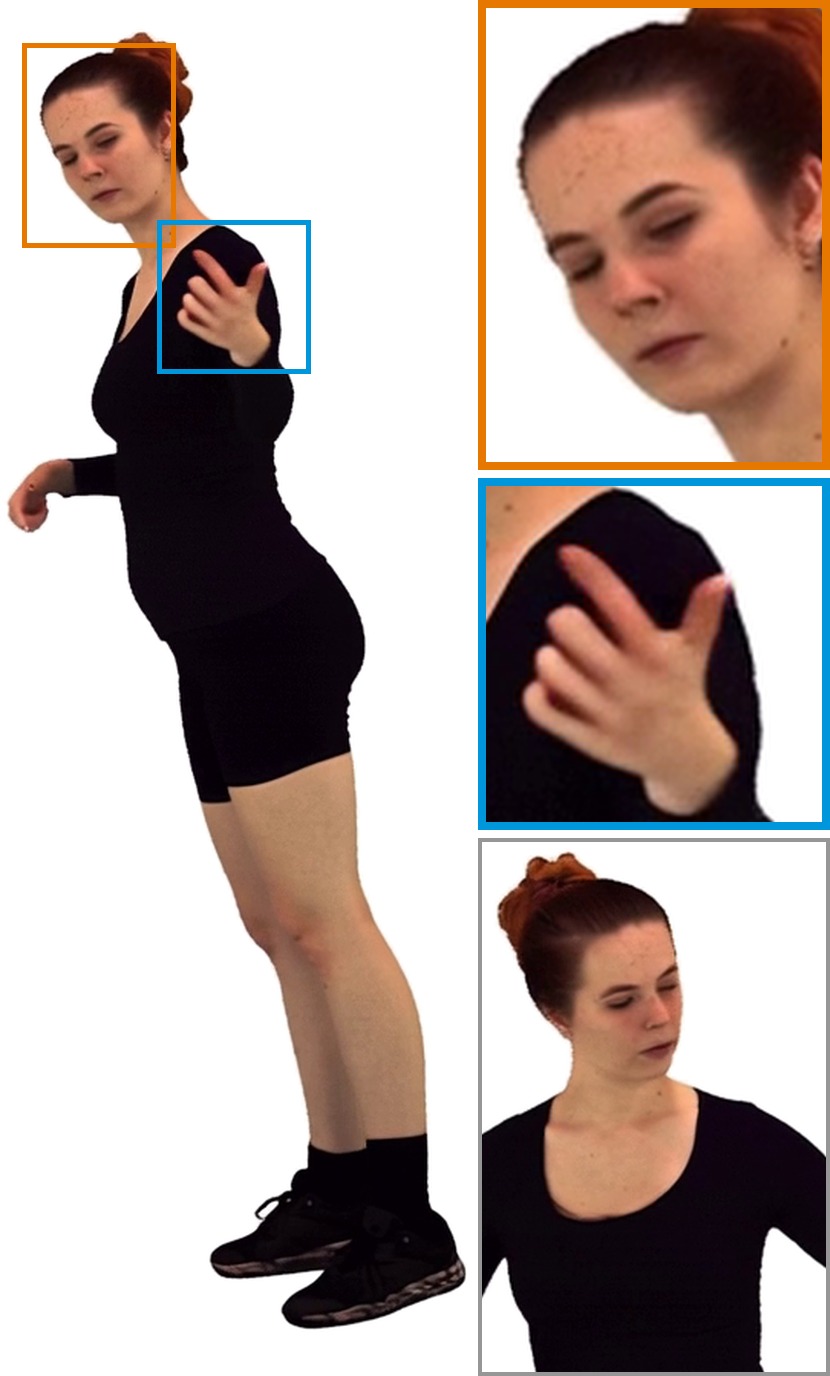} & \compcellT{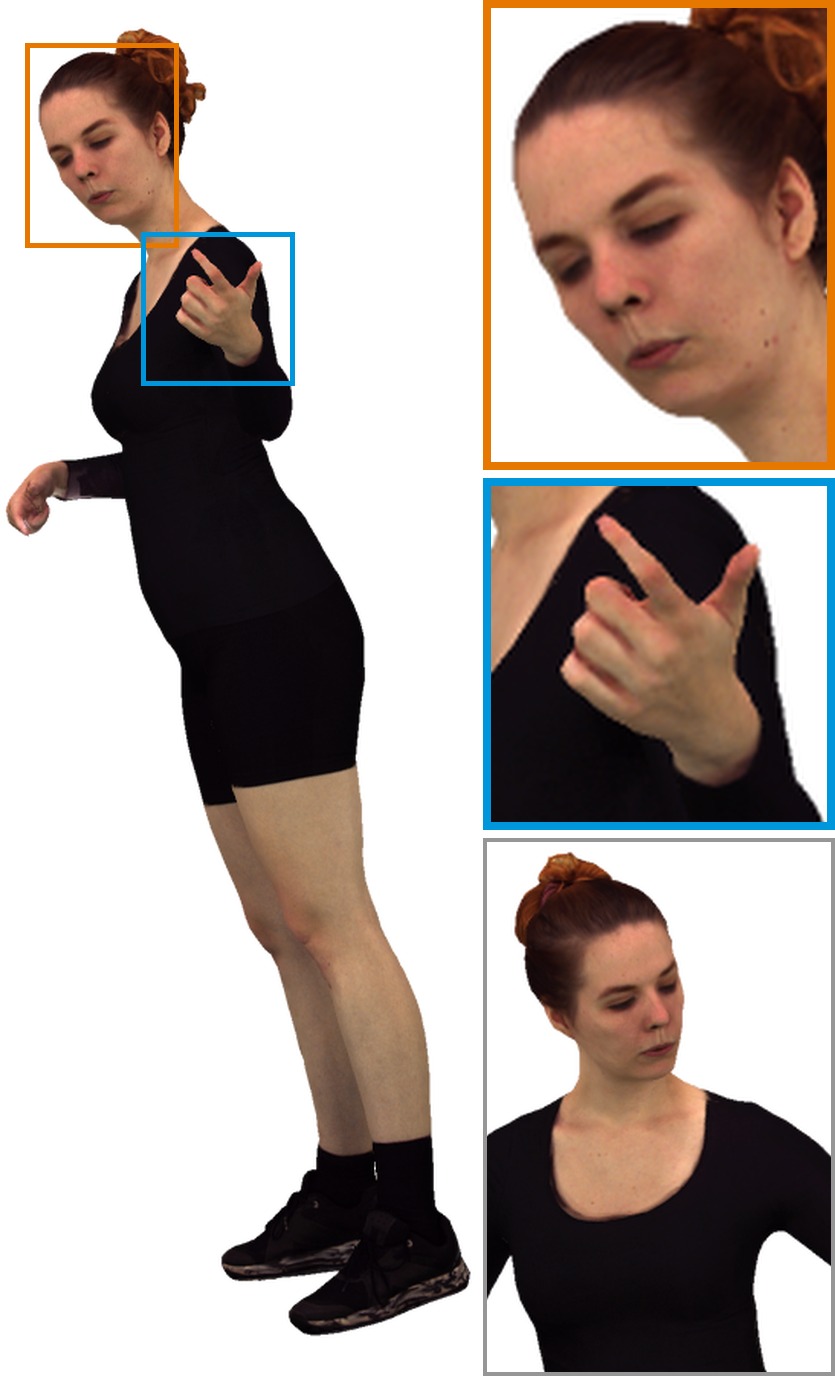} \\[2pt]
		\compcellT{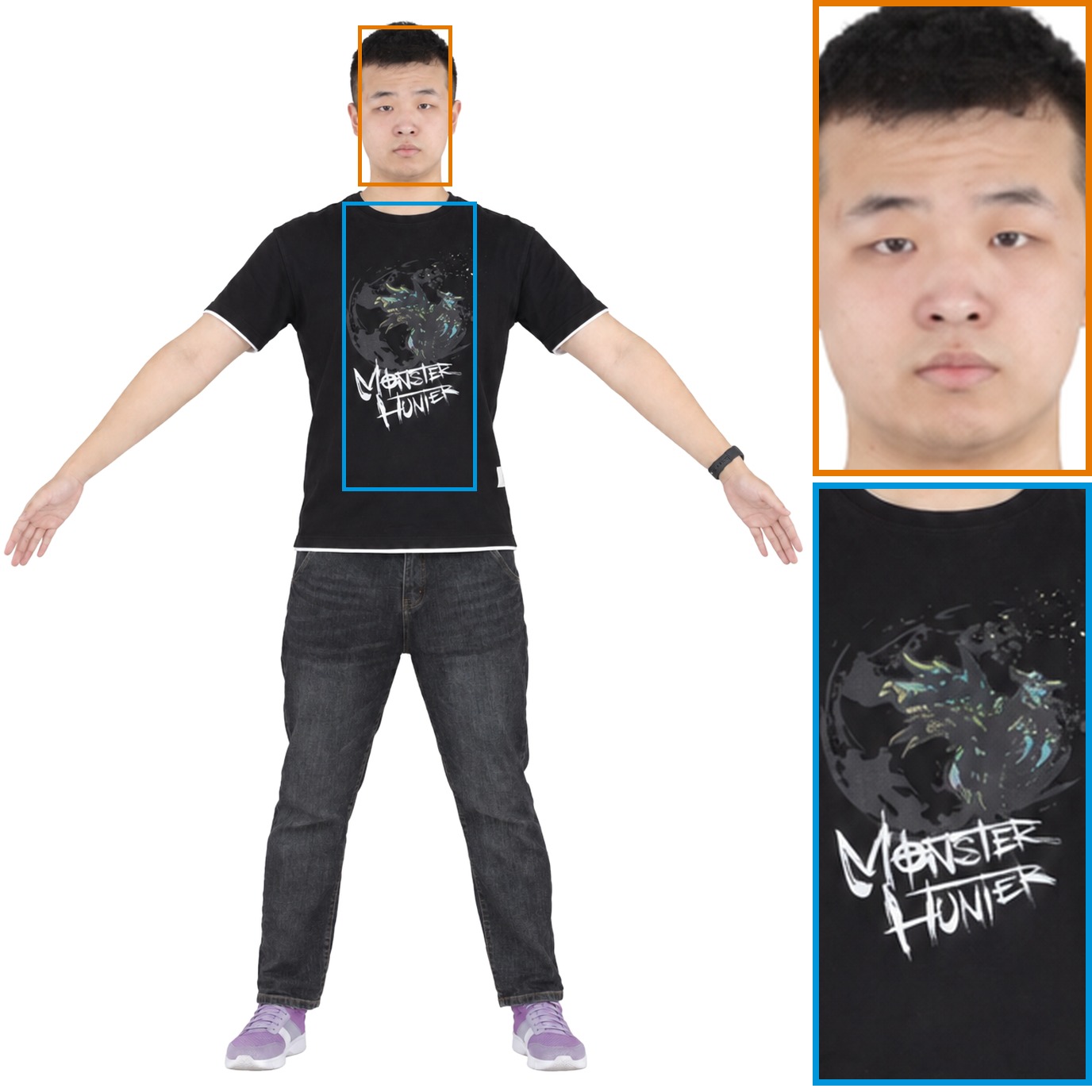} & \complabel{Animatable} & \compcellT{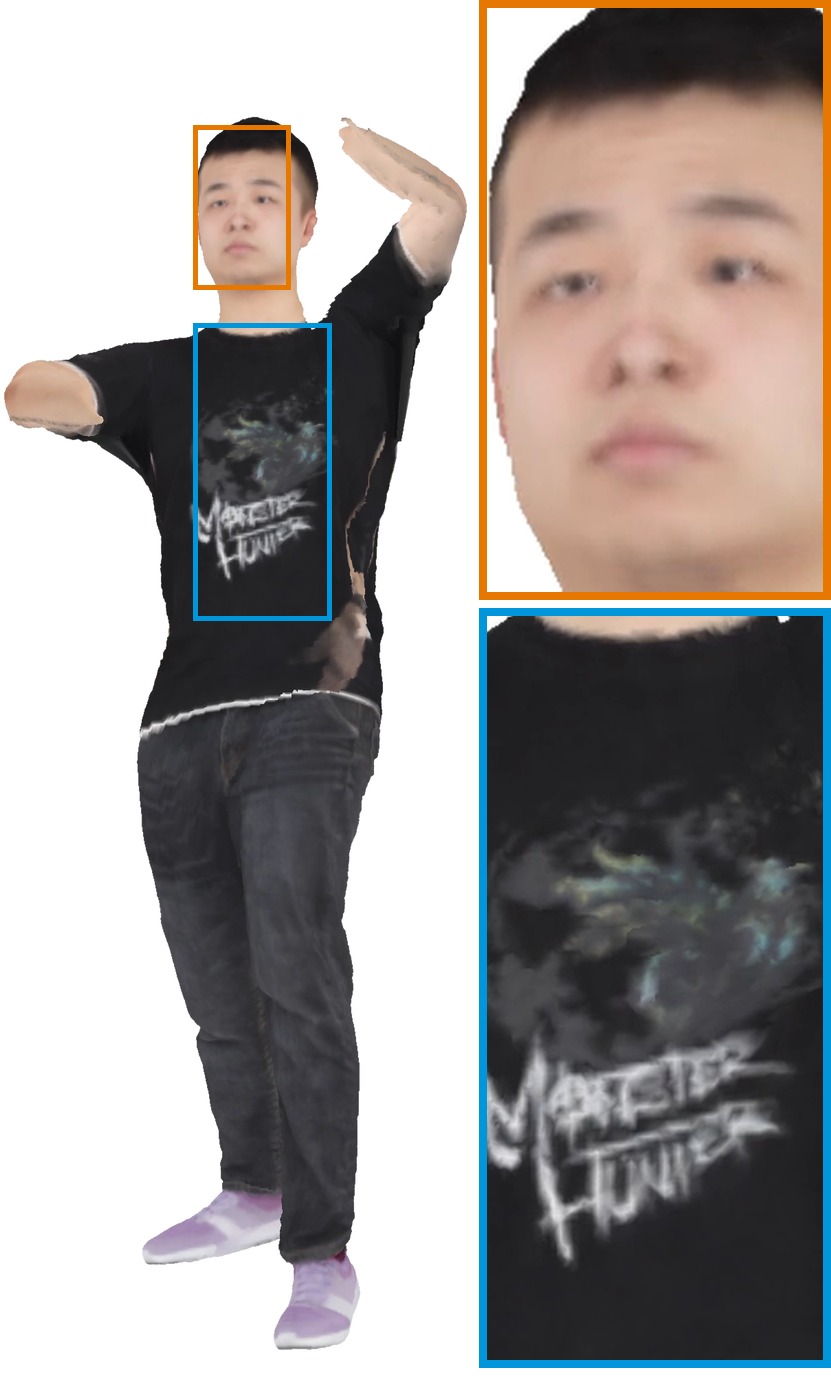} & \compcellT{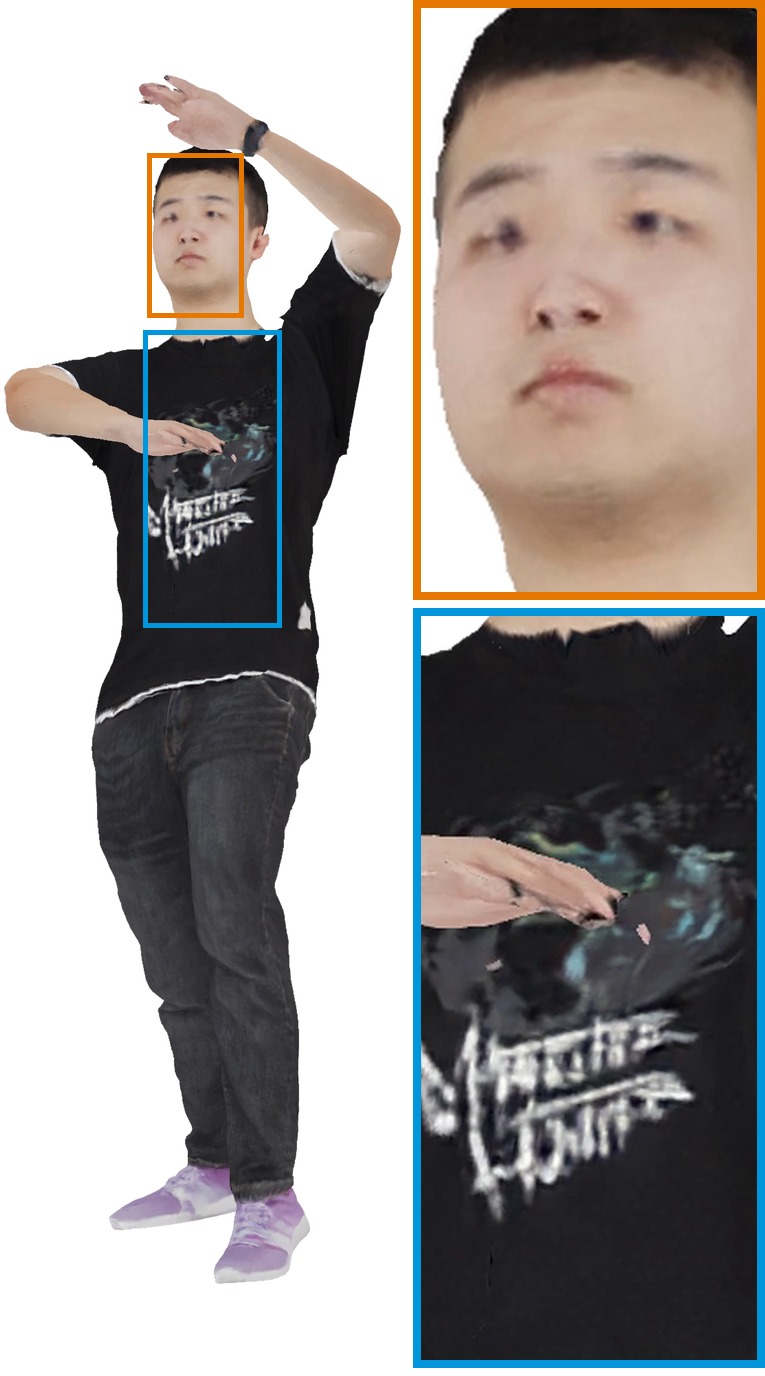} & \compcellT{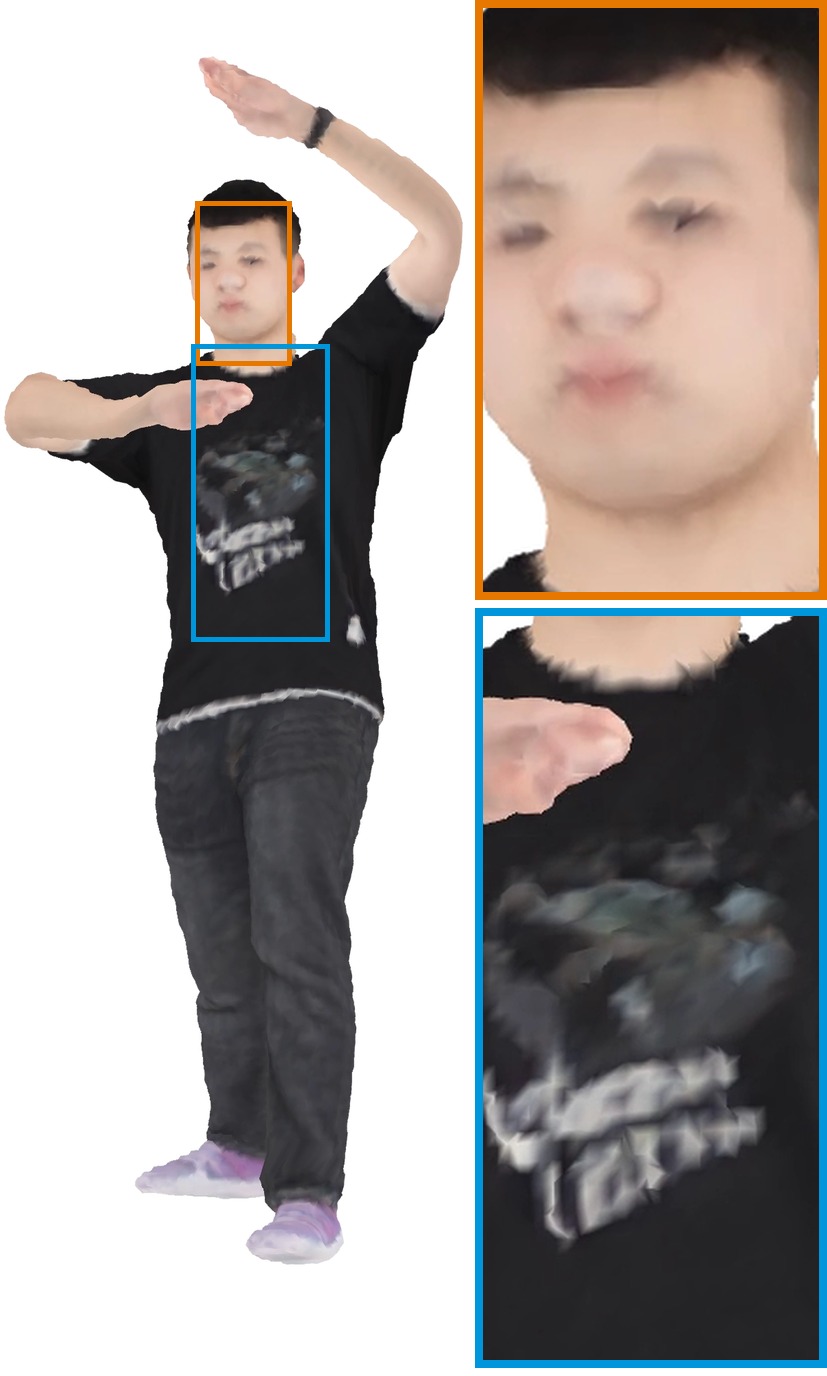} & \compcellT{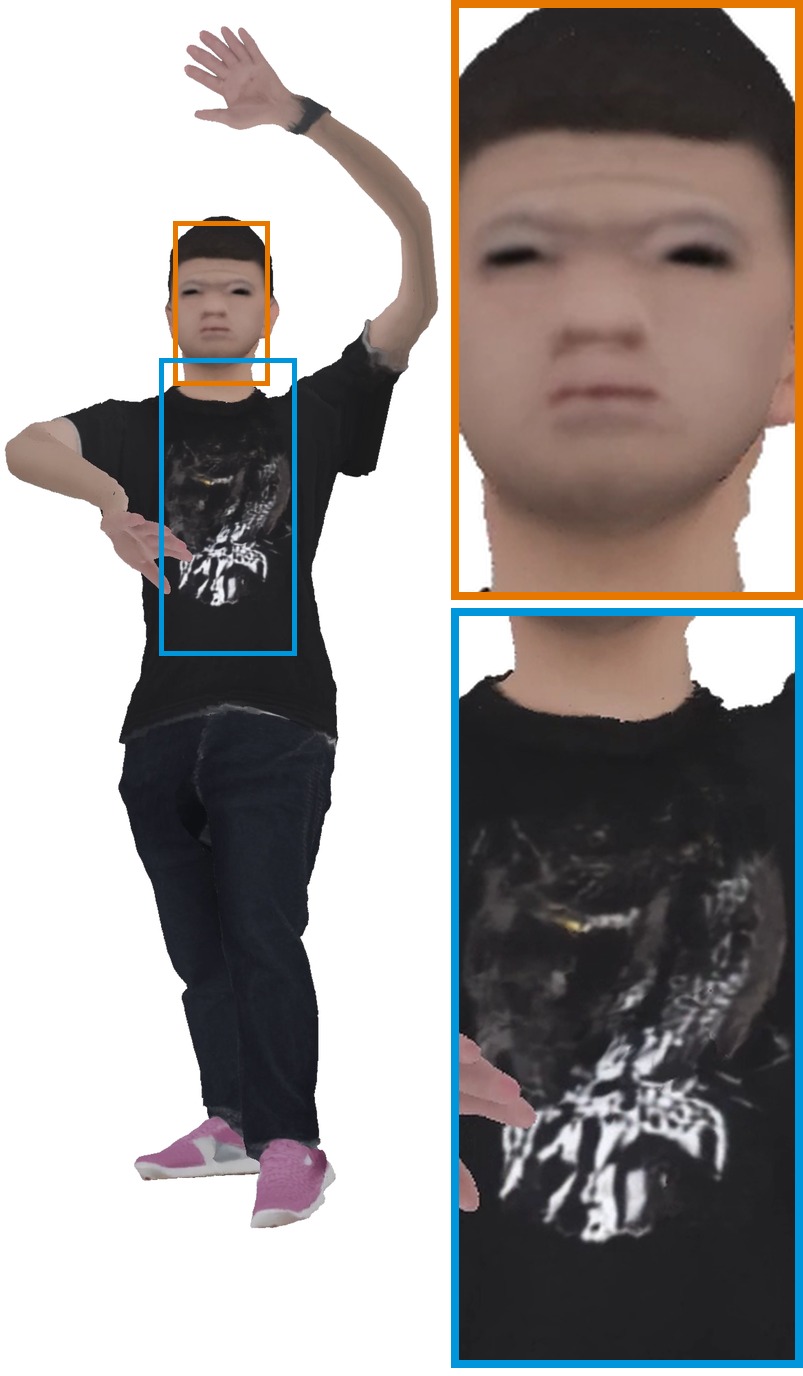} & \compcellT{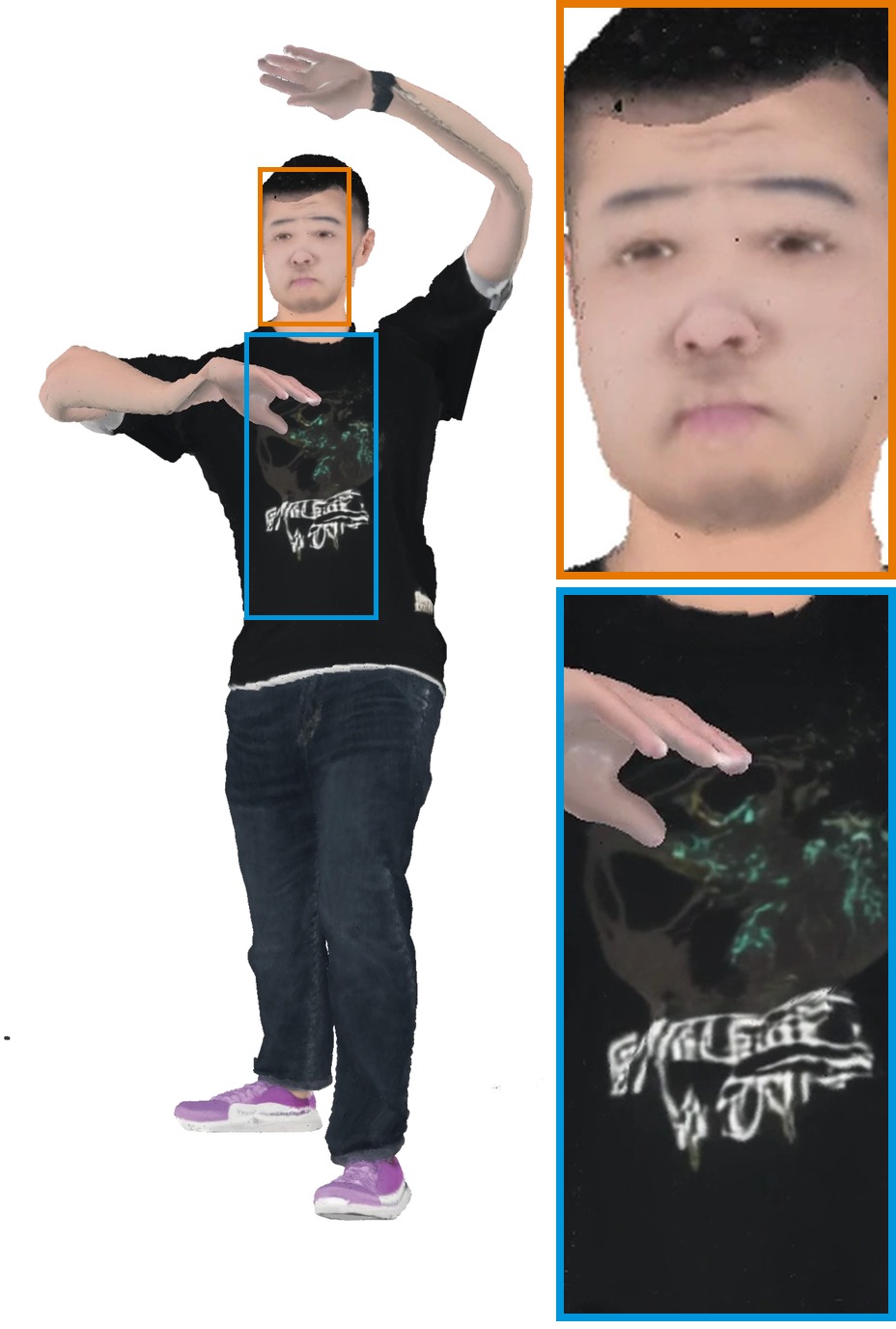} & \compcellT{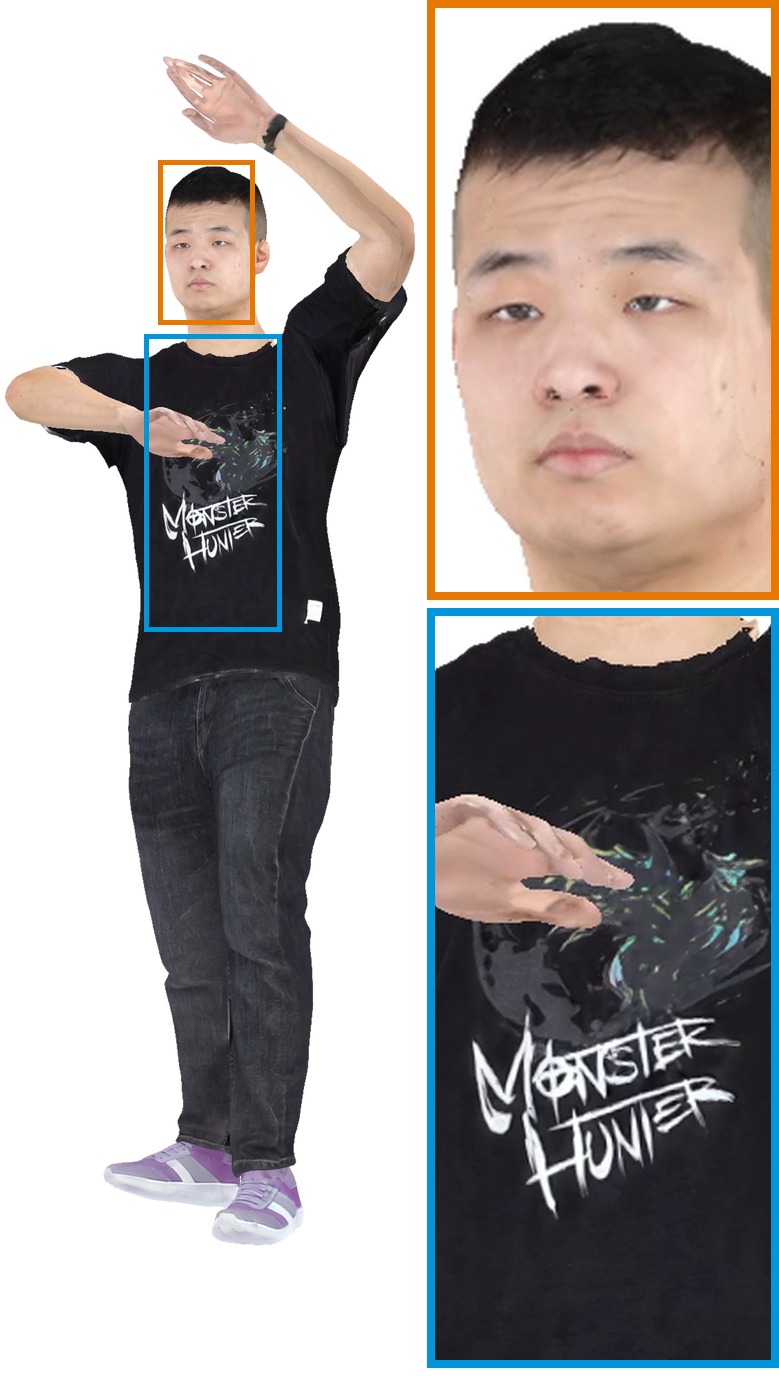} & \compcellT{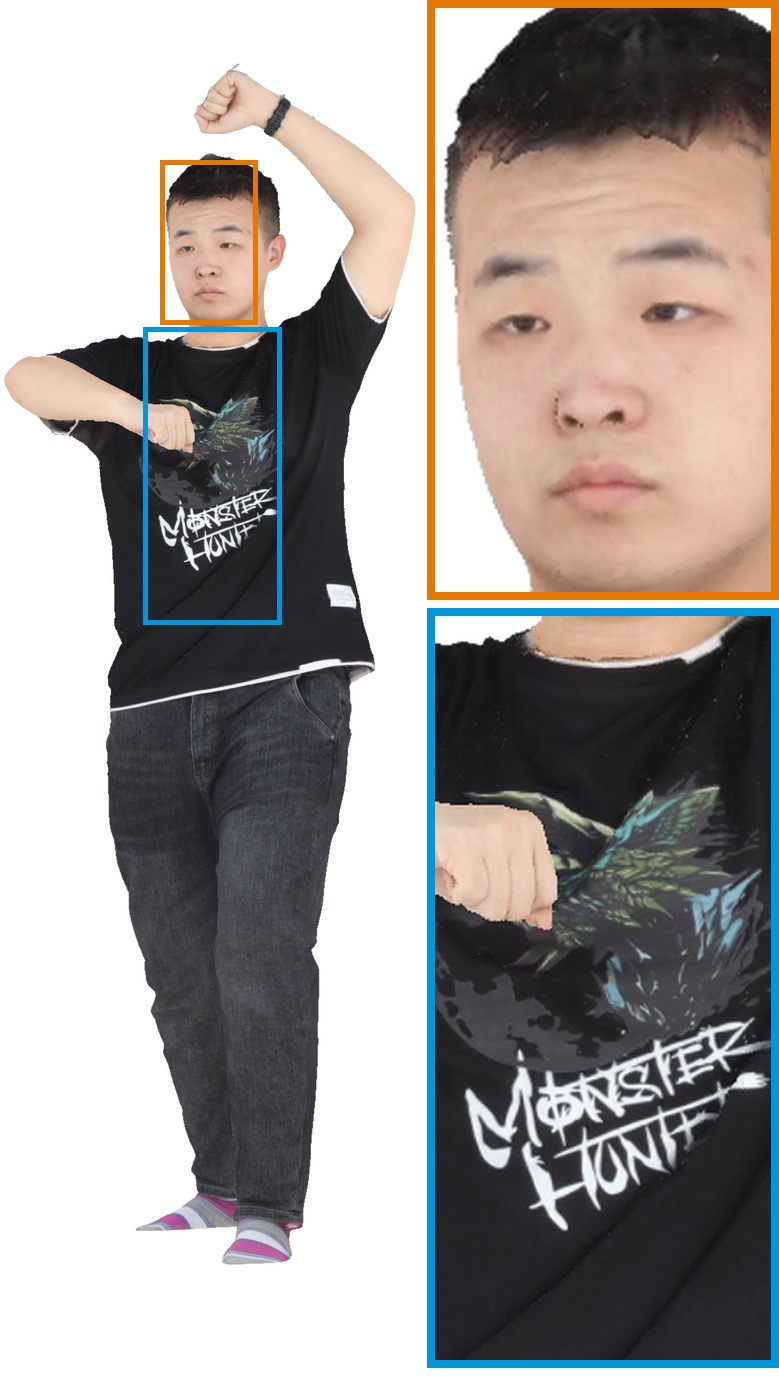} \\
	\end{tabular}
	\caption{\textbf{Comparison on scans with ground truth (Secs.~\ref{subsec:static} and~\ref{subsec:animatable}).} Meshes are aligned to the scan and rendered with one camera per row. \textbf{Static:} an X-Humans subject in the pose of the input image in a new view; zooms show the face (orange) and the left hand (blue), and an inset shows the upper body 30$^\circ$ from the front. \textbf{Animatable:} a THuman2.1 subject reconstructed from the canonical image and reposed to the scan, seen from a new view; zooms show the face (orange) and the shirt logo (blue). In both setups our reconstruction better matches the facial and other details of the GT picture.}
	\label{fig:comparison}
	\vspace{-6mm}
\end{figure*}
\iftablerow\input{tables/tables_row}\fi

\section{Experiments}
\label{sec:exp}

We compare AvatarAnyone with five baselines on static reconstruction (Sec.~\ref{subsec:static}), animatable reconstruction on scans (Sec.~\ref{subsec:animatable}), and YouTube videos (Secs.~\ref{subsec:animatable_youtube} and~\ref{subsec:user_study}).
We then show applications that the baselines lack: virtual try-on (Sec.~\ref{subsec:tryon}) and cloth simulation (Sec.~\ref{subsec:graphics}). We also ablate the main components of our method (Sec.~\ref{subsec:ablation}).

\noindent\textbf{Datasets.}
For static reconstruction, we use 20 X-Humans~\cite{shen2023xavatar} test scans, from 20 participants, with one front view of each as input.
SyncHuman trains on 2K2K~\cite{han2023highfidelity}, CustomHumans~\cite{ho2023customhumans}, and THuman2.1~\cite{yu2021function4d}, so the static setup does not use them.
For animatable reconstruction, we use 100 scans, 25 from each of THuman2.1, CustomHumans, 2K2K, and CAPE~\cite{ma2020cape}.
PSHuman and SyncHuman train on some of them, but the input is a generated canonical image, not a render (Sec.~\ref{subsec:animatable}).
CAPE scans have no texture, so the appearance metrics cover the other 75.
We also evaluate on YouTube videos (Secs.~\ref{subsec:animatable_youtube} and~\ref{subsec:user_study}).

\noindent\textbf{Baselines.}
We compare with three single-image human methods, PSHuman~\cite{li2025pshuman}, SyncHuman~\cite{chen2025synchuman}, and HumanNOVA~\cite{hu2026humannova}, and two general image-to-3D models, TRELLIS.2~\cite{xiang2026trellis2} and Pixal3D~\cite{li2026pixal3d}.
We run their official code and weights, and extract the mesh of HumanNOVA with Marching Cubes~\cite{lorensen1987marching}, as its authors do.
AvatarAnyone itself uses Pixal3D to initialize loose garments, so this comparison shows what our method adds.
We compare only with methods that output meshes, because our goal is avatars that graphics engines, cloth simulators, and robotics simulators use directly.
3DGS avatars such as AHOY~\cite{mir2026ahoy}, LHM~\cite{qiu2025lhm}, and IDOL~\cite{zhuang2025idol} are animatable and target rendering quality, but these tools cannot load them directly, and their garments cannot be simulated or swapped as meshes. These avatars also do not provide a mesh surface and cannot be scored on surface geometry.

\noindent\textbf{Metrics.}
We measure geometry with the Chamfer distance (CD) and the point-to-surface distance (P2S), both in cm, and with the F-score at 1\,cm~\cite{tatarchenko2019single}.
We measure appearance with PSNR in dB.
The animatable setup adds SSIM~\cite{wang2004ssim} and LPIPS~\cite{zhang2018lpips}. 
Following HumanNOVA~\cite{hu2026humannova}, we match the scale of each mesh to the scan and align the two with ICP~\cite{besl1992icp}. We also evaluate in-the-wild reconstruction with a perceptual study (Sec.~\ref{subsec:user_study}).
\subsection{Static reconstruction}
\label{subsec:static}

Single-image methods reconstruct the person in the pose of the input image, so we first compare all methods in this setup (Tab.~\ref{tab:quan_static}, Fig.~\ref{fig:comparison}, top). We use our fitting pipeline directly on the posed image without canonicalization, so all methods receive the same input image.
AvatarAnyone is best on all four metrics.
Against the strongest geometry baseline, Pixal3D, it lowers the Chamfer distance from 1.18 to 1.06\,cm. It raises PSNR by 1.2\,dB over the best appearance baseline, HumanNOVA.
In Tabs.~\ref{tab:quan_static} and~\ref{tab:quan_animatable}, the baselines solve the easier problem of one fused surface.
AvatarAnyone matches or beats them on this task, although it also recovers the garments as separate layers.

\iftablerow\else\input{tables/quan_static}\fi %

\subsection{Animatable reconstruction on scans}
\label{subsec:animatable}

\noindent\textbf{Protocol.}
To simulate a casual video with known geometry, we render a front-facing video of each scan. For CAPE, we only use the available posed renderings.
The image model~\cite{openai2026gptimage25} turns these into a canonical image (Sec.~\ref{subsec:inputs}) that no method has seen in training. All methods reconstruct the 3D human from the exact same image.
We repose each mesh to the SMPL-X fit of the scan with LBS~\cite{smpl2015loper,SMPL-X:2019}.
The baselines have no skeleton, so each of their vertices takes the weights of the nearest vertex of an SMPL-X fit.

\noindent\textbf{Results.}
Over all 100 scans, AvatarAnyone is best on every metric (Tab.~\ref{tab:quan_animatable}, Fig.~\ref{fig:comparison}, bottom).
It lowers the Chamfer distance of the best baseline from 1.92 to 1.36\,cm, by 29\%, against about 10\% in the static setup.
Per dataset, it is best or tied for best on every metric; the ties are on 2K2K, with PSHuman in SSIM and SyncHuman in LPIPS.

\input{tables/quan_animatable}

\subsection{Animatable reconstruction on YouTube videos}
\label{subsec:animatable_youtube}
\begin{figure*}[t]
	\centering
	\vspace*{2.5mm}%
	\begin{overpic}[width=\linewidth]{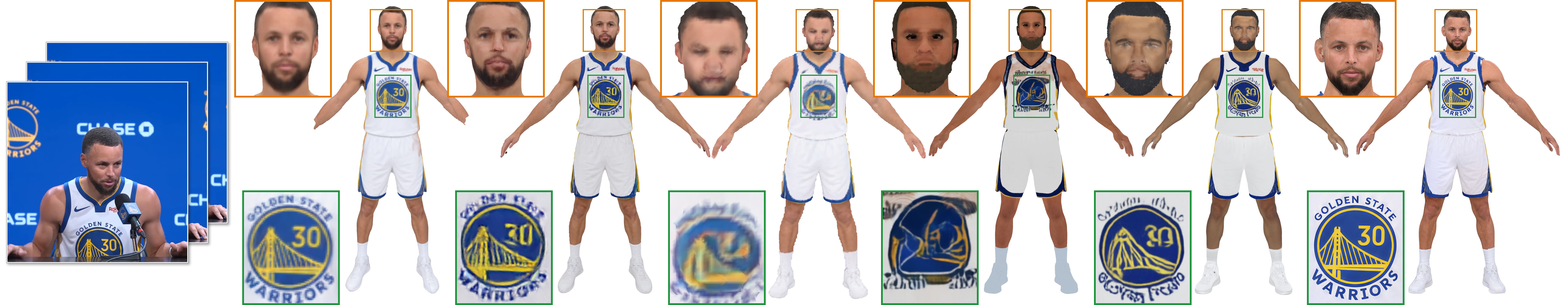}
		\put(7.5,20.4){\makebox(0,0)[b]{\scriptsize\strut Input video}}
		\put(20.6,20.4){\makebox(0,0)[b]{\scriptsize\strut PSHuman}}
		\put(34.2,20.4){\makebox(0,0)[b]{\scriptsize\strut SyncHuman}}
		\put(47.8,20.4){\makebox(0,0)[b]{\scriptsize\strut HumanNOVA}}
		\put(61.4,20.4){\makebox(0,0)[b]{\scriptsize\strut TRELLIS.2}}
		\put(74.9,20.4){\makebox(0,0)[b]{\scriptsize\strut Pixal3D}}
		\put(88.5,20.4){\makebox(0,0)[b]{\scriptsize\strut Ours}}
	\end{overpic}
	\caption{\textbf{Comparison on a YouTube video (Sec.~\ref{subsec:animatable_youtube}).} All methods start from the same canonical image; meshes are aligned to our avatar and rendered from the front with one camera. Zooms show the face (orange) and the jersey logo (green). Of all methods, our method preserves the jersey logo and facial details the best.}
	\label{fig:comparison_youtube}
	\vspace{-5mm}%
\end{figure*}

\begin{figure*}[t]
	\centering
	\vspace*{2.5mm}%
	\begin{overpic}[width=\linewidth]{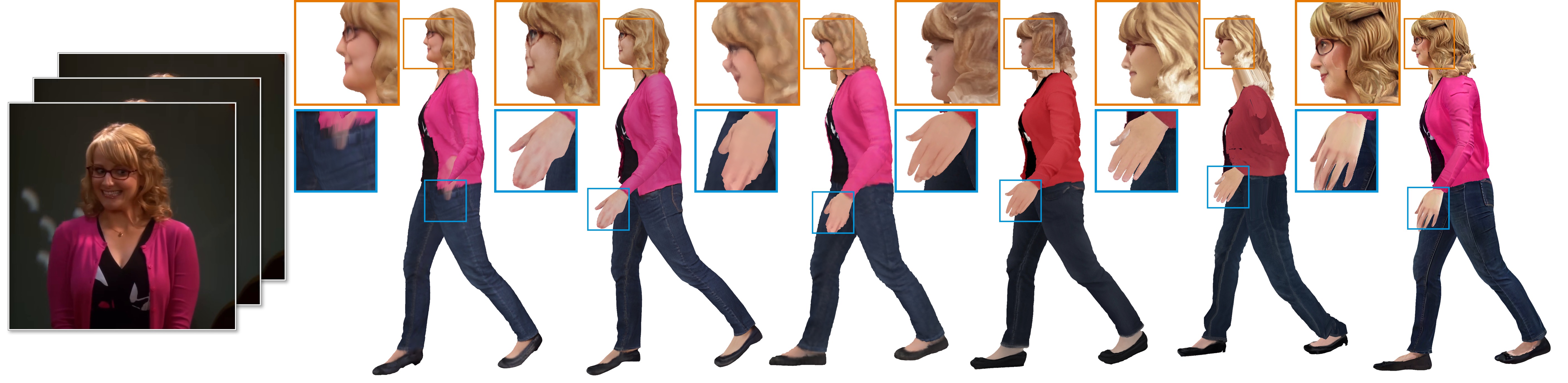}
		\put(9.4,25.3){\makebox(0,0)[b]{\scriptsize\strut Input video}}
		\put(23.9,25.3){\makebox(0,0)[b]{\scriptsize\strut PSHuman}}
		\put(36.6,25.3){\makebox(0,0)[b]{\scriptsize\strut SyncHuman}}
		\put(49.4,25.3){\makebox(0,0)[b]{\scriptsize\strut HumanNOVA}}
		\put(62.2,25.3){\makebox(0,0)[b]{\scriptsize\strut TRELLIS.2}}
		\put(74.9,25.3){\makebox(0,0)[b]{\scriptsize\strut Pixal3D}}
		\put(87.7,25.3){\makebox(0,0)[b]{\scriptsize\strut Ours}}
	\end{overpic}
	\caption{\textbf{Animatable avatars from a second YouTube video (Sec.~\ref{subsec:animatable_youtube}).} All methods start from the same canonical image, and each avatar is reposed to a walking pose (baselines: nearest-vertex rig of Sec.~\ref{subsec:animatable}; ours: its own rig). Zooms show the face (orange) and the hand (blue). Of all methods, our method reconstructs the most visually convincing facial and hand details.}
	\label{fig:comparison_youtube2}
	\vspace{-6mm}
\end{figure*}

We construct an evaluation set of 25 YouTube videos.
These videos occlude parts of the person and have no scans, so we evaluate in image space: the first 5 seconds are used as input for reconstruction and the next 5 for evaluation.
We repose each avatar to the SMPL-X pose of each held-out frame and render it with that frame's camera (both estimated using NLF~\cite{sarandi2024nlf}).
All methods use the same poses and cameras.
AvatarAnyone improves PSNR over all baselines (Tab.~\ref{tab:quan_youtube}, Figs.~\ref{fig:comparison_youtube} and~\ref{fig:comparison_youtube2}).

\iftablerow\else\input{tables/quan_youtube}\fi %

\subsection{Perceptual study on YouTube videos}
\label{subsec:user_study}

The metrics above score only the visible parts, so a perceptual study judges the whole reconstruction of mesh-based methods.
We select 12 videos with occlusions and varied clothing.
Participants see each input video next to textured 360$^\circ$ renders of the six reconstructions, in random order and unnamed, and are asked ``which 3D reconstruction matches the shown YouTube video best?''.
In total, 52 participants each judge 12 videos, which gives 624 answers.
They pick AvatarAnyone in 79.81\% of answers and SyncHuman in 12.50\% (Tab.~\ref{tab:quan_youtube}). See the supplementary material for the interface of the perceptual study.

\subsection{Layered reconstruction for virtual try-on}
\label{subsec:tryon}

\begin{figure*}[t]
	\centering
	\setlength{\tabcolsep}{1pt}
	\ifdefined\tryonh\else\newlength{\tryonh}\fi
	\setlength{\tryonh}{2.88cm}
	\newcommand{\tryonimg}[1]{\raisebox{-0.5\height}{\includegraphics[height=\tryonh]{#1}}}
	\newcommand{\tryonstack}[1]{\raisebox{-0.5\height}{\includegraphics[width=0.8\tryonh]{#1}}}
	\resizebox{\linewidth}{!}{%
	\begin{tabular}{@{}cccc@{\hspace{8pt}}cccc@{}}
		\scriptsize Input video & \scriptsize Avatar & \scriptsize Try-on 1 & \scriptsize Try-on 2 & \scriptsize Input video & \scriptsize Avatar & \scriptsize Try-on 1 & \scriptsize Try-on 2 \\
		\tryonstack{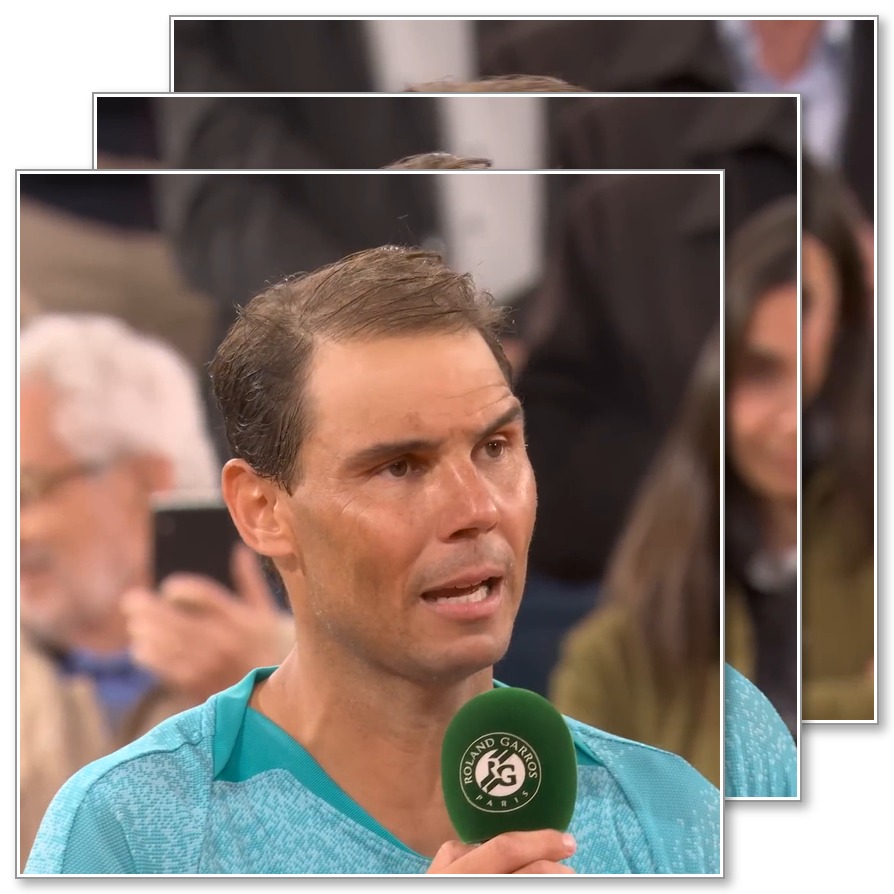} & \tryonimg{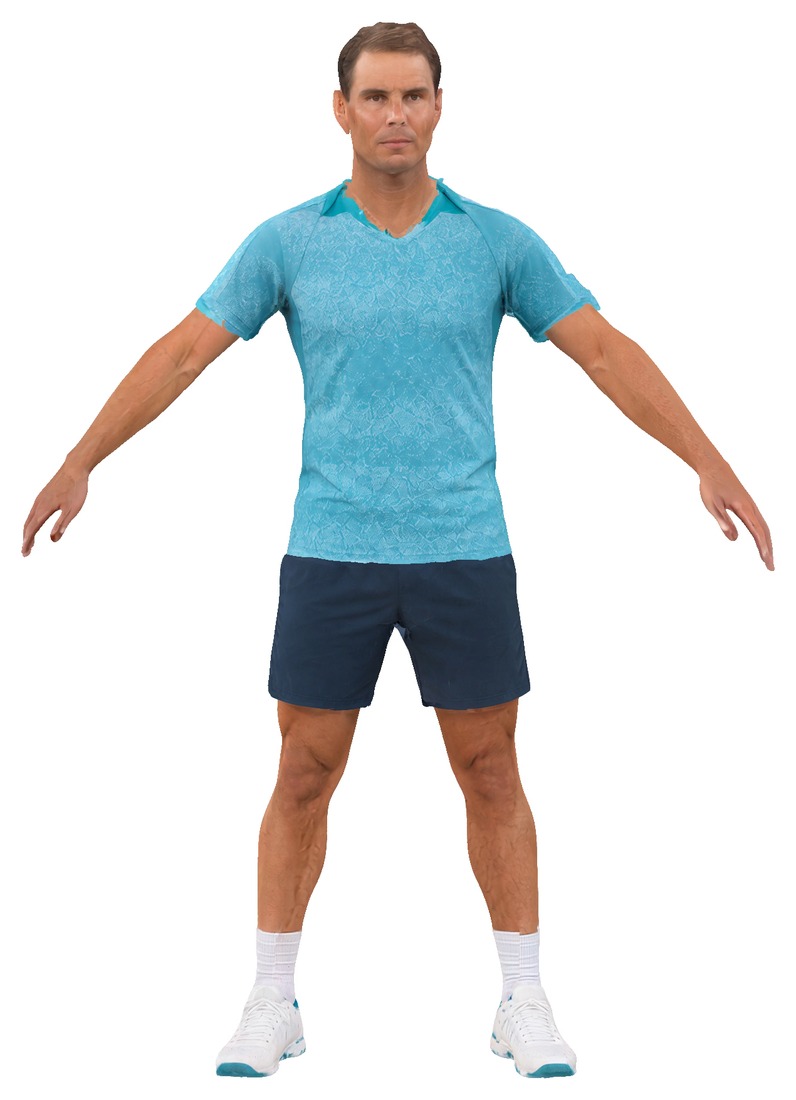} & \tryonimg{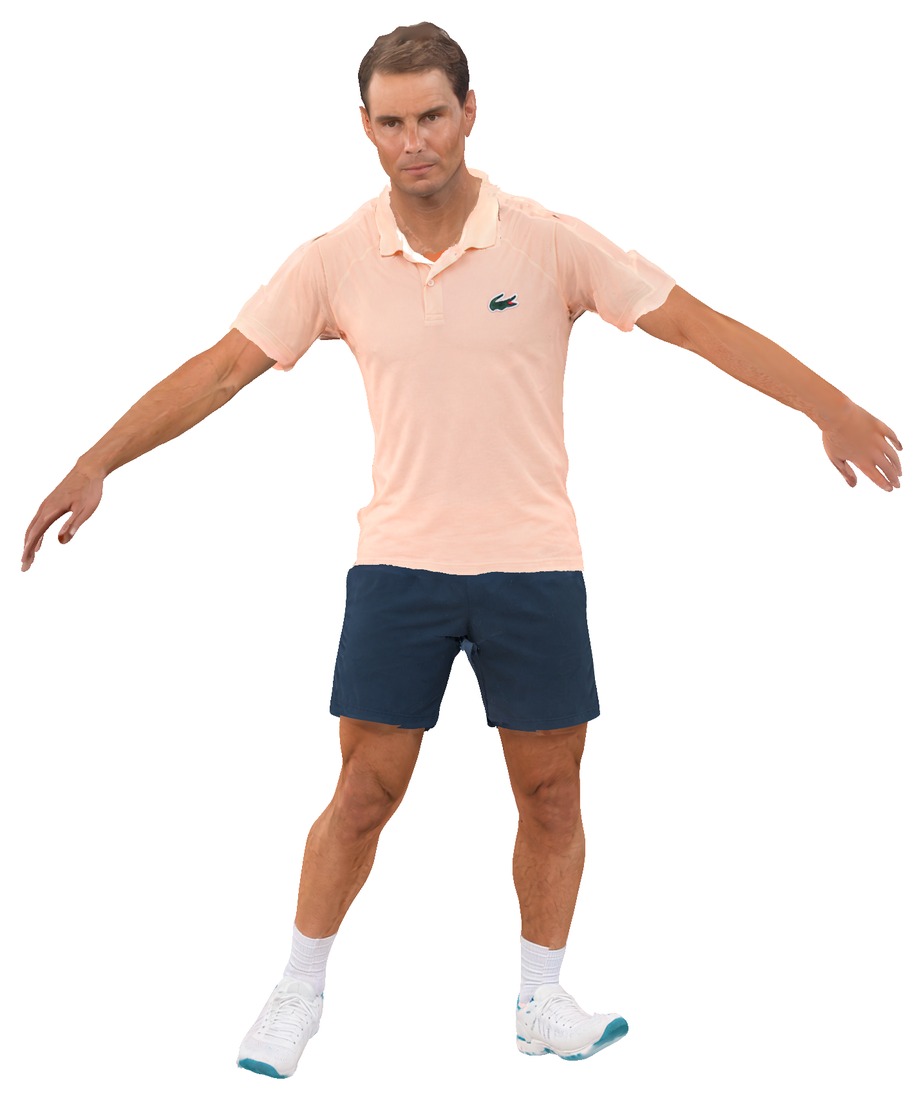} & \tryonimg{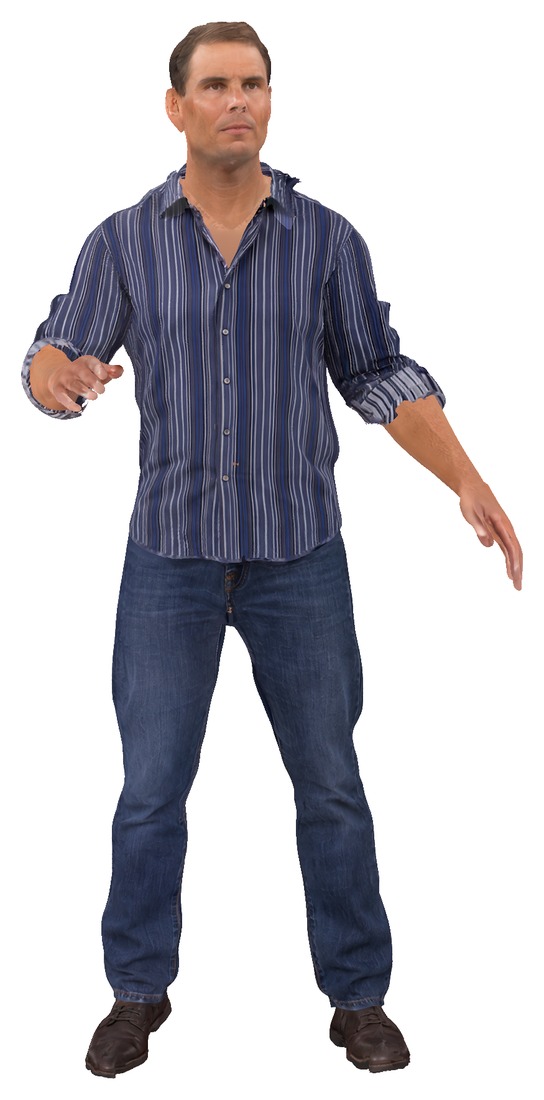} & \tryonstack{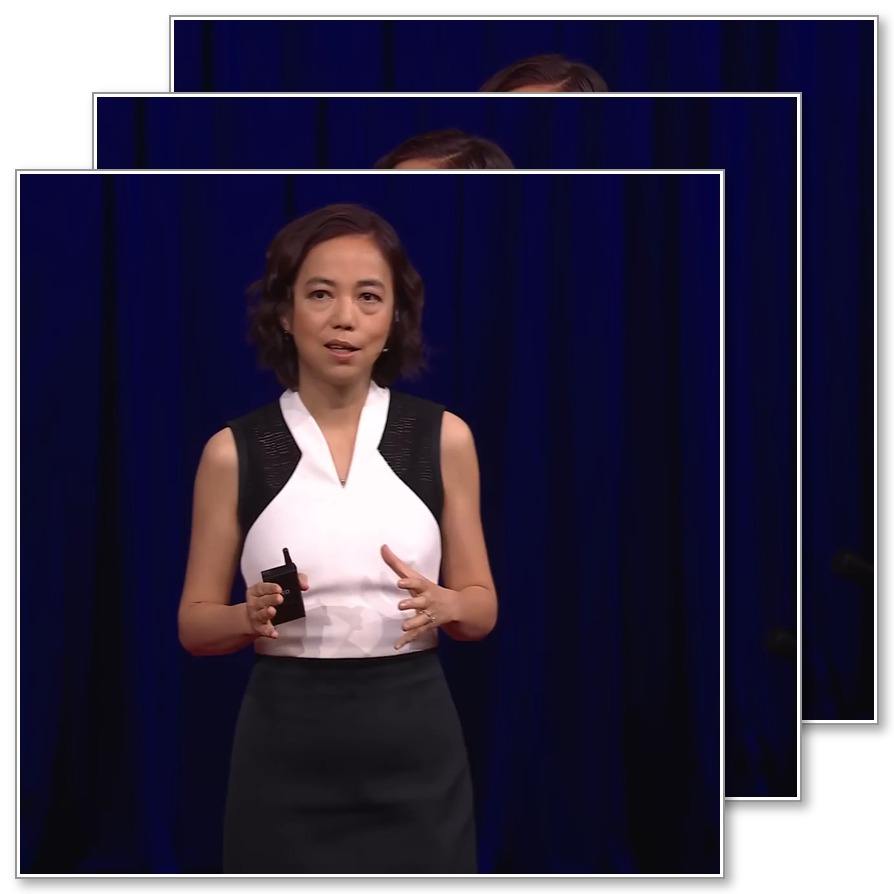} & \tryonimg{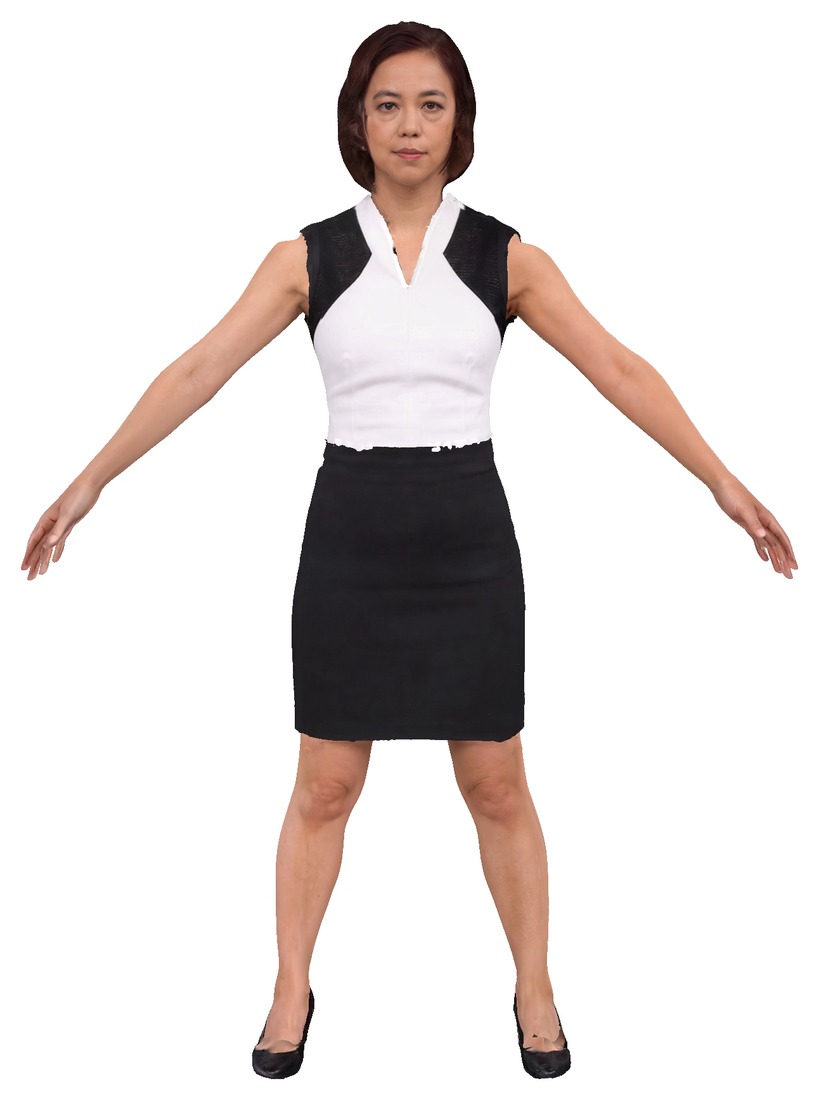} & \tryonimg{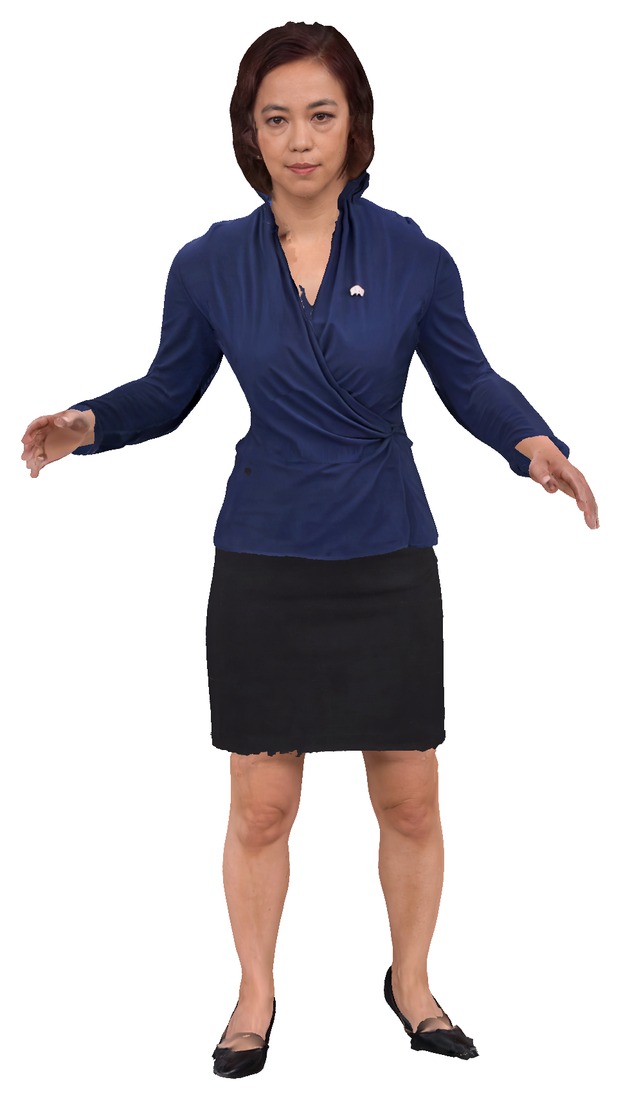} & \tryonimg{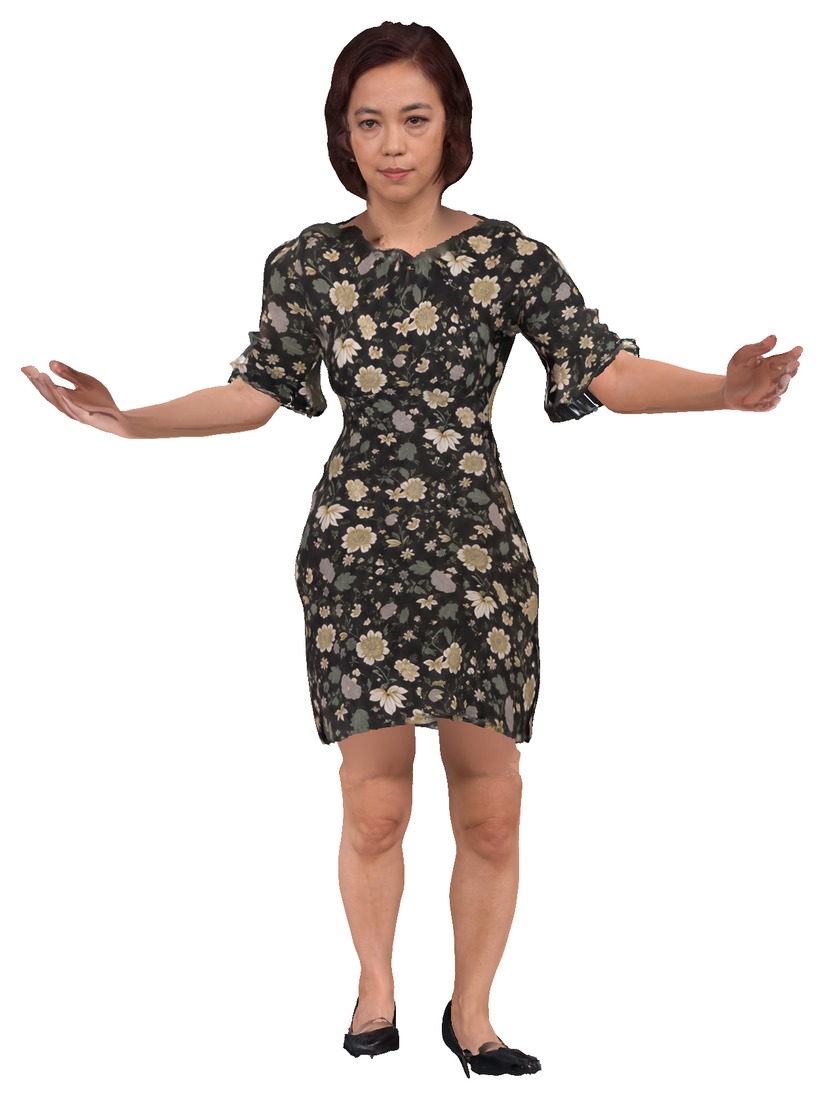} \\
		\tryonstack{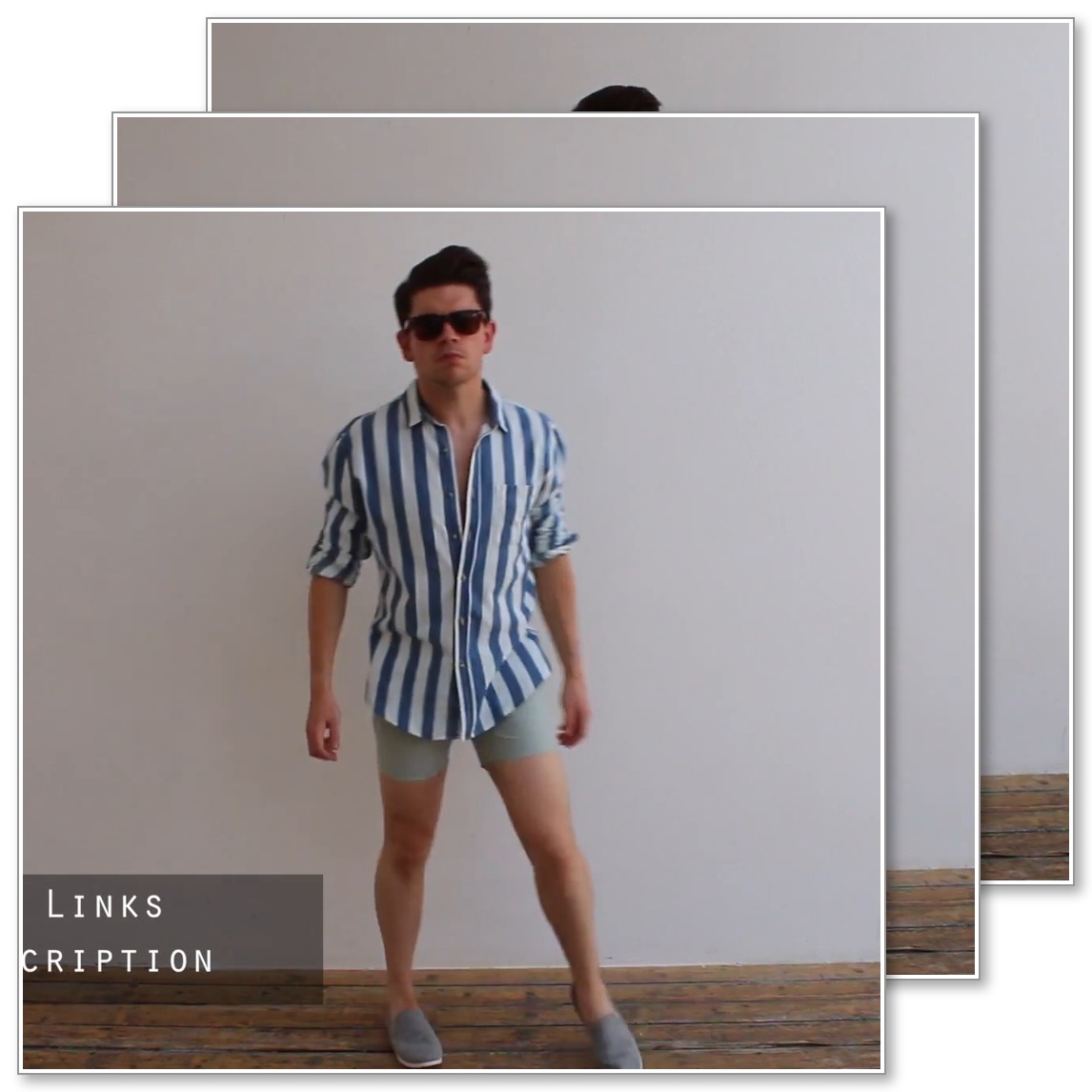} & \tryonimg{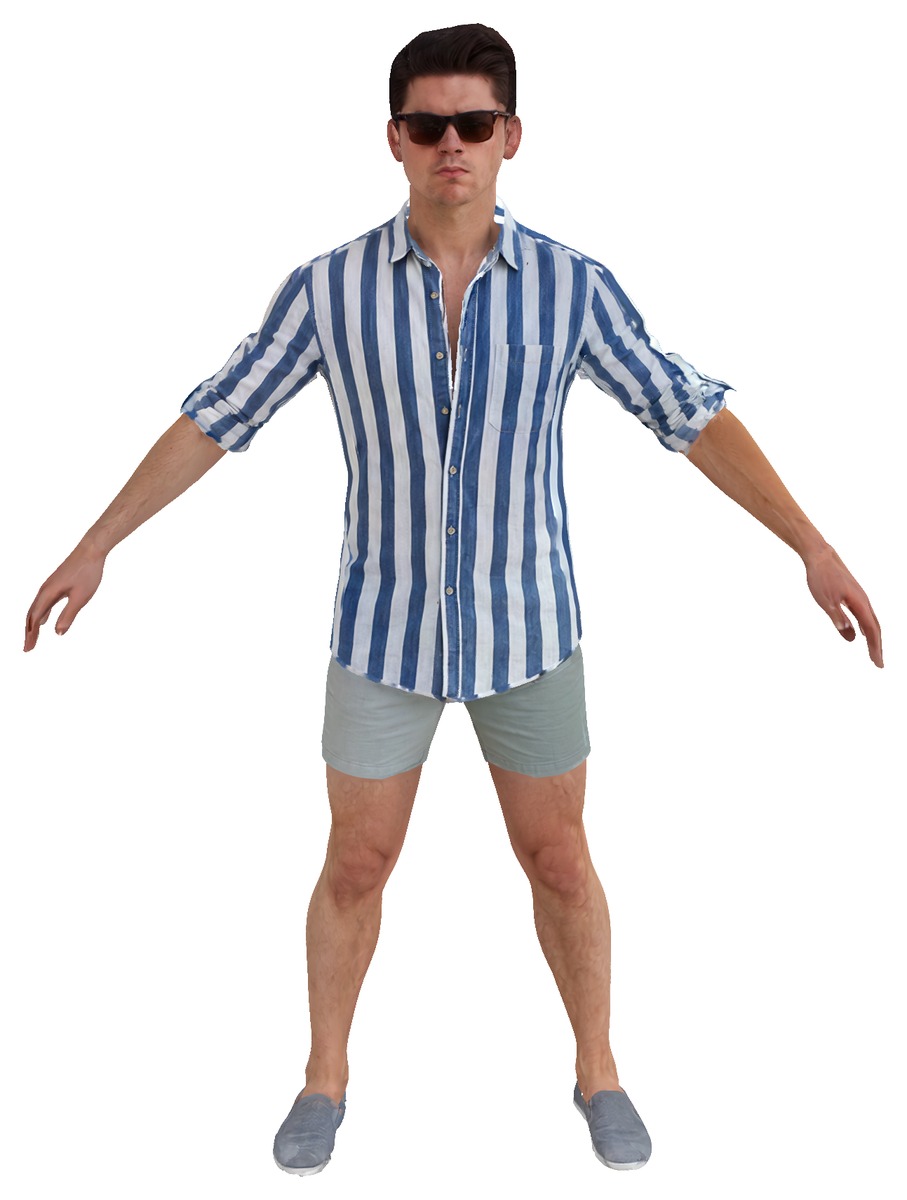} & \tryonimg{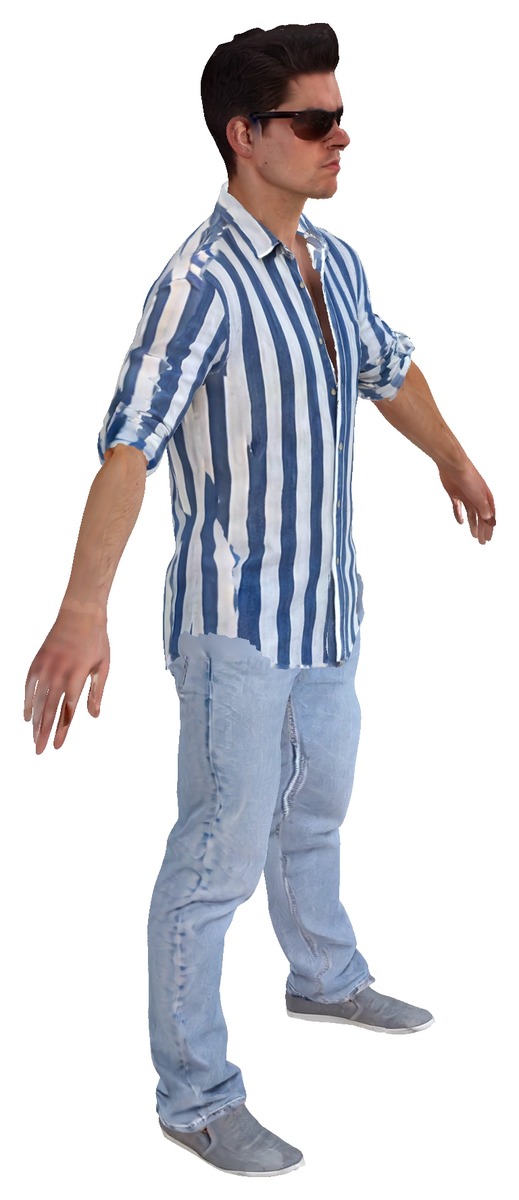} & \tryonimg{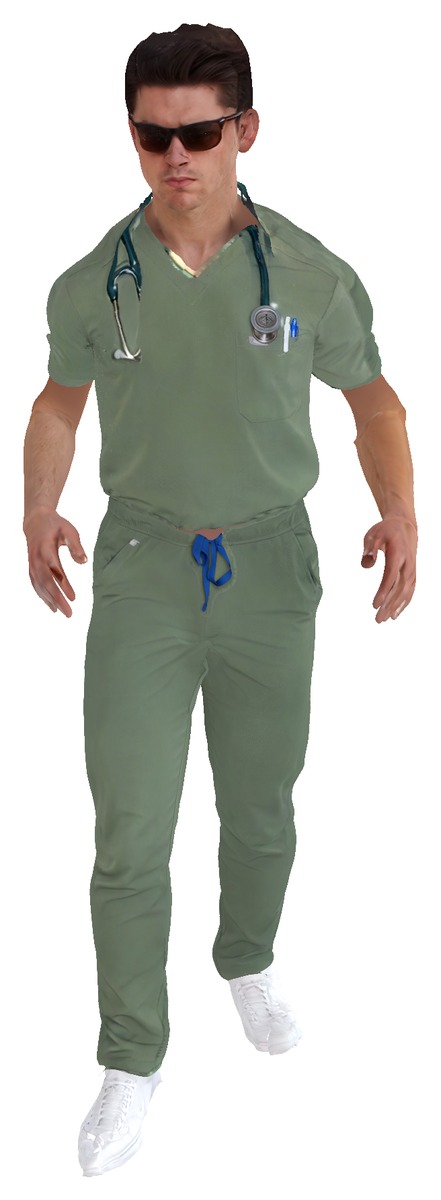} & \tryonstack{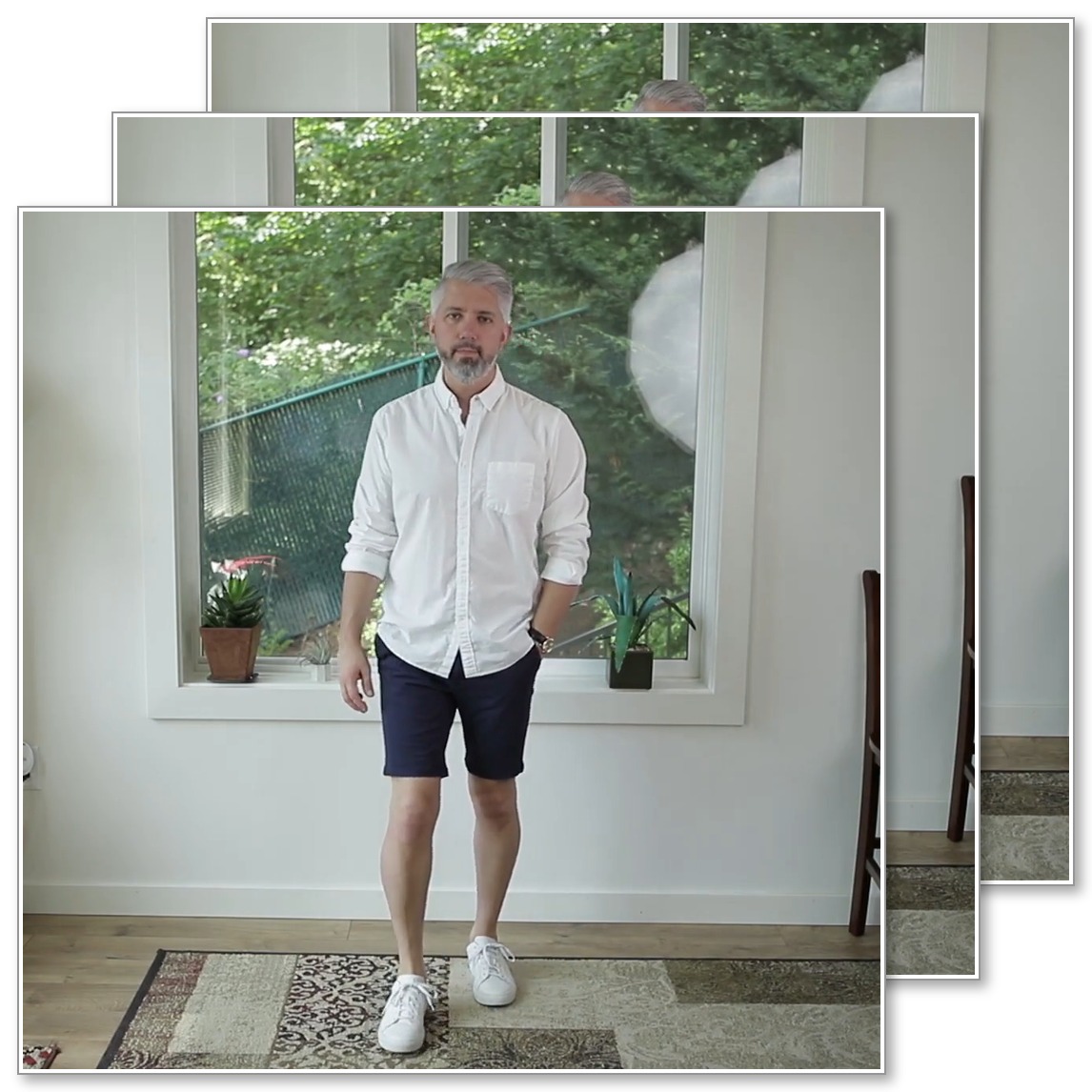} & \tryonimg{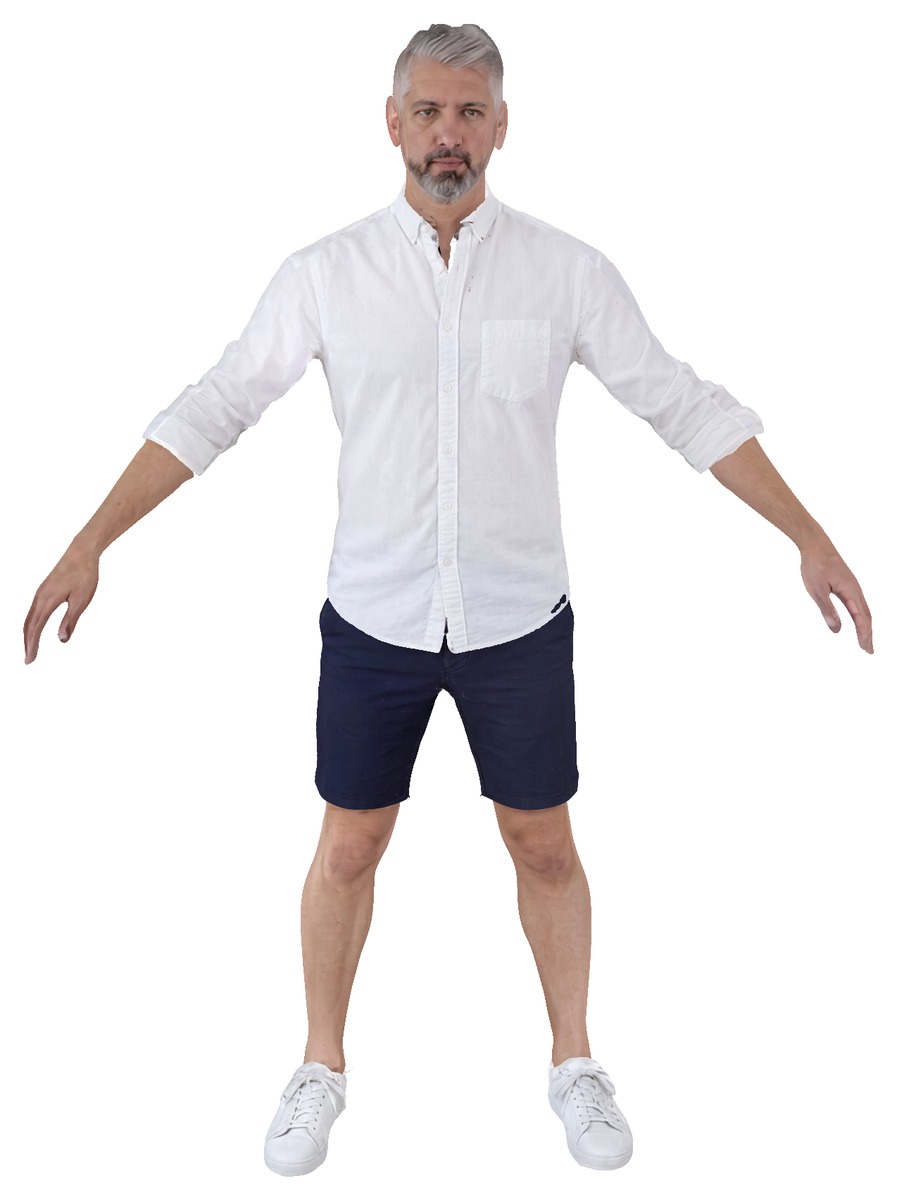} & \tryonimg{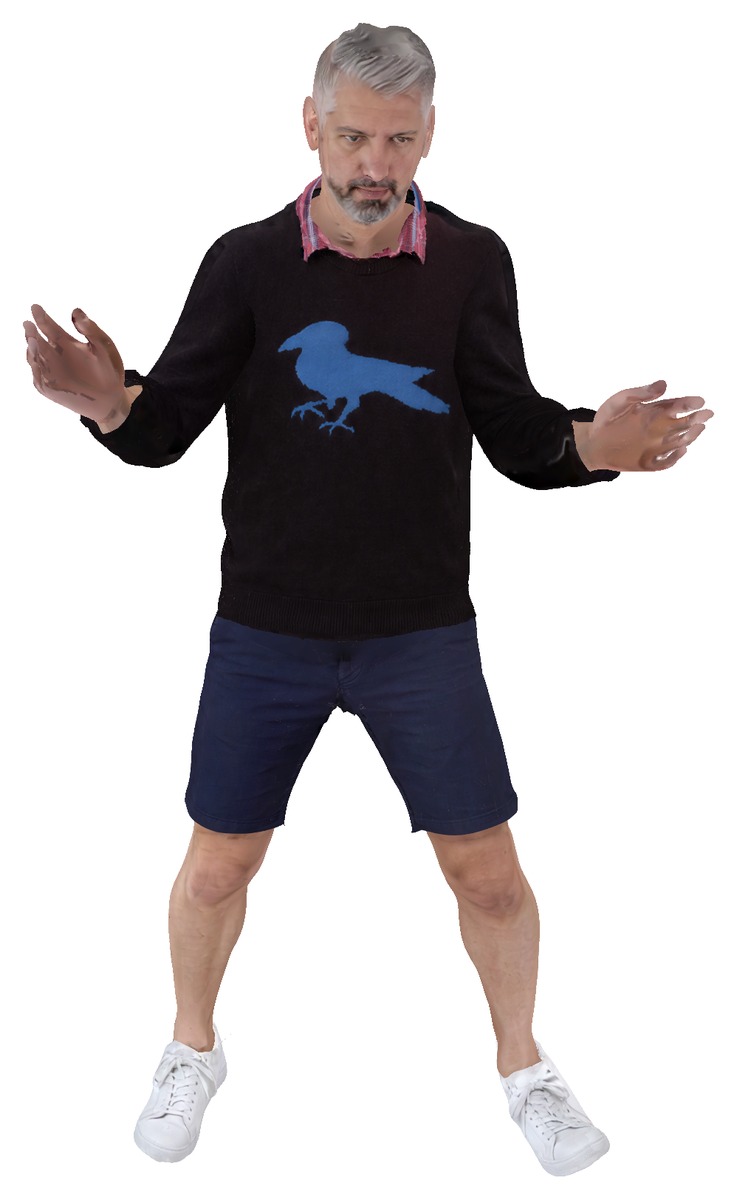} & \tryonimg{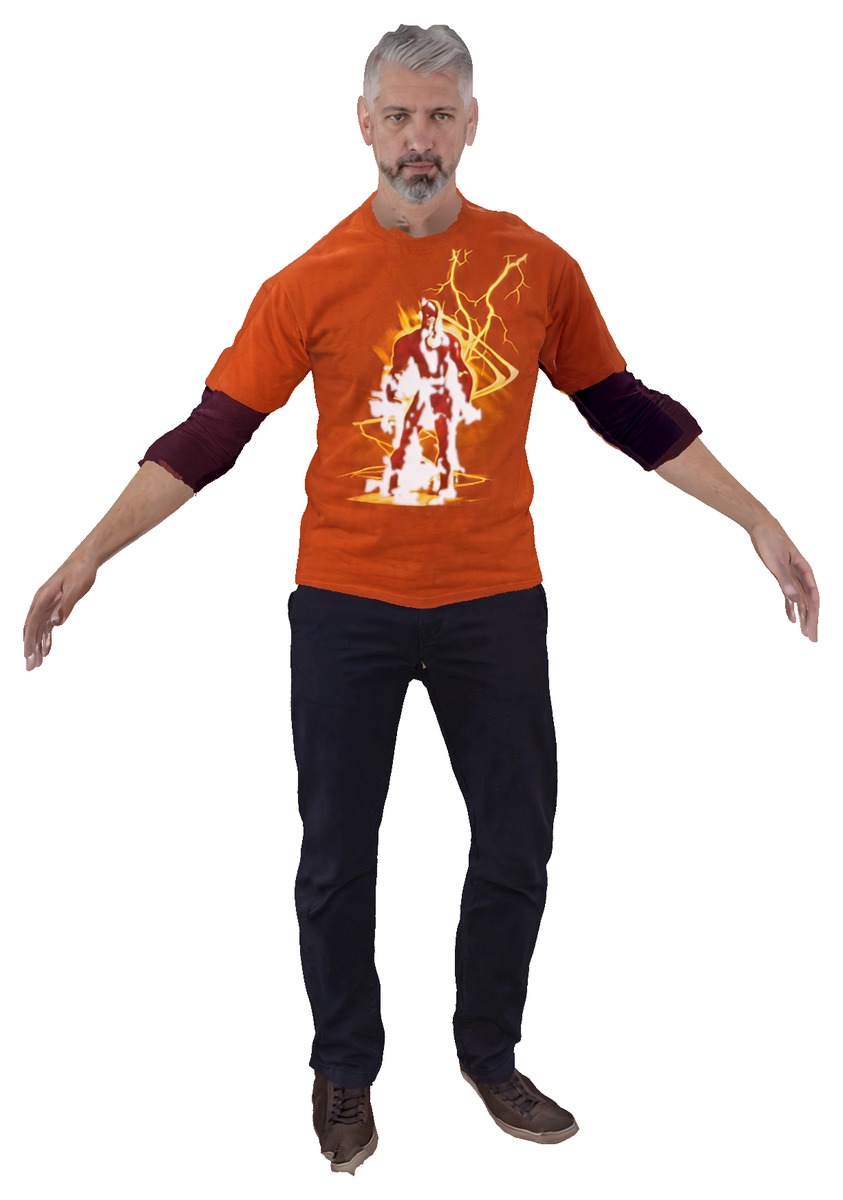} \\
	\end{tabular}%
	}
	\caption{\textbf{Virtual try-on with layered avatars (Sec.~\ref{subsec:tryon}).} As our avatars are layered meshes, the clothing can be replaced. Each row shows two people: input frames, our avatar, and two try-ons that replace garment layers with garments of other avatars and repose the avatar. }
	\label{fig:tryon}
	\vspace{-6mm}
\end{figure*}

In our reconstructions, each layer is a separate mesh with its own texture (Sec.~\ref{sec:method}).
We can therefore replace the top, the lower garment, or the shoes of an avatar with those of another avatar (Fig.~\ref{fig:tryon}).
The baselines output one fused surface and do not support this application, because removing a garment would leave a hole without a body.

\begin{figure*}[t]
	\centering
	\includegraphics[width=\linewidth]{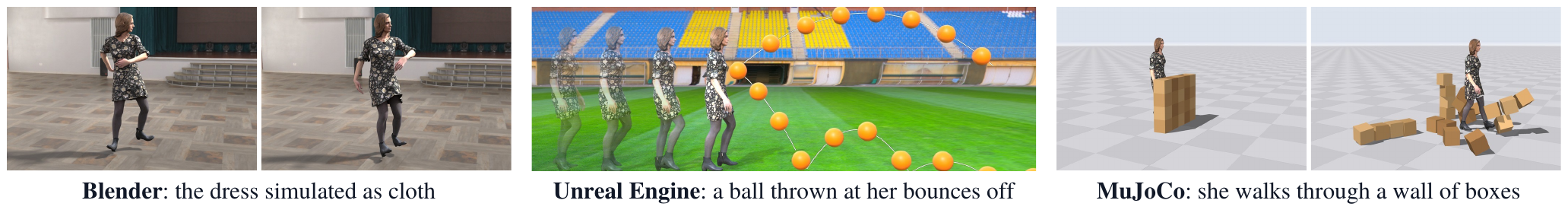}
	\caption{\textbf{One avatar in three tools (Sec.~\ref{subsec:graphics}).} \textbf{Blender:} two moments of a dance with the dress simulated as cloth, rendered with Cycles. \textbf{Unreal Engine:} a ball thrown at the walking avatar bounces off its body; the picture overlays the ball at regular time steps, its path, and four moments of the walk. \textbf{MuJoCo:} the avatar follows a motion-capture walk through a wall of 20 boxes (left: before, right: after).}
	\label{fig:engines}
	\vspace{-4mm}
\end{figure*}

\begin{figure*}[!t]
	\centering
	\includegraphics[width=\linewidth]{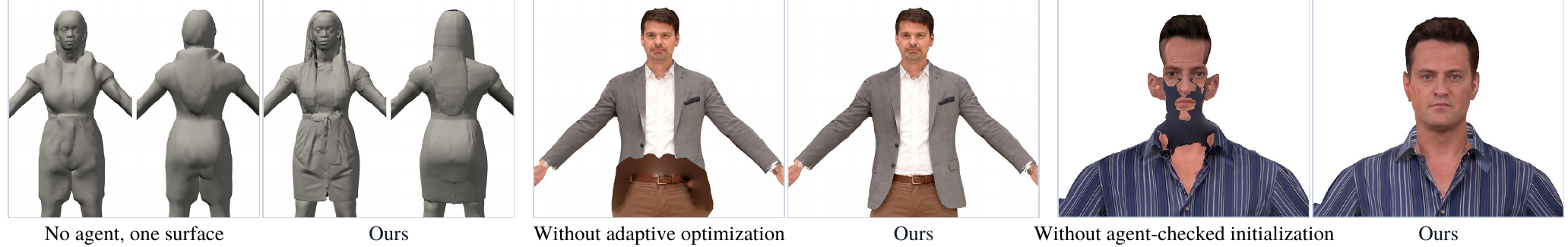}
	\caption{\textbf{Ablation study (Sec.~\ref{subsec:ablation}).} \textbf{Left:} geometry without any agents, using only SMPL-X optimization. \textbf{Middle:} without adaptive optimization and the layer order from the agent, the trousers are placed outside the jacket. \textbf{Right:} without the agent-checked initialization, the generated upper garment interferes with the optimization of the head and yields incorrect results.}
	\label{fig:ablation}
    \vspace{-6mm}

\end{figure*}

\subsection{Clothing, contact and physics simulation}
\label{subsec:graphics}
\label{subsec:qualitative}

Blender~\cite{blender2026blender} and Unreal Engine~\cite{epic2026unreal} import our avatars directly and animate them with their own tools (Fig.~\ref{fig:engines}).
In Blender, the cloth simulator moves the skirt of the dress over the dancing body.
In Unreal Engine, the ball bounces off collision bodies that follow the skeleton.
In MuJoCo~\cite{todorov2012mujoco}, the physics engine moves the boxes on contact with the walking avatar.
The baselines output one static surface that needs rigging and has no separate garments, so cloth simulation is impossible.
3DGS avatars do not load natively into these applications.

\subsection{Ablation studies}
\label{subsec:ablation}
We remove one component at a time and evaluate on 24 of the 100 scans, 6 from each dataset (Tab.~\ref{tab:ablation} and Fig.~\ref{fig:ablation}).
\emph{w/o agents and with one surface} fits SMPL-X to the whole person: the hair becomes a hood, and the dress closes between the legs.
\emph{w/o adaptive optimization} gives everyone the same layers and layer order, so trousers can cover a jacket.
\emph{w/o agent-checked initialization} skips the agent gate, so a shirt initialization with a head moves onto the face.
The full method is best on every metric.

\iftablerow\else\input{tables/ablation}\fi %

%% file: tables/tables_row.tex
\begin{table*}[t]
	\captionsetup{justification=raggedright,singlelinecheck=false}%
	\input{tables/quan_static}\hfill
	\input{tables/quan_youtube}\hfill
	\input{tables/ablation}
	\vspace{-3mm}
\end{table*}

%% file: tables/quan_static.tex
\begin{rowtable}{0.29\linewidth}
	\centering
	\rowtablefont
	\renewcommand{\arraystretch}{1.15}
	\caption{\textbf{Static reconstruction on X-Humans (Sec.~\ref{subsec:static}).} Green: best; yellow: second best.}
	\label{tab:quan_static}
	\vspace{-2mm}%
	\begin{tabular}{l|ccc|c}
		\toprule
		\rowcolor{headerbg}
		\textbf{Method} & \textbf{CD}\,$\downarrow$ & \textbf{P2S}\,$\downarrow$ & \textbf{F-score}\,$\uparrow$ & \textbf{PSNR}\,$\uparrow$ \\
		\midrule
		PSHuman~\cite{li2025pshuman} & 1.56 & 1.73 & 0.527 & 20.95 \\
		SyncHuman~\cite{chen2025synchuman} & 1.38 & 1.42 & 0.521 & 20.77 \\
		HumanNOVA~\cite{hu2026humannova} & 1.26 & 1.24 & 0.579 & \snd{21.64} \\
		TRELLIS.2~\cite{xiang2026trellis2} & 1.38 & 1.32 & 0.543 & 19.75 \\
		Pixal3D~\cite{li2026pixal3d} & \snd{1.18} & \snd{1.13} & \snd{0.609} & 21.59 \\
		\midrule
		\rowcolor{best} \textbf{Ours} & \fst{1.06} & \fst{1.01} & \fst{0.663} & \fst{22.84} \\
		\bottomrule
	\end{tabular}
\end{rowtable}

%% file: tables/quan_youtube.tex
\begin{rowtable}{0.335\linewidth}
	\centering
	\rowtablefont
	\renewcommand{\arraystretch}{1.15}
	\caption{\textbf{Comparison on YouTube videos.} Reposed renders on held-out frames (Sec.~\ref{subsec:animatable_youtube}) and the votes of the perceptual study in \% (Sec.~\ref{subsec:user_study}).}
	\label{tab:quan_youtube}
	\vspace{-2mm}%
	\begin{tabular}{l|cc|c}
		\toprule
		\rowcolor{headerbg}
		& \multicolumn{2}{c|}{\textbf{Reposed renders}} & \textbf{Perceptual study} \\
		\rowcolor{headerbg}
		\textbf{Method} & \textbf{PSNR}\,$\uparrow$ & \textbf{LPIPS}\,$\downarrow$ & \textbf{Votes}\,$\uparrow$ \\
		\midrule
		PSHuman~\cite{li2025pshuman} & 15.20 & 0.206 & 7.05 \\
		SyncHuman~\cite{chen2025synchuman} & 15.19 & \snd{0.202} & \snd{12.50} \\
		HumanNOVA~\cite{hu2026humannova} & 15.41 & 0.217 & 0.00 \\
		TRELLIS.2~\cite{xiang2026trellis2} & 15.34 & 0.225 & 0.16 \\
		Pixal3D~\cite{li2026pixal3d} & \snd{15.50} & 0.211 & 0.48 \\
		\midrule
		\rowcolor{best} \textbf{Ours} & \fst{15.64} & \fst{0.193} & \fst{79.81} \\
		\bottomrule
	\end{tabular}
\end{rowtable}

%% file: tables/ablation.tex
\begin{rowtable}{0.35\linewidth}
	\centering
	\rowtablefont
	\renewcommand{\arraystretch}{1.15}
	\caption{\textbf{Ablation study (Sec.~\ref{subsec:ablation}).} Each variant removes one component. The evaluation uses 24 scans, 6 from each dataset of Tab.~\ref{tab:quan_animatable}; the appearance metrics exclude the untextured CAPE scans. }
	\label{tab:ablation}
	\vspace{-2mm}%
	\begin{tabular}{l|c|ccc}
		\toprule
		\rowcolor{headerbg}
		\textbf{Variant} & \textbf{CD}\,$\downarrow$ & \textbf{PSNR}\,$\uparrow$ & \textbf{SSIM}\,$\uparrow$ & \textbf{LPIPS}\,$\downarrow$ \\
		\midrule
		\rowcolor{best} \textbf{Full} & \fst{1.40} & \fst{19.01} & \fst{0.901} & \fst{0.090} \\
		\midrule
		w/o agents and with one surface & 1.99 & 17.31 & 0.840 & 0.115 \\
		w/o adaptive optimization & 1.80 & 18.12 & 0.881 & \snd{0.102} \\
		w/o agent-checked initialization & \snd{1.76} & \snd{18.42} & \snd{0.892} & 0.105 \\
		\bottomrule
	\end{tabular}
\end{rowtable}

%% file: tables/quan_animatable.tex
\begin{table*}[t]
	\centering
	\fontsize{6.6}{8}\selectfont
	\setlength{\tabcolsep}{1.3pt}
	\renewcommand{\arraystretch}{1.15}
	\caption{\textbf{Animatable reconstruction on four scan datasets (Sec.~\ref{subsec:animatable}).} Columns: all 100 scans (All), THuman2.1~\cite{yu2021function4d} (TH), CustomHumans~\cite{ho2023customhumans} (CH), 2K2K~\cite{han2023highfidelity} (2K), and CAPE~\cite{ma2020cape} (CA), 25 scans each; the appearance metrics of All exclude the untextured CAPE scans. Green: best; yellow: second best.}
	\label{tab:quan_animatable}
	\vspace{-2mm}%
	\begin{tabular}{l|ccccc|ccccc|ccccc|cccc|cccc|cccc}
		\toprule
		\rowcolor{headerbg}
		& \multicolumn{5}{c|}{\textbf{CD}\,$\downarrow$} & \multicolumn{5}{c|}{\textbf{P2S}\,$\downarrow$} & \multicolumn{5}{c|}{\textbf{F-score}\,$\uparrow$} & \multicolumn{4}{c|}{\textbf{PSNR}\,$\uparrow$} & \multicolumn{4}{c|}{\textbf{SSIM}\,$\uparrow$} & \multicolumn{4}{c}{\textbf{LPIPS}\,$\downarrow$} \\
		\rowcolor{headerbg}
		\textbf{Method} & All & TH & CH & 2K & CA & All & TH & CH & 2K & CA & All & TH & CH & 2K & CA & All & TH & CH & 2K & All & TH & CH & 2K & All & TH & CH & 2K \\
		\midrule
		PSHuman & \snd{1.92} & \snd{2.01} & 1.83 & \snd{1.69} & 2.14 & 2.20 & 2.28 & 2.17 & 1.85 & 2.48 & 0.393 & 0.348 & 0.440 & 0.395 & 0.388 & \snd{18.24} & 17.93 & 18.16 & \snd{18.63} & 0.895 & 0.888 & 0.889 & \fst{0.909} & 0.103 & 0.115 & 0.109 & \snd{0.085} \\
		SyncHuman & 1.94 & 2.11 & \snd{1.73} & 1.78 & 2.15 & 2.08 & 2.33 & 1.84 & 1.92 & 2.24 & 0.365 & 0.319 & 0.413 & 0.378 & 0.348 & 18.15 & 17.60 & 18.23 & \snd{18.63} & \snd{0.896} & 0.889 & 0.891 & \snd{0.908} & \snd{0.099} & \snd{0.113} & \snd{0.099} & \fst{0.084} \\
		HumanNOVA & 1.93 & 2.03 & 1.93 & 1.80 & 1.97 & \snd{1.90} & \snd{2.07} & 1.92 & \snd{1.72} & 1.87 & 0.415 & \snd{0.383} & 0.427 & 0.414 & \snd{0.435} & 18.08 & \snd{17.97} & 18.07 & 18.19 & 0.894 & \snd{0.890} & 0.886 & 0.905 & 0.112 & 0.120 & 0.117 & 0.098 \\
		TRELLIS.2 & 2.10 & 2.63 & 1.81 & 2.11 & \snd{1.84} & 1.95 & 2.46 & 1.76 & 1.83 & \snd{1.73} & 0.391 & 0.321 & 0.438 & 0.380 & 0.425 & 17.21 & 16.34 & 18.10 & 17.18 & 0.886 & 0.872 & \snd{0.892} & 0.894 & 0.122 & 0.143 & 0.112 & 0.112 \\
		Pixal3D & 2.06 & 2.35 & 1.77 & 1.91 & 2.19 & 1.91 & 2.28 & \snd{1.58} & 1.81 & 1.97 & \snd{0.417} & 0.355 & \snd{0.470} & \snd{0.421} & 0.422 & 17.93 & 17.20 & \snd{18.37} & 18.22 & 0.889 & 0.878 & 0.887 & 0.903 & 0.117 & 0.134 & 0.113 & 0.103 \\
		\midrule
		\rowcolor{best} \textbf{Ours} & \fst{1.36} & \fst{1.67} & \fst{1.05} & \fst{1.48} & \fst{1.25} & \fst{1.37} & \fst{1.72} & \fst{1.03} & \fst{1.49} & \fst{1.22} & \fst{0.511} & \fst{0.400} & \fst{0.615} & \fst{0.477} & \fst{0.550} & \fst{19.09} & \fst{18.39} & \fst{19.79} & \fst{19.10} & \fst{0.900} & \fst{0.892} & \fst{0.900} & \fst{0.909} & \fst{0.087} & \fst{0.100} & \fst{0.078} & \fst{0.084} \\
		\bottomrule
	\end{tabular}
	\vspace{-5mm}%
\end{table*}

%% file: sec/5_conc.tex
\section{Conclusion}
\label{sec:conc}

AvatarAnyone reconstructs a layered, animatable mesh avatar from a casual video, with the body, the hair, and garments as separate layers.
Its key idea is to treat the structure of a person as a set of visual questions.
Agents answer these questions and select the initialization and the terms of each layer.
One joint optimization then fits the geometry.
On 100 scans, it lowers the Chamfer distance of the best baseline from 1.92 to 1.36\,cm, and it receives 79.81\% of the votes in a perceptual study.
The layers support virtual try-on and cloth simulation. Baselines that reconstruct humans as one mesh cannot support such applications. The avatars also load natively into Unreal Engine and MuJoCo. 

\noindent\textbf{Limitations.}
The avatar has one upper and one lower garment layer, so a jacket and the shirt under it form one layer.
The method depends on proprietary image and language models.
One avatar costs about \$3 in API calls. The fit takes about 2\,h on one GPU.
Avatars of real people can be misused, and the supplementary material discusses this risk.

\noindent\textbf{Future work.}
We believe our work opens a new area of research at the intersection of agentic reconstruction and 3D humans. We hope
future work can shorten the optimization and reconstruct every clothing layer, such as a jacket over a shirt.

%% file: main.bbl
\begin{thebibliography}{171}
\providecommand{\natexlab}[1]{#1}
\providecommand{\url}[1]{\texttt{#1}}
\expandafter\ifx\csname urlstyle\endcsname\relax
  \providecommand{\doi}[1]{doi: #1}\else
  \providecommand{\doi}{doi: \begingroup \urlstyle{rm}\Url}\fi

\bibitem[Abdal et~al.(2024)Abdal, Yifan, Shi, Xu, Po, Kuang, Chen, Yeung, and
  Wetzstein]{abdal2023gaussian}
Rameen Abdal, Wang Yifan, Zifan Shi, Yinghao Xu, Ryan Po, Zhengfei Kuang,
  Qifeng Chen, Dit-Yan Yeung, and Gordon Wetzstein.
\newblock {Gaussian} shell maps for efficient {3D} human generation.
\newblock In \emph{Proceedings of the IEEE/CVF Conference on Computer Vision
  and Pattern Recognition (CVPR)}, 2024.

\bibitem[Akenine-M{\"o}ller et~al.(2018)Akenine-M{\"o}ller, Haines, Hoffman,
  Pesce, Iwanicki, and Hillaire]{akenine2018realtime}
Tomas Akenine-M{\"o}ller, Eric Haines, Naty Hoffman, Angelo Pesce, Micha{\l}
  Iwanicki, and S{\'e}bastien Hillaire.
\newblock \emph{Real-Time Rendering}.
\newblock A K Peters/CRC Press, fourth edition, 2018.

\bibitem[Alldieck et~al.(2018)Alldieck, Magnor, Xu, Theobalt, and
  Pons-Moll]{alldieck2018video}
Thiemo Alldieck, Marcus Magnor, Weipeng Xu, Christian Theobalt, and Gerard
  Pons-Moll.
\newblock Video based reconstruction of {3D} people models.
\newblock In \emph{Proceedings of the IEEE/CVF Conference on Computer Vision
  and Pattern Recognition (CVPR)}, 2018.

\bibitem[Alldieck et~al.(2019)Alldieck, Magnor, Bhatnagar, Theobalt, and
  Pons-Moll]{alldieck19cvpr}
Thiemo Alldieck, Marcus Magnor, Bharat~Lal Bhatnagar, Christian Theobalt, and
  Gerard Pons-Moll.
\newblock Learning to reconstruct people in clothing from a single {RGB}
  camera.
\newblock In \emph{Proceedings of the IEEE/CVF Conference on Computer Vision
  and Pattern Recognition (CVPR)}, 2019.

\bibitem[Alldieck et~al.(2021)Alldieck, Xu, and
  Sminchisescu]{alldieck2021imghum}
Thiemo Alldieck, Hongyi Xu, and Cristian Sminchisescu.
\newblock {imGHUM}: Implicit generative models of {3D} human shape and
  articulated pose.
\newblock In \emph{Proceedings of the IEEE/CVF International Conference on
  Computer Vision (ICCV)}, 2021.

\bibitem[Anthropic(2026)]{anthropic2026claudeopus55}
Anthropic.
\newblock {Claude Opus 5.5}.
\newblock \url{https://www.anthropic.com/claude}, 2026.
\newblock Large language model, September 2026 version.

\bibitem[Barron et~al.(2022)Barron, Mildenhall, Verbin, Srinivasan, and
  Hedman]{barron2022mip}
Jonathan~T Barron, Ben Mildenhall, Dor Verbin, Pratul~P Srinivasan, and Peter
  Hedman.
\newblock {Mip-NeRF} 360: Unbounded anti-aliased neural radiance fields.
\newblock In \emph{Proceedings of the IEEE/CVF Conference on Computer Vision
  and Pattern Recognition (CVPR)}, 2022.

\bibitem[Barron et~al.(2023)Barron, Mildenhall, Verbin, Srinivasan, and
  Hedman]{barron2023zipnerf}
Jonathan~T. Barron, Ben Mildenhall, Dor Verbin, Pratul~P. Srinivasan, and Peter
  Hedman.
\newblock {Zip-NeRF}: Anti-aliased grid-based neural radiance fields.
\newblock In \emph{Proceedings of the IEEE/CVF International Conference on
  Computer Vision (ICCV)}, 2023.

\bibitem[Besl and McKay(1992)]{besl1992icp}
Paul~J. Besl and Neil~D. McKay.
\newblock A method for registration of {3-D} shapes.
\newblock \emph{IEEE Transactions on Pattern Analysis and Machine
  Intelligence}, 14\penalty0 (2), 1992.

\bibitem[Bhatnagar et~al.(2019)Bhatnagar, Tiwari, Theobalt, and
  Pons-Moll]{bhatnagar2019mgn}
Bharat~Lal Bhatnagar, Garvita Tiwari, Christian Theobalt, and Gerard Pons-Moll.
\newblock {Multi-Garment Net}: Learning to dress {3D} people from images.
\newblock In \emph{Proceedings of the IEEE/CVF International Conference on
  Computer Vision (ICCV)}, 2019.

\bibitem[Bian et~al.(2025)Bian, Xu, Xiu, Grigorev, Liu, Lu, Black, and
  Feng]{bian2025chatgarment}
Siyuan Bian, Chenghao Xu, Yuliang Xiu, Artur Grigorev, Zhen Liu, Cewu Lu,
  Michael~J. Black, and Yao Feng.
\newblock {ChatGarment}: Garment estimation, generation and editing via large
  language models.
\newblock In \emph{Proceedings of the IEEE/CVF Conference on Computer Vision
  and Pattern Recognition (CVPR)}, 2025.

\bibitem[Blattmann et~al.(2023)Blattmann, Dockhorn, Kulal, Mendelevitch,
  Kilian, Lorenz, Levi, English, Voleti, Letts, Jampani, and
  Rombach]{blattmann2023svd}
Andreas Blattmann, Tim Dockhorn, Sumith Kulal, Daniel Mendelevitch, Maciej
  Kilian, Dominik Lorenz, Yam Levi, Zion English, Vikram Voleti, Adam Letts,
  Varun Jampani, and Robin Rombach.
\newblock Stable video diffusion: Scaling latent video diffusion models to
  large datasets.
\newblock \emph{arXiv preprint arXiv:2311.15127}, 2023.

\bibitem[{Blender Online Community}(2026)]{blender2026blender}
{Blender Online Community}.
\newblock {Blender}: A {3D} modelling and rendering package.
\newblock \url{https://www.blender.org/}, 2026.
\newblock Accessed: 2026-11-15.

\bibitem[Bogo et~al.(2016)Bogo, Kanazawa, Lassner, Gehler, Romero, and
  Black]{Bogo2016keepitsmpl}
Federica Bogo, Angjoo Kanazawa, Christoph Lassner, Peter Gehler, Javier Romero,
  and Michael~J. Black.
\newblock Keep it {SMPL}: Automatic estimation of {3D} human pose and shape
  from a single image.
\newblock In \emph{European Conference on Computer Vision (ECCV)}, 2016.

\bibitem[Buehler et~al.(2026)Buehler, Yuan, Li, Huang, Nagano, and
  Iqbal]{buehler2026dreamliftanimate}
Marcel~C. Buehler, Ye Yuan, Xueting Li, Yangyi Huang, Koki Nagano, and Umar
  Iqbal.
\newblock Dream, lift, animate: From single images to animatable {Gaussian}
  avatars.
\newblock In \emph{International Conference on 3D Vision (3DV)}, 2026.

\bibitem[Carion et~al.(2026)]{carion2025sam3}
Nicolas Carion et~al.
\newblock {SAM} 3: Segment anything with concepts.
\newblock In \emph{International Conference on Learning Representations
  (ICLR)}, 2026.

\bibitem[Chen et~al.(2025{\natexlab{a}})Chen, Peng, Tao, and
  Zhang]{chen2025d3human}
Honghu Chen, Bo Peng, Yunfan Tao, and Juyong Zhang.
\newblock {D$^3$-Human}: Dynamic disentangled digital human from monocular
  video.
\newblock In \emph{Proceedings of the IEEE/CVF Conference on Computer Vision
  and Pattern Recognition (CVPR)}, 2025{\natexlab{a}}.

\bibitem[Chen et~al.(2026)Chen, Peng, Xiang, Zhou, Chen, and
  Zhang]{chen2026singlecamera}
Honghu Chen, Bo Peng, Jun Xiang, Jingtao Zhou, Xujing Chen, and Juyong Zhang.
\newblock Single camera based simulation and try-on ready avatar modeling.
\newblock \emph{IEEE Transactions on Visualization and Computer Graphics},
  2026.

\bibitem[Chen(2024)]{chen2024ggavatar}
Jingxuan Chen.
\newblock {GGAvatar}: Reconstructing garment-separated {3D} {Gaussian
  Splatting} avatars from monocular video.
\newblock In \emph{Proceedings of the ACM International Conference on
  Multimedia in Asia (MMAsia)}, 2024.

\bibitem[Chen et~al.(2025{\natexlab{b}})Chen, Li, Zheng, Zhao, Li, Zhu, Dou,
  Wang, and Liu]{chen2025synchuman}
Wenyue Chen, Peng Li, Wangguandong Zheng, Chengfeng Zhao, Mengfei Li, Yaolong
  Zhu, Zhiyang Dou, Ronggang Wang, and Yuan Liu.
\newblock {SyncHuman}: Synchronizing {2D} and {3D} generative models for
  single-view human reconstruction.
\newblock In \emph{Advances in Neural Information Processing Systems
  (NeurIPS)}, 2025{\natexlab{b}}.

\bibitem[Chen et~al.(2021)Chen, Zheng, Black, Hilliges, and
  Geiger]{chen2021snarf}
Xu Chen, Yufeng Zheng, Michael~J Black, Otmar Hilliges, and Andreas Geiger.
\newblock {SNARF}: Differentiable forward skinning for animating non-rigid
  neural implicit shapes.
\newblock In \emph{Proceedings of the IEEE/CVF International Conference on
  Computer Vision (ICCV)}, 2021.

\bibitem[{CLO Virtual Fashion}(2026)]{clo2026marvelous}
{CLO Virtual Fashion}.
\newblock {Marvelous Designer}.
\newblock \url{https://www.marvelousdesigner.com/}, 2026.
\newblock Accessed: 2026-11-15.

\bibitem[Corona et~al.(2021)Corona, Pumarola, Aleny{\`a}, Pons-Moll, and
  Moreno-Noguer]{corona2021smplicit}
Enric Corona, Albert Pumarola, Guillem Aleny{\`a}, Gerard Pons-Moll, and
  Francesc Moreno-Noguer.
\newblock {SMPLicit}: Topology-aware generative model for clothed people.
\newblock In \emph{Proceedings of the IEEE/CVF Conference on Computer Vision
  and Pattern Recognition (CVPR)}, 2021.

\bibitem[Deng et~al.(2020)Deng, Lewis, Jeruzalski, Pons-Moll, Hinton, Norouzi,
  and Tagliasacchi]{deng2020nasa}
Boyang Deng, John~P Lewis, Timothy Jeruzalski, Gerard Pons-Moll, Geoffrey
  Hinton, Mohammad Norouzi, and Andrea Tagliasacchi.
\newblock {NASA}: Neural articulated shape approximation.
\newblock In \emph{European Conference on Computer Vision (ECCV)}, 2020.

\bibitem[Dhamo et~al.(2024)Dhamo, Nie, Moreau, Song, Shaw, Zhou, and
  P{\'e}rez-Pellitero]{dhamo2023headgas}
Helisa Dhamo, Yinyu Nie, Arthur Moreau, Jifei Song, Richard Shaw, Yiren Zhou,
  and Eduardo P{\'e}rez-Pellitero.
\newblock {HeadGaS}: Real-time animatable head avatars via {3D} {Gaussian
  Splatting}.
\newblock In \emph{European Conference on Computer Vision (ECCV)}, 2024.

\bibitem[Dong et~al.(2025)Dong, Duan, Song, Black, and Geiger]{dong2025moga}
Zijian Dong, Longteng Duan, Jie Song, Michael~J. Black, and Andreas Geiger.
\newblock {MoGA}: {3D} generative avatar prior for monocular {Gaussian} avatar
  reconstruction.
\newblock In \emph{Proceedings of the IEEE/CVF International Conference on
  Computer Vision (ICCV)}, 2025.

\bibitem[{Epic Games}(2026{\natexlab{a}})]{epic2026metahuman}
{Epic Games}.
\newblock {MetaHuman}.
\newblock \url{https://www.metahuman.com/}, 2026{\natexlab{a}}.
\newblock Accessed: 2026-11-15.

\bibitem[{Epic Games}(2026{\natexlab{b}})]{epic2026unreal}
{Epic Games}.
\newblock {Unreal Engine}.
\newblock \url{https://www.unrealengine.com/}, 2026{\natexlab{b}}.
\newblock Accessed: 2026-11-15.

\bibitem[Feng et~al.(2022)Feng, Yang, Pollefeys, Black, and
  Bolkart]{feng2022scarf}
Yao Feng, Jinlong Yang, Marc Pollefeys, Michael~J. Black, and Timo Bolkart.
\newblock Capturing and animation of body and clothing from monocular video.
\newblock In \emph{ACM SIGGRAPH Asia Conference Papers}, 2022.

\bibitem[Feng et~al.(2023)Feng, Liu, Bolkart, Yang, Pollefeys, and
  Black]{feng2023delta}
Yao Feng, Weiyang Liu, Timo Bolkart, Jinlong Yang, Marc Pollefeys, and
  Michael~J. Black.
\newblock Learning disentangled avatars with hybrid {3D} representations.
\newblock \emph{arXiv preprint arXiv:2309.06441}, 2023.

\bibitem[Feng et~al.(2024)Feng, Lin, Dwivedi, Sun, Patel, and
  Black]{feng2024chatpose}
Yao Feng, Jing Lin, Sai~Kumar Dwivedi, Yu Sun, Priyanka Patel, and Michael~J.
  Black.
\newblock {ChatPose}: Chatting about {3D} human pose.
\newblock In \emph{Proceedings of the IEEE/CVF Conference on Computer Vision
  and Pattern Recognition (CVPR)}, 2024.

\bibitem[{Google DeepMind}(2026)]{google2026gemini4argon}
{Google DeepMind}.
\newblock {Gemini 4 Argon}.
\newblock \url{https://ai.google.dev/gemini-api/docs/models/gemini-4-argon},
  2026.
\newblock Large language model, October 2026 version.

\bibitem[Gregory(2018)]{gregory2018gameengine}
Jason Gregory.
\newblock \emph{Game Engine Architecture}.
\newblock A K Peters/CRC Press, third edition, 2018.

\bibitem[G{\"u}ler et~al.(2018)G{\"u}ler, Neverova, and
  Kokkinos]{guler2018densepose}
R{\i}za~Alp G{\"u}ler, Natalia Neverova, and Iasonas Kokkinos.
\newblock {DensePose}: Dense human pose estimation in the wild.
\newblock In \emph{Proceedings of the IEEE/CVF Conference on Computer Vision
  and Pattern Recognition (CVPR)}, 2018.

\bibitem[Guo et~al.(2023)Guo, Jiang, Chen, Song, and
  Hilliges]{guo2023vid2avatar}
Chen Guo, Tianjian Jiang, Xu Chen, Jie Song, and Otmar Hilliges.
\newblock {Vid2Avatar}: {3D} avatar reconstruction from videos in the wild via
  self-supervised scene decomposition.
\newblock In \emph{Proceedings of the IEEE/CVF Conference on Computer Vision
  and Pattern Recognition (CVPR)}, 2023.

\bibitem[Guo et~al.(2024)Guo, Jiang, Kaufmann, Zheng, Valentin, Song, and
  Hilliges]{guo2024reloo}
Chen Guo, Tianjian Jiang, Manuel Kaufmann, Chengwei Zheng, Julien Valentin, Jie
  Song, and Otmar Hilliges.
\newblock {ReLoo}: Reconstructing humans dressed in loose garments from
  monocular video in the wild.
\newblock In \emph{European Conference on Computer Vision (ECCV)}, 2024.

\bibitem[Guo et~al.(2025)Guo, Li, Kant, Sheikh, Saito, and
  Cao]{guo2025vid2avatarpro}
Chen Guo, Junxuan Li, Yash Kant, Yaser Sheikh, Shunsuke Saito, and Chen Cao.
\newblock {Vid2Avatar-Pro}: Authentic avatar from videos in the wild via
  universal prior.
\newblock In \emph{Proceedings of the IEEE/CVF Conference on Computer Vision
  and Pattern Recognition (CVPR)}, 2025.

\bibitem[Gupta and Kembhavi(2023)]{gupta2023visprog}
Tanmay Gupta and Aniruddha Kembhavi.
\newblock Visual programming: Compositional visual reasoning without training.
\newblock In \emph{Proceedings of the IEEE/CVF Conference on Computer Vision
  and Pattern Recognition (CVPR)}, 2023.

\bibitem[Guzov et~al.(2021)Guzov, Mir, Sattler, and Pons-Moll]{mir20hps}
Vladimir Guzov, Aymen Mir, Torsten Sattler, and Gerard Pons-Moll.
\newblock Human {POSEitioning} system ({HPS}): {3D} human pose estimation and
  self-localization in large scenes from body-mounted sensors.
\newblock In \emph{Proceedings of the IEEE/CVF Conference on Computer Vision
  and Pattern Recognition (CVPR)}, 2021.

\bibitem[Habermann et~al.(2023)Habermann, Liu, Xu, Pons-Moll, Zollhoefer, and
  Theobalt]{haberman2023hdhumans}
Marc Habermann, Lingjie Liu, Weipeng Xu, Gerard Pons-Moll, Michael Zollhoefer,
  and Christian Theobalt.
\newblock {HDHumans}: A hybrid approach for high-fidelity digital humans.
\newblock \emph{Proceedings of the ACM on Computer Graphics and Interactive
  Techniques}, 6\penalty0 (3), 2023.

\bibitem[Han et~al.(2023)Han, Park, Yoon, Kang, Park, and
  Jeon]{han2023highfidelity}
Sang-Hun Han, Min-Gyu Park, Ju~Hong Yoon, Ju-Mi Kang, Young-Jae Park, and
  Hae-Gon Jeon.
\newblock High-fidelity {3D} human digitization from single {2K} resolution
  images.
\newblock In \emph{Proceedings of the IEEE/CVF Conference on Computer Vision
  and Pattern Recognition (CVPR)}, 2023.

\bibitem[He et~al.(2026)He, Luo, Ma, and Averbuch-Elor]{he2026thinking}
Guangzhao He, Rundong Luo, Wei-Chiu Ma, and Hadar Averbuch-Elor.
\newblock Thinking in {Blender}: Staged executable inverse graphics with
  vision-language models.
\newblock \emph{arXiv preprint arXiv:2606.02580}, 2026.

\bibitem[He et~al.(2021)He, Xu, Saito, Soatto, and Tung]{he2021arch++}
Tong He, Yuanlu Xu, Shunsuke Saito, Stefano Soatto, and Tony Tung.
\newblock Arch++: Animation-ready clothed human reconstruction revisited.
\newblock In \emph{Proceedings of the IEEE/CVF International Conference on
  Computer Vision (ICCV)}, 2021.

\bibitem[He et~al.(2025)He, Wu, Li, Kang, Zhang, Ye, Chen, Gao, Zhang, and
  Zhuang]{he2024magicman}
Xu He, Zhiyong Wu, Xiaoyu Li, Di Kang, Chaopeng Zhang, Jiangnan Ye, Liyang
  Chen, Xiangjun Gao, Han Zhang, and Haolin Zhuang.
\newblock {MagicMan}: Generative novel view synthesis of humans with {3D}-aware
  diffusion and iterative refinement.
\newblock In \emph{Proceedings of the AAAI Conference on Artificial
  Intelligence}, 2025.

\bibitem[Ho et~al.(2023)Ho, Xue, Song, and Hilliges]{ho2023customhumans}
Hsuan-I Ho, Lixin Xue, Jie Song, and Otmar Hilliges.
\newblock Learning locally editable virtual humans.
\newblock In \emph{Proceedings of the IEEE/CVF Conference on Computer Vision
  and Pattern Recognition (CVPR)}, 2023.

\bibitem[Ho et~al.(2020)Ho, Jain, and Abbeel]{Ho2020DDPM}
Jonathan Ho, Ajay Jain, and Pieter Abbeel.
\newblock Denoising diffusion probabilistic models.
\newblock In \emph{Advances in Neural Information Processing Systems
  (NeurIPS)}, 2020.

\bibitem[Hong et~al.(2024)Hong, Zhang, Gu, Bi, Zhou, Liu, Liu, Sunkavalli, Bui,
  and Tan]{hong2023lrm}
Yicong Hong, Kai Zhang, Jiuxiang Gu, Sai Bi, Yang Zhou, Difan Liu, Feng Liu,
  Kalyan Sunkavalli, Trung Bui, and Hao Tan.
\newblock {LRM}: Large reconstruction model for single image to {3D}.
\newblock In \emph{International Conference on Learning Representations
  (ICLR)}, 2024.

\bibitem[Hu et~al.(2026)Hu, Zhao, Guo, Jiang, Liu, Fan, Wang, Wang, and
  Pavlakos]{hu2026humannova}
Hezhen Hu, Wangbo Zhao, Lanqing Guo, Hanwen Jiang, Jonathan~C. Liu, Zhiwen Fan,
  Kai Wang, Zhangyang Wang, and Georgios Pavlakos.
\newblock {HumanNOVA}: Photorealistic, universal and rapid {3D} human avatar
  modeling from a single image.
\newblock In \emph{Proceedings of the IEEE/CVF Conference on Computer Vision
  and Pattern Recognition (CVPR)}, 2026.

\bibitem[Hu et~al.(2024{\natexlab{a}})Hu, Gao, Zhang, Sun, Zhang, and
  Bo]{hu2024animateanyone}
Li Hu, Xin Gao, Peng Zhang, Ke Sun, Bang Zhang, and Liefeng Bo.
\newblock Animate anyone: Consistent and controllable image-to-video synthesis
  for character animation.
\newblock In \emph{Proceedings of the IEEE/CVF Conference on Computer Vision
  and Pattern Recognition (CVPR)}, 2024{\natexlab{a}}.

\bibitem[Hu et~al.(2024{\natexlab{b}})Hu, Zhang, Zhang, Zhou, Liu, Zhang, and
  Nie]{hu2024gaussianavatar}
Liangxiao Hu, Hongwen Zhang, Yuxiang Zhang, Boyao Zhou, Boning Liu, Shengping
  Zhang, and Liqiang Nie.
\newblock {GaussianAvatar}: Towards realistic human avatar modeling from a
  single video via animatable {3D} {Gaussians}.
\newblock In \emph{Proceedings of the IEEE/CVF Conference on Computer Vision
  and Pattern Recognition (CVPR)}, 2024{\natexlab{b}}.

\bibitem[Hu et~al.(2024{\natexlab{c}})Hu, Hu, and Liu]{hu2024gauhuman}
Shoukang Hu, Tao Hu, and Ziwei Liu.
\newblock {GauHuman}: Articulated {Gaussian Splatting} from monocular human
  videos.
\newblock In \emph{Proceedings of the IEEE/CVF Conference on Computer Vision
  and Pattern Recognition (CVPR)}, 2024{\natexlab{c}}.

\bibitem[Hu et~al.(2024{\natexlab{d}})Hu, Iscen, Jain, Kipf, Yue, Ross, Schmid,
  and Fathi]{hu2024scenecraft}
Ziniu Hu, Ahmet Iscen, Aashi Jain, Thomas Kipf, Yisong Yue, David~A. Ross,
  Cordelia Schmid, and Alireza Fathi.
\newblock {SceneCraft}: An {LLM} agent for synthesizing {3D} scenes as
  {Blender} code.
\newblock In \emph{International Conference on Machine Learning (ICML)},
  2024{\natexlab{d}}.

\bibitem[Huang et~al.(2024{\natexlab{a}})Huang, Yang, and
  Guibas]{huang2024blenderalchemy}
Ian Huang, Guandao Yang, and Leonidas Guibas.
\newblock {BlenderAlchemy}: Editing {3D} graphics with vision-language models.
\newblock In \emph{European Conference on Computer Vision (ECCV)},
  2024{\natexlab{a}}.

\bibitem[Huang et~al.(2024{\natexlab{b}})Huang, Yi, Xiu, Liao, Tang, Cai, and
  Thies]{huang2024tech}
Yangyi Huang, Hongwei Yi, Yuliang Xiu, Tingting Liao, Jiaxiang Tang, Deng Cai,
  and Justus Thies.
\newblock {TeCH}: Text-guided reconstruction of lifelike clothed humans.
\newblock In \emph{International Conference on 3D Vision (3DV)},
  2024{\natexlab{b}}.

\bibitem[Huang et~al.(2025)Huang, Yuan, Li, Kautz, and
  Iqbal]{huang2025adahuman}
Yangyi Huang, Ye Yuan, Xueting Li, Jan Kautz, and Umar Iqbal.
\newblock {AdaHuman}: Animatable detailed {3D} human generation with
  compositional multiview diffusion.
\newblock In \emph{Proceedings of the IEEE/CVF International Conference on
  Computer Vision (ICCV)}, 2025.

\bibitem[Huang et~al.(2020)Huang, Xu, Lassner, Li, and Tung]{huang2020arch}
Zeng Huang, Yuanlu Xu, Christoph Lassner, Hao Li, and Tony Tung.
\newblock Arch: Animatable reconstruction of clothed humans.
\newblock In \emph{Proceedings of the IEEE/CVF Conference on Computer Vision
  and Pattern Recognition (CVPR)}, 2020.

\bibitem[Huang et~al.(2026)Huang, Li, Chiu, Lyu, Zhou, Yao, Lasenby, and
  Wu]{huang2026litereality}
Zhening Huang, Yueyan Li, Johnathan Chiu, Xiaoyang Lyu, Matt Zhou, Yuxin Yao,
  Joan Lasenby, and Shangzhe Wu.
\newblock {LiteReality-Agent}: An agentic system for interactable {3D} indoor
  scene reconstruction.
\newblock \emph{arXiv preprint arXiv:2610.01863}, 2026.

\bibitem[Huang-Menders et~al.(2025)Huang-Menders, Liu, Xu, Zhang, Tang, and
  Tai]{huangmenders2025smartavatar}
Alexander Huang-Menders, Xinhang Liu, Andy Xu, Yuyao Zhang, Chi-Keung Tang, and
  Yu-Wing Tai.
\newblock {SmartAvatar}: Text- and image-guided human avatar generation with
  {VLM} {AI} agents.
\newblock \emph{arXiv preprint arXiv:2506.04606}, 2025.

\bibitem[Jiang et~al.(2020)Jiang, Zhang, Hong, Luo, Liu, and
  Bao]{jiang2020bcnet}
Boyi Jiang, Juyong Zhang, Yang Hong, Jinhao Luo, Ligang Liu, and Hujun Bao.
\newblock {BCNet}: Learning body and cloth shape from a single image.
\newblock In \emph{European Conference on Computer Vision (ECCV)}, 2020.

\bibitem[Jiang et~al.(2022{\natexlab{a}})Jiang, Hong, Bao, and
  Zhang]{jiang2022selfrecon}
Boyi Jiang, Yang Hong, Hujun Bao, and Juyong Zhang.
\newblock {SelfRecon}: Self reconstruction your digital avatar from monocular
  video.
\newblock In \emph{Proceedings of the IEEE/CVF Conference on Computer Vision
  and Pattern Recognition (CVPR)}, 2022{\natexlab{a}}.

\bibitem[Jiang et~al.(2026)Jiang, Chen, Tao, Li, Guo, Wu, Yang, and
  Jiang]{jiang2026dressagent}
Nanxiang Jiang, Yunuo Chen, Zeng Tao, Zizhang Li, Qi Guo, Jiajun Wu, Yin Yang,
  and Chenfanfu Jiang.
\newblock Inverse garment design with visual agents.
\newblock \emph{arXiv preprint arXiv:2610.15582}, 2026.

\bibitem[Jiang et~al.(2022{\natexlab{b}})Jiang, Yi, Samei, Tuzel, and
  Ranjan]{jian2022neuman}
Wei Jiang, Kwang~Moo Yi, Golnoosh Samei, Oncel Tuzel, and Anurag Ranjan.
\newblock {NeuMan}: Neural human radiance field from a single video.
\newblock In \emph{European Conference on Computer Vision (ECCV)},
  2022{\natexlab{b}}.

\bibitem[Jiang et~al.(2024)Jiang, Shen, Hong, Guo, Wu, Zhang, Yu, and
  Xu]{jiang2024robust}
Yuheng Jiang, Zhehao Shen, Yu Hong, Chengcheng Guo, Yize Wu, Yingliang Zhang,
  Jingyi Yu, and Lan Xu.
\newblock Robust dual {Gaussian Splatting} for immersive human-centric
  volumetric videos.
\newblock \emph{ACM Transactions on Graphics}, 43\penalty0 (6), 2024.

\bibitem[Junkawitsch et~al.(2025)Junkawitsch, Sun, Zhu, Theobalt, and
  Habermann]{junkawitsch2025eva}
Hendrik Junkawitsch, Guoxing Sun, Heming Zhu, Christian Theobalt, and Marc
  Habermann.
\newblock {EVA}: Expressive virtual avatars from multi-view videos.
\newblock In \emph{ACM SIGGRAPH Conference Proceedings}, 2025.

\bibitem[Kanazawa et~al.(2018)Kanazawa, Black, Jacobs, and
  Malik]{kanazawaHMR18}
Angjoo Kanazawa, Michael~J. Black, David~W. Jacobs, and Jitendra Malik.
\newblock End-to-end recovery of human shape and pose.
\newblock In \emph{Proceedings of the IEEE/CVF Conference on Computer Vision
  and Pattern Recognition (CVPR)}, 2018.

\bibitem[Kerbl et~al.(2023)Kerbl, Kopanas, Leimk{\"u}hler, and
  Drettakis]{kerbl2023gaussians}
Bernhard Kerbl, Georgios Kopanas, Thomas Leimk{\"u}hler, and George Drettakis.
\newblock {3D} {Gaussian Splatting} for real-time radiance field rendering.
\newblock \emph{ACM Transactions on Graphics}, 42\penalty0 (4), 2023.

\bibitem[Khirodkar et~al.(2026)Khirodkar, Wen, Martinez, Dong, Su, and
  Saito]{khirodkar2026sapiens2}
Rawal Khirodkar, He Wen, Julieta Martinez, Yuan Dong, Zhaoen Su, and Shunsuke
  Saito.
\newblock {Sapiens2}.
\newblock In \emph{International Conference on Learning Representations
  (ICLR)}, 2026.

\bibitem[Kim et~al.(2024)Kim, Kim, Saito, and Joo]{kim2024gala}
Taeksoo Kim, Byungjun Kim, Shunsuke Saito, and Hanbyul Joo.
\newblock {GALA}: Generating animatable layered assets from a single scan.
\newblock In \emph{Proceedings of the IEEE/CVF Conference on Computer Vision
  and Pattern Recognition (CVPR)}, 2024.

\bibitem[Kocabas et~al.(2024)Kocabas, Chang, Gabriel, Tuzel, and
  Ranjan]{kocabas2023hugs}
Muhammed Kocabas, Jen-Hao~Rick Chang, James Gabriel, Oncel Tuzel, and Anurag
  Ranjan.
\newblock {HUGS}: Human {Gaussian} splats.
\newblock In \emph{Proceedings of the IEEE/CVF Conference on Computer Vision
  and Pattern Recognition (CVPR)}, 2024.

\bibitem[Kong et~al.(2024)]{kong2025hunyuanvideo}
Weijie Kong et~al.
\newblock {HunyuanVideo}: A systematic framework for large video generative
  models.
\newblock \emph{arXiv preprint arXiv:2412.03603}, 2024.

\bibitem[Korosteleva and Sorkine-Hornung(2023)]{korosteleva2023garmentcode}
Maria Korosteleva and Olga Sorkine-Hornung.
\newblock {GarmentCode}: Programming parametric sewing patterns.
\newblock \emph{ACM Transactions on Graphics (SIGGRAPH Asia)}, 2023.

\bibitem[Kulits et~al.(2024)Kulits, Feng, Liu, Fern{\'a}ndez~Abrevaya, and
  Black]{kulits2024rethinking}
Peter Kulits, Haiwen Feng, Weiyang Liu, Victoria Fern{\'a}ndez~Abrevaya, and
  Michael~J. Black.
\newblock Re-thinking inverse graphics with large language models.
\newblock \emph{Transactions on Machine Learning Research}, 2024.

\bibitem[Lassner and Zollh\"ofer(2021)]{lassner2021pulsar}
Christoph Lassner and Michael Zollh\"ofer.
\newblock Pulsar: Efficient sphere-based neural rendering.
\newblock In \emph{Proceedings of the IEEE/CVF Conference on Computer Vision
  and Pattern Recognition (CVPR)}, 2021.

\bibitem[Le et~al.(2025)Le, Xie, Liang, Wang, Yang, Ma, Vedder, Krishna,
  Jayaraman, and Eaton]{le2025articulateanything}
Long Le, Jason Xie, William Liang, Hung-Ju Wang, Yue Yang, Yecheng~Jason Ma,
  Kyle Vedder, Arjun Krishna, Dinesh Jayaraman, and Eric Eaton.
\newblock {Articulate-Anything}: Automatic modeling of articulated objects via
  a vision-language foundation model.
\newblock In \emph{International Conference on Learning Representations
  (ICLR)}, 2025.

\bibitem[Lei et~al.(2024)Lei, Wang, Pavlakos, Liu, and Daniilidis]{lei2023gart}
Jiahui Lei, Yufu Wang, Georgios Pavlakos, Lingjie Liu, and Kostas Daniilidis.
\newblock {GART}: {Gaussian} articulated template models.
\newblock In \emph{Proceedings of the IEEE/CVF Conference on Computer Vision
  and Pattern Recognition (CVPR)}, 2024.

\bibitem[Li et~al.(2026{\natexlab{a}})Li, Zhao, Chen, Hu, Guo, Zhang, Shan, and
  Hu]{li2026pixal3d}
Dong-Yang Li, Wang Zhao, Yuxin Chen, Wenbo Hu, Meng-Hao Guo, Fang-Lue Zhang,
  Ying Shan, and Shi-Min Hu.
\newblock {Pixal3D}: Pixel-aligned {3D} generation from images.
\newblock In \emph{ACM SIGGRAPH Conference Proceedings}, 2026{\natexlab{a}}.

\bibitem[Li et~al.(2026{\natexlab{b}})Li, Wu, Zhang, Shi, Wang, Zhao, Su, Yu,
  Zhang, Ding, Wen, Zhang, Liu, and Li]{li2026hat4d}
Jiaxin Li, Yuxiang Wu, Zhenkai Zhang, Xinrui Shi, Haoyuan Wang, Yichen Zhao,
  Linxiang Su, Chenyang Yu, Mingyu Zhang, Yifan Ding, Boran Wen, Li Zhang,
  Ruiyang Liu, and Yong-Lu Li.
\newblock {HAT-4D}: Lifting monocular video for {4D} multi-object interactions
  via human-agent collaboration.
\newblock In \emph{European Conference on Computer Vision (ECCV)},
  2026{\natexlab{b}}.

\bibitem[Li et~al.(2025{\natexlab{a}})Li, Zheng, Liu, Yu, Li, Qi, Chi, Xia,
  Cao, Xue, Luo, and Guo]{li2025pshuman}
Peng Li, Wangguandong Zheng, Yuan Liu, Tao Yu, Yangguang Li, Xingqun Qi,
  Xiaowei Chi, Siyu Xia, Yan-Pei Cao, Wei Xue, Wenhan Luo, and Yike Guo.
\newblock {PSHuman}: Photorealistic single-image {3D} human reconstruction
  using cross-scale multiview diffusion and explicit remeshing.
\newblock In \emph{Proceedings of the IEEE/CVF Conference on Computer Vision
  and Pattern Recognition (CVPR)}, 2025{\natexlab{a}}.

\bibitem[Li et~al.(2022)Li, Tanke, Vo, Zollhofer, Gall, Kanazawa, and
  Lassner]{li2022tava}
Ruilong Li, Julian Tanke, Minh Vo, Michael Zollhofer, Jurgen Gall, Angjoo
  Kanazawa, and Christoph Lassner.
\newblock {TAVA}: Template-free animatable volumetric actors.
\newblock In \emph{European Conference on Computer Vision (ECCV)}, 2022.

\bibitem[Li et~al.(2025{\natexlab{b}})Li, Yu, Du, Jiang, Xie, Chen, Yang, and
  Jiang]{li2025dress1to3}
Xuan Li, Chang Yu, Wenxin Du, Ying Jiang, Tianyi Xie, Yunuo Chen, Yin Yang, and
  Chenfanfu Jiang.
\newblock {Dress-1-to-3}: Single image to simulation-ready {3D} outfit with
  diffusion prior and differentiable physics.
\newblock \emph{ACM Transactions on Graphics}, 2025{\natexlab{b}}.

\bibitem[Li et~al.(2026{\natexlab{c}})Li, Shi, Dong, Zhang, Xu, Lee, Chang,
  Xiang, Pan, and Jiang]{li2026legoanything}
Xirui Li, Peng Shi, Mingwen Dong, Sheng Zhang, Zhuoyan Xu, Dongkyu Lee,
  Shuaichen Chang, Yi Xiang, Lin Pan, and Jiarong Jiang.
\newblock {LEGO-Anything}: Coding agents for {3D} scene reconstruction.
\newblock \emph{arXiv preprint arXiv:2609.36380}, 2026{\natexlab{c}}.

\bibitem[Li et~al.(2024{\natexlab{a}})Li, Sun, Zheng, Wang, Zhang, and
  Liu]{li2024relightableandanimatable}
Zhe Li, Yipengjing Sun, Zerong Zheng, Lizhen Wang, Shengping Zhang, and Yebin
  Liu.
\newblock Animatable and relightable {Gaussians} for high-fidelity human avatar
  modeling.
\newblock \emph{arXiv preprint arXiv:2311.16096}, 2024{\natexlab{a}}.

\bibitem[Li et~al.(2024{\natexlab{b}})Li, Zheng, Wang, and
  Liu]{li2024animatablegaussians}
Zhe Li, Zerong Zheng, Lizhen Wang, and Yebin Liu.
\newblock Animatable {Gaussians}: Learning pose-dependent {Gaussian} maps for
  high-fidelity human avatar modeling.
\newblock In \emph{Proceedings of the IEEE/CVF Conference on Computer Vision
  and Pattern Recognition (CVPR)}, 2024{\natexlab{b}}.

\bibitem[Lin et~al.(2025{\natexlab{a}})Lin, Jiang, Yang, Zheng, Liang, Zhang,
  and Liu]{lin2025omnihuman}
Gaojie Lin, Jianwen Jiang, Jiaqi Yang, Zerong Zheng, Chao Liang, Yuan Zhang,
  and Jingtuo Liu.
\newblock {OmniHuman-1}: Rethinking the scaling-up of one-stage conditioned
  human animation models.
\newblock In \emph{Proceedings of the IEEE/CVF International Conference on
  Computer Vision (ICCV)}, 2025{\natexlab{a}}.

\bibitem[Lin et~al.(2025{\natexlab{b}})Lin, Feng, Liu, and
  Black]{lin2025chathuman}
Jing Lin, Yao Feng, Weiyang Liu, and Michael~J. Black.
\newblock {ChatHuman}: Chatting about {3D} humans with tools.
\newblock In \emph{Proceedings of the IEEE/CVF Conference on Computer Vision
  and Pattern Recognition (CVPR)}, 2025{\natexlab{b}}.

\bibitem[Lin et~al.(2024)Lin, Li, Su, Zheng, Zhang, and Liu]{lin2024layga}
Siyou Lin, Zhe Li, Zhaoqi Su, Zerong Zheng, Hongwen Zhang, and Yebin Liu.
\newblock {LayGA}: Layered {Gaussian} avatars for animatable clothing transfer.
\newblock In \emph{ACM SIGGRAPH Conference Proceedings}, 2024.

\bibitem[Liu et~al.(2021)Liu, Habermann, Rudnev, Sarkar, Gu, and
  Theobalt]{liu2021neural}
Lingjie Liu, Marc Habermann, Viktor Rudnev, Kripasindhu Sarkar, Jiatao Gu, and
  Christian Theobalt.
\newblock Neural actor: Neural free-view synthesis of human actors with pose
  control.
\newblock \emph{ACM Transactions on Graphics}, 2021.

\bibitem[Loper et~al.(2015)Loper, Mahmood, Romero, Pons-Moll, and
  Black]{smpl2015loper}
Matthew Loper, Naureen Mahmood, Javier Romero, Gerard Pons-Moll, and Michael~J.
  Black.
\newblock {SMPL}: A skinned multi-person linear model.
\newblock \emph{ACM Transactions on Graphics}, 34\penalty0 (6), 2015.

\bibitem[Lorensen and Cline(1987)]{lorensen1987marching}
William~E. Lorensen and Harvey~E. Cline.
\newblock Marching cubes: A high resolution {3D} surface construction
  algorithm.
\newblock In \emph{ACM SIGGRAPH Conference Proceedings}, 1987.

\bibitem[Lu et~al.(2025)Lu, Dong, Kwon, Zhao, Dai, and De~la Torre]{lu2025gas}
Yixing Lu, Junting Dong, Youngjoong Kwon, Qin Zhao, Bo Dai, and Fernando De~la
  Torre.
\newblock {GAS}: Generative avatar synthesis from a single image.
\newblock In \emph{Proceedings of the IEEE/CVF International Conference on
  Computer Vision (ICCV)}, 2025.

\bibitem[Luo et~al.(2026)Luo, Liu, Cao, Tancik, Buch, Stearns, and
  Xian]{luo2026scenerig}
Rundong Luo, Yunong Liu, Di Cao, Matthew Tancik, Shyamal Buch, Colton Stearns,
  and Wenqi Xian.
\newblock {SceneRig}: An agentic system for simulation-ready {3D} scene
  reconstruction from single images.
\newblock \emph{arXiv preprint arXiv:2610.28585}, 2026.

\bibitem[Ma et~al.(2020)Ma, Yang, Ranjan, Pujades, Pons-Moll, Tang, and
  Black]{ma2020cape}
Qianli Ma, Jinlong Yang, Anurag Ranjan, Sergi Pujades, Gerard Pons-Moll, Siyu
  Tang, and Michael~J. Black.
\newblock Learning to dress {3D} people in generative clothing.
\newblock In \emph{Proceedings of the IEEE/CVF Conference on Computer Vision
  and Pattern Recognition (CVPR)}, 2020.

\bibitem[Mandi et~al.(2025)Mandi, Weng, Bauer, and Song]{mandi2025real2code}
Zhao Mandi, Yijia Weng, Dominik Bauer, and Shuran Song.
\newblock {Real2Code}: Reconstruct articulated objects via code generation.
\newblock In \emph{International Conference on Learning Representations
  (ICLR)}, 2025.

\bibitem[Mazur et~al.(2026)Mazur, Karaev, Chang, Malik, and
  Shafiullah]{mazur2026agentstar}
Kirill Mazur, Nikita Karaev, Matthew Chang, Jitendra Malik, and Nur
  Muhammad~Mahi Shafiullah.
\newblock {AgentSTAR}: Agentic shape tracking and reconstruction from monocular
  videos.
\newblock \emph{arXiv preprint arXiv:2609.24487}, 2026.

\bibitem[Mescheder et~al.(2019)Mescheder, Oechsle, Niemeyer, Nowozin, and
  Geiger]{mescheder2019occupancy}
Lars Mescheder, Michael Oechsle, Michael Niemeyer, Sebastian Nowozin, and
  Andreas Geiger.
\newblock Occupancy networks: Learning {3D} reconstruction in function space.
\newblock In \emph{Proceedings of the IEEE/CVF Conference on Computer Vision
  and Pattern Recognition (CVPR)}, 2019.

\bibitem[Mildenhall et~al.(2020)Mildenhall, Srinivasan, Tancik, Barron,
  Ramamoorthi, and Ng]{mildenhall2020nerf}
Ben Mildenhall, Pratul~P. Srinivasan, Matthew Tancik, Jonathan~T. Barron, Ravi
  Ramamoorthi, and Ren Ng.
\newblock {NeRF}: Representing scenes as neural radiance fields for view
  synthesis.
\newblock In \emph{European Conference on Computer Vision (ECCV)}, 2020.

\bibitem[Mir et~al.(2020)Mir, Alldieck, and Pons-Moll]{mir2020pix2surf}
Aymen Mir, Thiemo Alldieck, and Gerard Pons-Moll.
\newblock Learning to transfer texture from clothing images to {3D} humans.
\newblock In \emph{Proceedings of the IEEE/CVF Conference on Computer Vision
  and Pattern Recognition (CVPR)}, 2020.

\bibitem[Mir et~al.(2024)Mir, Puig, Kanazawa, and Pons-Moll]{mir2024origin}
Aymen Mir, Xavier Puig, Angjoo Kanazawa, and Gerard Pons-Moll.
\newblock Generating continual human motion in diverse {3D} scenes.
\newblock In \emph{International Conference on 3D Vision (3DV)}, 2024.

\bibitem[Mir et~al.(2026)Mir, Guler, Tang, Wonka, and Pons-Moll]{mir2026ahoy}
Aymen Mir, Riza~Alp Guler, Xiangjun Tang, Peter Wonka, and Gerard Pons-Moll.
\newblock {AHOY}! animatable humans under occlusion from {YouTube} videos with
  {Gaussian Splatting} and video diffusion priors.
\newblock In \emph{European Conference on Computer Vision (ECCV)}, 2026.

\bibitem[Moon et~al.(2024)Moon, Shiratori, and Saito]{moon2024exavatar}
Gyeongsik Moon, Takaaki Shiratori, and Shunsuke Saito.
\newblock Expressive whole-body {3D} {Gaussian} avatar.
\newblock In \emph{European Conference on Computer Vision (ECCV)}, 2024.

\bibitem[Moreau et~al.(2024)Moreau, Song, Dhamo, Shaw, Zhou, and
  P{\'e}rez-Pellitero]{moreau2024human}
Arthur Moreau, Jifei Song, Helisa Dhamo, Richard Shaw, Yiren Zhou, and Eduardo
  P{\'e}rez-Pellitero.
\newblock Human {Gaussian Splatting}: Real-time rendering of animatable
  avatars.
\newblock In \emph{Proceedings of the IEEE/CVF Conference on Computer Vision
  and Pattern Recognition (CVPR)}, 2024.

\bibitem[M\"uller et~al.(2022)M\"uller, Evans, Schied, and
  Keller]{mueller2022instant}
Thomas M\"uller, Alex Evans, Christoph Schied, and Alexander Keller.
\newblock Instant neural graphics primitives with a multiresolution hash
  encoding.
\newblock \emph{ACM Transactions on Graphics}, 41\penalty0 (4), 2022.

\bibitem[Nakayama et~al.(2025)Nakayama, Ackermann, Kesdogan, Zheng,
  Korosteleva, Sorkine-Hornung, Guibas, Yang, and
  Wetzstein]{nakayama2025aipparel}
Kiyohiro Nakayama, Jan Ackermann, Timur~Levent Kesdogan, Yang Zheng, Maria
  Korosteleva, Olga Sorkine-Hornung, Leonidas~J. Guibas, Guandao Yang, and
  Gordon Wetzstein.
\newblock {AIpparel}: A multimodal foundation model for digital garments.
\newblock In \emph{Proceedings of the IEEE/CVF Conference on Computer Vision
  and Pattern Recognition (CVPR)}, 2025.

\bibitem[Neverova et~al.(2018)Neverova, G{\"u}ler, and
  Kokkinos]{neverova2018denseposetransfer}
Natalia Neverova, R{\i}za~Alp G{\"u}ler, and Iasonas Kokkinos.
\newblock Dense pose transfer.
\newblock In \emph{European Conference on Computer Vision (ECCV)}, 2018.

\bibitem[Neverova et~al.(2020)Neverova, Novotny, Szafraniec, Khalidov, Labatut,
  and Vedaldi]{neverova2020cse}
Natalia Neverova, David Novotny, Marc Szafraniec, Vasil Khalidov, Patrick
  Labatut, and Andrea Vedaldi.
\newblock Continuous surface embeddings.
\newblock In \emph{Advances in Neural Information Processing Systems
  (NeurIPS)}, 2020.

\bibitem[OpenAI(2026{\natexlab{a}})]{openai2026gpt6astra}
OpenAI.
\newblock {GPT-6 Astra}.
\newblock \url{https://developers.openai.com/api/docs/models/gpt-6-astra},
  2026{\natexlab{a}}.
\newblock Large language model, September 2026 version.

\bibitem[OpenAI(2026{\natexlab{b}})]{openai2026gptimage25}
OpenAI.
\newblock {GPT-Image-2.5 Sunburst}.
\newblock
  \url{https://developers.openai.com/api/docs/models/gpt-image-2.5-sunburst},
  2026{\natexlab{b}}.
\newblock Image generation model, September 2026 version.

\bibitem[Pan et~al.(2024)Pan, Su, Lin, Fan, Zhang, Li, Shen, Mu, and
  Liu]{pan2025humansplat}
Panwang Pan, Zhuo Su, Chenguo Lin, Zhen Fan, Yongjie Zhang, Zeming Li, Tingting
  Shen, Yadong Mu, and Yebin Liu.
\newblock {HumanSplat}: Generalizable single-image human {Gaussian Splatting}
  with structure priors.
\newblock In \emph{Advances in Neural Information Processing Systems
  (NeurIPS)}, 2024.

\bibitem[Pang et~al.(2024)Pang, Zhu, Kortylewski, Theobalt, and
  Habermann]{pang2024ash}
Haokai Pang, Heming Zhu, Adam Kortylewski, Christian Theobalt, and Marc
  Habermann.
\newblock {ASH}: Animatable {Gaussian} splats for efficient and photoreal human
  rendering.
\newblock In \emph{Proceedings of the IEEE/CVF Conference on Computer Vision
  and Pattern Recognition (CVPR)}, 2024.

\bibitem[Pang et~al.(2025)Pang, Liu, Cai, Yang, Zhang, and
  Liu]{pang2024disco4d}
Hui~En Pang, Shuai Liu, Zhongang Cai, Lei Yang, Tianwei Zhang, and Ziwei Liu.
\newblock {DISCO4D}: Disentangled {4D} human generation and animation from a
  single image.
\newblock In \emph{Proceedings of the IEEE/CVF Conference on Computer Vision
  and Pattern Recognition (CVPR)}, 2025.

\bibitem[Park et~al.(2019)Park, Florence, Straub, Newcombe, and
  Lovegrove]{park2019deepsdf}
Jeong~Joon Park, Peter Florence, Julian Straub, Richard Newcombe, and Steven
  Lovegrove.
\newblock {DeepSDF}: Learning continuous signed distance functions for shape
  representation.
\newblock In \emph{Proceedings of the IEEE/CVF Conference on Computer Vision
  and Pattern Recognition (CVPR)}, 2019.

\bibitem[Patel et~al.(2020)Patel, Liao, and Pons-Moll]{patel2020tailornet}
Chaitanya Patel, Zhouyingcheng Liao, and Gerard Pons-Moll.
\newblock {TailorNet}: Predicting clothing in {3D} as a function of human pose,
  shape and garment style.
\newblock In \emph{Proceedings of the IEEE/CVF Conference on Computer Vision
  and Pattern Recognition (CVPR)}, 2020.

\bibitem[Pavlakos et~al.(2019)Pavlakos, Choutas, Ghorbani, Bolkart, Osman,
  Tzionas, and Black]{SMPL-X:2019}
Georgios Pavlakos, Vasileios Choutas, Nima Ghorbani, Timo Bolkart, Ahmed A.~A.
  Osman, Dimitrios Tzionas, and Michael~J. Black.
\newblock Expressive body capture: {3D} hands, face, and body from a single
  image.
\newblock In \emph{Proceedings of the IEEE/CVF Conference on Computer Vision
  and Pattern Recognition (CVPR)}, 2019.

\bibitem[Peng et~al.(2024)Peng, Zhang, Guo, Cao, and Hu]{peng2024charactergen}
Hao-Yang Peng, Jia-Peng Zhang, Meng-Hao Guo, Yan-Pei Cao, and Shi-Min Hu.
\newblock {CharacterGen}: Efficient {3D} character generation from single
  images with multi-view pose canonicalization.
\newblock \emph{ACM Transactions on Graphics}, 2024.

\bibitem[Peng et~al.(2021)Peng, Dong, Wang, Zhang, Shuai, Zhou, and
  Bao]{peng2021animatable}
Sida Peng, Junting Dong, Qianqian Wang, Shangzhan Zhang, Qing Shuai, Xiaowei
  Zhou, and Hujun Bao.
\newblock Animatable neural radiance fields for modeling dynamic human bodies.
\newblock In \emph{Proceedings of the IEEE/CVF International Conference on
  Computer Vision (ICCV)}, 2021.

\bibitem[Pons-Moll et~al.(2017)Pons-Moll, Pujades, Hu, and
  Black]{ponsmoll2017clothcap}
Gerard Pons-Moll, Sergi Pujades, Sonny Hu, and Michael~J. Black.
\newblock {ClothCap}: Seamless {4D} clothing capture and retargeting.
\newblock \emph{ACM Transactions on Graphics}, 2017.

\bibitem[Qian et~al.(2024)Qian, Wang, Mihajlovic, Geiger, and
  Tang]{qian20233dgsavatar}
Zhiyin Qian, Shaofei Wang, Marko Mihajlovic, Andreas Geiger, and Siyu Tang.
\newblock {3DGS-Avatar}: Animatable avatars via deformable {3D} {Gaussian
  Splatting}.
\newblock In \emph{Proceedings of the IEEE/CVF Conference on Computer Vision
  and Pattern Recognition (CVPR)}, 2024.

\bibitem[Qiu et~al.(2023)Qiu, Chen, Zhou, Xu, Wang, and Han]{qiu2023recmv}
Lingteng Qiu, Guanying Chen, Jiapeng Zhou, Mutian Xu, Junle Wang, and Xiaoguang
  Han.
\newblock {REC-MV}: {RE}constructing {3D} dynamic cloth from monocular videos.
\newblock In \emph{Proceedings of the IEEE/CVF Conference on Computer Vision
  and Pattern Recognition (CVPR)}, 2023.

\bibitem[Qiu et~al.(2025{\natexlab{a}})Qiu, Gu, Li, Zuo, Shen, Zhang, Qiu,
  Yuan, Chen, Dong, and Bo]{qiu2025lhm}
Lingteng Qiu, Xiaodong Gu, Peihao Li, Qi Zuo, Weichao Shen, Junfei Zhang, Kejie
  Qiu, Weihao Yuan, Guanying Chen, Zilong Dong, and Liefeng Bo.
\newblock {LHM}: Large animatable human reconstruction model for single image
  to {3D} in seconds.
\newblock In \emph{Proceedings of the IEEE/CVF International Conference on
  Computer Vision (ICCV)}, 2025{\natexlab{a}}.

\bibitem[Qiu et~al.(2025{\natexlab{b}})Qiu, Li, Li, Zuo, Gu, Dong, Yuan, Peng,
  Zhu, Han, Chen, and Dong]{qiu2025pflhm}
Lingteng Qiu, Peihao Li, Heyuan Li, Qi Zuo, Xiaodong Gu, Yuan Dong, Weihao
  Yuan, Rui Peng, Siyu Zhu, Xiaoguang Han, Guanying Chen, and Zilong Dong.
\newblock {LHM++}: An efficient large human reconstruction model for pose-free
  images to {3D}.
\newblock \emph{arXiv preprint arXiv:2506.13766}, 2025{\natexlab{b}}.

\bibitem[Qiu et~al.(2025{\natexlab{c}})Qiu, Zhu, Zuo, Gu, Dong, Zhang, Xu, Li,
  Yuan, Bo, Chen, and Dong]{qiu2025anigs}
Lingteng Qiu, Shenhao Zhu, Qi Zuo, Xiaodong Gu, Yuan Dong, Junfei Zhang, Chao
  Xu, Zhe Li, Weihao Yuan, Liefeng Bo, Guanying Chen, and Zilong Dong.
\newblock {AniGS}: Animatable {Gaussian} avatar from a single image with
  inconsistent {Gaussian} reconstruction.
\newblock In \emph{Proceedings of the IEEE/CVF Conference on Computer Vision
  and Pattern Recognition (CVPR)}, 2025{\natexlab{c}}.

\bibitem[RenderPeople(2026)]{RenderPeople}
RenderPeople.
\newblock {RenderPeople}.
\newblock \url{https://renderpeople.com/}, 2026.
\newblock Accessed: 2026-11-15.

\bibitem[Rombach et~al.(2022)Rombach, Blattmann, Lorenz, Esser, and
  Ommer]{Rombach2022LDM}
Robin Rombach, Andreas Blattmann, Dominik Lorenz, Patrick Esser, and Bj{\"o}rn
  Ommer.
\newblock High-resolution image synthesis with latent diffusion models.
\newblock In \emph{Proceedings of the IEEE/CVF Conference on Computer Vision
  and Pattern Recognition (CVPR)}, 2022.

\bibitem[Saito et~al.(2019)Saito, Huang, Natsume, Morishima, Kanazawa, and
  Li]{saito2019pifu}
Shunsuke Saito, Zeng Huang, Ryota Natsume, Shigeo Morishima, Angjoo Kanazawa,
  and Hao Li.
\newblock {PIFu}: Pixel-aligned implicit function for high-resolution clothed
  human digitization.
\newblock In \emph{Proceedings of the IEEE/CVF International Conference on
  Computer Vision (ICCV)}, 2019.

\bibitem[Saito et~al.(2020)Saito, Simon, Saragih, and Joo]{saito2020pifuhd}
Shunsuke Saito, Tomas Simon, Jason Saragih, and Hanbyul Joo.
\newblock {PIFuHD}: Multi-level pixel-aligned implicit function for
  high-resolution {3D} human digitization.
\newblock In \emph{Proceedings of the IEEE/CVF Conference on Computer Vision
  and Pattern Recognition (CVPR)}, 2020.

\bibitem[S{\'a}r{\'a}ndi and Pons-Moll(2024)]{sarandi2024nlf}
Istv{\'a}n S{\'a}r{\'a}ndi and Gerard Pons-Moll.
\newblock Neural localizer fields for continuous {3D} human pose and shape
  estimation.
\newblock In \emph{Advances in Neural Information Processing Systems
  (NeurIPS)}, 2024.

\bibitem[Shao et~al.(2024)Shao, Pang, Zheng, Sun, and Liu]{shao2024human4dit}
Ruizhi Shao, Youxin Pang, Zerong Zheng, Jingxiang Sun, and Yebin Liu.
\newblock {Human4DiT}: 360-degree human video generation with {4D} diffusion
  transformer.
\newblock \emph{ACM Transactions on Graphics}, 2024.

\bibitem[Shen et~al.(2023)Shen, Guo, Kaufmann, Zarate, Valentin, Song, and
  Hilliges]{shen2023xavatar}
Kaiyue Shen, Chen Guo, Manuel Kaufmann, Juan~Jose Zarate, Julien Valentin, Jie
  Song, and Otmar Hilliges.
\newblock {X-Avatar}: Expressive human avatars.
\newblock In \emph{Proceedings of the IEEE/CVF Conference on Computer Vision
  and Pattern Recognition (CVPR)}, 2023.

\bibitem[Shen et~al.(2026{\natexlab{a}})Shen, Kova{\v{c}}i{\v{c}}, Kulits,
  Wang, Li, Tenenbaum, Yuille, Chen, and Wu]{shen2026fourdcodebench}
Ruihong Shen, {\v{Z}}iga Kova{\v{c}}i{\v{c}}, Peter Kulits, Xingrui Wang,
  Zizhang Li, Joshua~B. Tenenbaum, Alan Yuille, Jieneng Chen, and Jiajun Wu.
\newblock {4DCodeBench}: Benchmarking agents on inverse graphics of dynamic
  scenes.
\newblock \emph{arXiv preprint arXiv:2610.03715}, 2026{\natexlab{a}}.

\bibitem[Shen et~al.(2026{\natexlab{b}})Shen, Wang, Yin, Liu, Yang, Liang, Cai,
  and Lin]{shen2026vlmguided}
Wenhao Shen, Hao Wang, Wanqi Yin, Fayao Liu, Xulei Yang, Chao Liang, Zhongang
  Cai, and Guosheng Lin.
\newblock {VLM}-guided group preference alignment for diffusion-based human
  mesh recovery.
\newblock In \emph{Proceedings of the IEEE/CVF Conference on Computer Vision
  and Pattern Recognition (CVPR)}, 2026{\natexlab{b}}.

\bibitem[Shi et~al.(2026)Shi, Ye, Liu, Liu, Xu, He, Li, Chen, and
  Shen]{shi2026agile}
Jin-Chuan Shi, Binhong Ye, Tao Liu, Xiaoyang Liu, Yangjinhui Xu, Junzhe He,
  Zeju Li, Hao Chen, and Chunhua Shen.
\newblock {AGILE}: Hand-object interaction reconstruction from video via
  agentic generation.
\newblock In \emph{ACM SIGGRAPH Conference Proceedings}, 2026.

\bibitem[Sim and Moon(2025)]{sim2025persona}
Geonhee Sim and Gyeongsik Moon.
\newblock {PERSONA}: Personalized whole-body {3D} avatar with pose-driven
  deformations from a single image.
\newblock In \emph{Proceedings of the IEEE/CVF International Conference on
  Computer Vision (ICCV)}, 2025.

\bibitem[Sun et~al.(2025)Sun, Han, Deng, Wang, Qin, and Gould]{sun20253dgpt}
Chunyi Sun, Junlin Han, Weijian Deng, Xinlong Wang, Zishan Qin, and Stephen
  Gould.
\newblock {3D-GPT}: Procedural {3D} modeling with large language models.
\newblock In \emph{International Conference on 3D Vision (3DV)}, 2025.

\bibitem[Sun et~al.(2026)Sun, Wang, Song, Gu, and Liu]{sun2026harmony}
Shufan Sun, Chen Wang, Enxin Song, Jiatao Gu, and Lingjie Liu.
\newblock {HARMONY}: Hierarchical agentic reasoning for {MONocular}
  image-to-scene synthesis.
\newblock \emph{arXiv preprint arXiv:2609.26793}, 2026.

\bibitem[Sur{\'i}s et~al.(2023)Sur{\'i}s, Menon, and
  Vondrick]{suris2023vipergpt}
D{\'i}dac Sur{\'i}s, Sachit Menon, and Carl Vondrick.
\newblock {ViperGPT}: Visual inference via {Python} execution for reasoning.
\newblock In \emph{Proceedings of the IEEE/CVF International Conference on
  Computer Vision (ICCV)}, 2023.

\bibitem[Tan et~al.(2025)Tan, Xiang, Tulsiani, Ramanan, and
  Yang]{tan2025dressrecon}
Jeff Tan, Donglai Xiang, Shubham Tulsiani, Deva Ramanan, and Gengshan Yang.
\newblock {DressRecon}: Freeform {4D} human reconstruction from monocular
  video.
\newblock In \emph{International Conference on 3D Vision (3DV)}, 2025.

\bibitem[Tancik et~al.(2023)Tancik, Weber, Ng, Li, Yi, Kerr, Wang,
  Kristoffersen, Austin, Salahi, Ahuja, McAllister, and Kanazawa]{nerfstudio}
Matthew Tancik, Ethan Weber, Evonne Ng, Ruilong Li, Brent Yi, Justin Kerr,
  Terrance Wang, Alexander Kristoffersen, Jake Austin, Kamyar Salahi, Abhik
  Ahuja, David McAllister, and Angjoo Kanazawa.
\newblock Nerfstudio: A modular framework for neural radiance field
  development.
\newblock In \emph{ACM SIGGRAPH Conference Proceedings}, 2023.

\bibitem[Tatarchenko et~al.(2019)Tatarchenko, Richter, Ranftl, Li, Koltun, and
  Brox]{tatarchenko2019single}
Maxim Tatarchenko, Stephan~R. Richter, Ren{\'e} Ranftl, Zhuwen Li, Vladlen
  Koltun, and Thomas Brox.
\newblock What do single-view {3D} reconstruction networks learn?
\newblock In \emph{Proceedings of the IEEE/CVF Conference on Computer Vision
  and Pattern Recognition (CVPR)}, 2019.

\bibitem[Taubner et~al.(2025)Taubner, Zhang, Tuli, and
  Lindell]{taubner2025cap4d}
Felix Taubner, Ruihang Zhang, Mathieu Tuli, and David~B. Lindell.
\newblock {CAP4D}: Creating animatable {4D} portrait avatars with morphable
  multi-view diffusion models.
\newblock In \emph{Proceedings of the IEEE/CVF Conference on Computer Vision
  and Pattern Recognition (CVPR)}, 2025.

\bibitem[Tiwari et~al.(2020)Tiwari, Bhatnagar, Tung, and
  Pons-Moll]{tiwari2020sizer}
Garvita Tiwari, Bharat~Lal Bhatnagar, Tony Tung, and Gerard Pons-Moll.
\newblock {SIZER}: A dataset and model for parsing {3D} clothing and learning
  size sensitive {3D} clothing.
\newblock In \emph{European Conference on Computer Vision (ECCV)}, 2020.

\bibitem[Todorov et~al.(2012)Todorov, Erez, and Tassa]{todorov2012mujoco}
Emanuel Todorov, Tom Erez, and Yuval Tassa.
\newblock {MuJoCo}: A physics engine for model-based control.
\newblock In \emph{IEEE/RSJ International Conference on Intelligent Robots and
  Systems (IROS)}, 2012.

\bibitem[Tu et~al.(2025)Tu, Xing, Han, Cheng, Dai, Luo, and
  Wu]{tu2025stableanimator}
Shuyuan Tu, Zhen Xing, Xintong Han, Zhi-Qi Cheng, Qi Dai, Chong Luo, and Zuxuan
  Wu.
\newblock {StableAnimator}: High-quality identity-preserving human image
  animation.
\newblock In \emph{Proceedings of the IEEE/CVF Conference on Computer Vision
  and Pattern Recognition (CVPR)}, 2025.

\bibitem[Waechter et~al.(2014)Waechter, Moehrle, and
  Goesele]{waechter2014color}
Michael Waechter, Nils Moehrle, and Michael Goesele.
\newblock Let there be color! large-scale texturing of {3D} reconstructions.
\newblock In \emph{European Conference on Computer Vision (ECCV)}, 2014.

\bibitem[{Wan Team}(2025)]{wan2025wan}
{Wan Team}.
\newblock Wan: Open and advanced large-scale video generative models.
\newblock \emph{arXiv preprint arXiv:2503.20314}, 2025.

\bibitem[Wang et~al.(2025)Wang, Zhang, Tang, Zhang, Gao, Wang, and
  Sang]{wang2025unianimate_dit}
Xiang Wang, Shiwei Zhang, Longxiang Tang, Yingya Zhang, Changxin Gao, Yuehuan
  Wang, and Nong Sang.
\newblock {UniAnimate-DiT}: Human image animation with large-scale video
  diffusion transformer.
\newblock \emph{arXiv preprint arXiv:2504.11289}, 2025.

\bibitem[Wang et~al.(2004)Wang, Bovik, Sheikh, and Simoncelli]{wang2004ssim}
Zhou Wang, Alan~C. Bovik, Hamid~R. Sheikh, and Eero~P. Simoncelli.
\newblock Image quality assessment: from error visibility to structural
  similarity.
\newblock \emph{IEEE Transactions on Image Processing}, 2004.

\bibitem[Weng et~al.(2022)Weng, Curless, Srinivasan, Barron, and
  Kemelmacher-Shlizerman]{weng_humannerf_2022_cvpr}
Chung-Yi Weng, Brian Curless, Pratul~P. Srinivasan, Jonathan~T. Barron, and Ira
  Kemelmacher-Shlizerman.
\newblock Human{N}e{RF}: Free-viewpoint rendering of moving people from
  monocular video.
\newblock In \emph{Proceedings of the IEEE/CVF Conference on Computer Vision
  and Pattern Recognition (CVPR)}, 2022.

\bibitem[Westover(1991)]{westover1991phdsplatting}
Lee~Alan Westover.
\newblock \emph{Splatting: a parallel, feed-forward volume rendering
  algorithm}.
\newblock PhD thesis, University of North Carolina at Chapel Hill, 1991.

\bibitem[Wu et~al.(2026)Wu, Birchfield, Rabeti, Chang, and Wen]{wu2026gator}
Qirui Wu, Stan Birchfield, Hesam Rabeti, Angel~X. Chang, and Bowen Wen.
\newblock {GATOR}: Generative and agentic {3D} object reconstruction from
  casual images.
\newblock \emph{arXiv preprint arXiv:2610.11215}, 2026.

\bibitem[Xiang et~al.(2026)Xiang, Chen, Xu, Wang, Lv, Deng, Zhu, Dong, Zhao,
  Yuan, and Yang]{xiang2026trellis2}
Jianfeng Xiang, Xiaoxue Chen, Sicheng Xu, Ruicheng Wang, Zelong Lv, Yu Deng,
  Hongyuan Zhu, Yue Dong, Hao Zhao, Nicholas~Jing Yuan, and Jiaolong Yang.
\newblock Native and compact structured latents for {3D} generation.
\newblock In \emph{Proceedings of the IEEE/CVF Conference on Computer Vision
  and Pattern Recognition (CVPR)}, 2026.

\bibitem[Xie et~al.(2026)Xie, Wang, Lin, Wang, and Li]{xie2026codimrecon}
Shuzhao Xie, Lelin Wang, Guying Lin, Zhi Wang, and Minchen Li.
\newblock {CoDimRecon}: Agentic reconstruction of sim-ready {3D} scenes with
  deformable curves, surfaces, and volumes.
\newblock \emph{arXiv preprint arXiv:2609.36024}, 2026.

\bibitem[Xie et~al.(2022)Xie, Takikawa, Saito, Litany, Yan, Khan, Tombari,
  Tompkin, Sitzmann, and Sridhar]{nerf_review}
Yiheng Xie, Towaki Takikawa, Shunsuke Saito, Or Litany, Shiqin Yan, Numair
  Khan, Federico Tombari, James Tompkin, Vincent Sitzmann, and Srinath Sridhar.
\newblock Neural fields in visual computing and beyond.
\newblock \emph{Computer Graphics Forum}, 2022.

\bibitem[Xiu et~al.(2023)Xiu, Yang, Cao, Tzionas, and Black]{xiu2023econ}
Yuliang Xiu, Jinlong Yang, Xu Cao, Dimitrios Tzionas, and Michael~J. Black.
\newblock {ECON}: Explicit clothed humans optimized via normal integration.
\newblock In \emph{Proceedings of the IEEE/CVF Conference on Computer Vision
  and Pattern Recognition (CVPR)}, 2023.

\bibitem[Xu et~al.(2021)Xu, Alldieck, and Sminchisescu]{xu2021hnerf}
Hongyi Xu, Thiemo Alldieck, and Cristian Sminchisescu.
\newblock {H-NeRF}: Neural radiance fields for rendering and temporal
  reconstruction of humans in motion.
\newblock In \emph{Advances in Neural Information Processing Systems
  (NeurIPS)}, 2021.

\bibitem[Xu et~al.(2024{\natexlab{a}})Xu, Chen, Li, Zhang, Wang, Zheng, and
  Liu]{xu2024gaussian}
Yuelang Xu, Benwang Chen, Zhe Li, Hongwen Zhang, Lizhen Wang, Zerong Zheng, and
  Yebin Liu.
\newblock {Gaussian} head avatar: Ultra high-fidelity head avatar via dynamic
  {Gaussians}.
\newblock In \emph{Proceedings of the IEEE/CVF Conference on Computer Vision
  and Pattern Recognition (CVPR)}, 2024{\natexlab{a}}.

\bibitem[Xu et~al.(2024{\natexlab{b}})Xu, Zhang, Liew, Yan, Liu, Zhang, Feng,
  and Shou]{xu2024magicanimate}
Zhongcong Xu, Jianfeng Zhang, Jun~Hao Liew, Hanshu Yan, Jia-Wei Liu, Chenxu
  Zhang, Jiashi Feng, and Mike~Zheng Shou.
\newblock {MagicAnimate}: Temporally consistent human image animation using
  diffusion model.
\newblock In \emph{Proceedings of the IEEE/CVF Conference on Computer Vision
  and Pattern Recognition (CVPR)}, 2024{\natexlab{b}}.

\bibitem[Yan et~al.(2026)Yan, Guo, Yang, and Tang]{yan2026scail2}
Wenhao Yan, Fengjia Guo, Zhuoyi Yang, and Jie Tang.
\newblock {SCAIL-2}: Unifying controlled character animation with end-to-end
  in-context conditioning.
\newblock \emph{arXiv preprint arXiv:2606.10804}, 2026.

\bibitem[Yang et~al.(2024)Yang, Chen, Gao, Wang, Han, and
  Wang]{yang2024havefun}
Xihe Yang, Xingyu Chen, Daiheng Gao, Shaohui Wang, Xiaoguang Han, and Baoyuan
  Wang.
\newblock {HAVE-FUN}: Human avatar reconstruction from few-shot unconstrained
  images.
\newblock In \emph{Proceedings of the IEEE/CVF Conference on Computer Vision
  and Pattern Recognition (CVPR)}, 2024.

\bibitem[Yang et~al.(2025)Yang, Teng, Zheng, Ding, Huang, Xu, Yang, Hong,
  Zhang, Feng, Yin, Zhang, Wang, Cheng, Xu, Gu, Dong, and
  Tang]{yang2024cogvideox}
Zhuoyi Yang, Jiayan Teng, Wendi Zheng, Ming Ding, Shiyu Huang, Jiazheng Xu,
  Yuanming Yang, Wenyi Hong, Xiaohan Zhang, Guanyu Feng, Da Yin, Yuxuan Zhang,
  Weihan Wang, Yean Cheng, Bin Xu, Xiaotao Gu, Yuxiao Dong, and Jie Tang.
\newblock {CogVideoX}: Text-to-video diffusion models with an expert
  transformer.
\newblock In \emph{International Conference on Learning Representations
  (ICLR)}, 2025.

\bibitem[Yin et~al.(2026)Yin, Ge, Wang, Wang, Li, Black, Darrell, Kanazawa, and
  Feng]{yin2026viga}
Shaofeng Yin, Jiaxin Ge, Zora~Zhiruo Wang, Chenyang Wang, Xiuyu Li, Michael~J.
  Black, Trevor Darrell, Angjoo Kanazawa, and Haiwen Feng.
\newblock Vision-as-inverse-graphics agent via interleaved multimodal
  reasoning.
\newblock In \emph{European Conference on Computer Vision (ECCV)}, 2026.

\bibitem[Yu et~al.(2021)Yu, Zheng, Guo, Liu, Dai, and Liu]{yu2021function4d}
Tao Yu, Zerong Zheng, Kaiwen Guo, Pengpeng Liu, Qionghai Dai, and Yebin Liu.
\newblock {Function4D}: Real-time human volumetric capture from very sparse
  consumer {RGBD} sensors.
\newblock In \emph{Proceedings of the IEEE/CVF Conference on Computer Vision
  and Pattern Recognition (CVPR)}, 2021.

\bibitem[Zhang et~al.(2018)Zhang, Isola, Efros, Shechtman, and
  Wang]{zhang2018lpips}
Richard Zhang, Phillip Isola, Alexei~A. Efros, Eli Shechtman, and Oliver Wang.
\newblock The unreasonable effectiveness of deep features as a perceptual
  metric.
\newblock In \emph{Proceedings of the IEEE/CVF Conference on Computer Vision
  and Pattern Recognition (CVPR)}, 2018.

\bibitem[Zhang et~al.(2025)Zhang, Gu, Wang, Wang, Cheng, Zhu, and
  Zou]{zhang2024mimicmotion}
Yuang Zhang, Jiaxi Gu, Li-Wen Wang, Han Wang, Junqi Cheng, Yuefeng Zhu, and
  Fangyuan Zou.
\newblock {MimicMotion}: High-quality human motion video generation with
  confidence-aware pose guidance.
\newblock In \emph{International Conference on Machine Learning (ICML)}, 2025.

\bibitem[Zhang et~al.(2024)Zhang, Yang, and Yang]{zhang2024sifu}
Zechuan Zhang, Zongxin Yang, and Yi Yang.
\newblock {SIFU}: Side-view conditioned implicit function for real-world usable
  clothed human reconstruction.
\newblock In \emph{Proceedings of the IEEE/CVF Conference on Computer Vision
  and Pattern Recognition (CVPR)}, 2024.

\bibitem[Zheng et~al.(2024)Zheng, Zhou, Shao, Liu, Zhang, Nie, and
  Liu]{zheng2024gpsgaussian}
Shunyuan Zheng, Boyao Zhou, Ruizhi Shao, Boning Liu, Shengping Zhang, Liqiang
  Nie, and Yebin Liu.
\newblock {GPS-Gaussian}: Generalizable pixel-wise {3D} {Gaussian Splatting}
  for real-time human novel view synthesis.
\newblock In \emph{Proceedings of the IEEE/CVF Conference on Computer Vision
  and Pattern Recognition (CVPR)}, 2024.

\bibitem[Zheng et~al.(2022)Zheng, Yu, Liu, and Dai]{zheng2020pamir}
Zerong Zheng, Tao Yu, Yebin Liu, and Qionghai Dai.
\newblock {PaMIR}: Parametric model-conditioned implicit representation for
  image-based human reconstruction.
\newblock \emph{IEEE Transactions on Pattern Analysis and Machine
  Intelligence}, 2022.

\bibitem[Zhou et~al.(2026)Zhou, Li, Lyu, Song, Huang, Zheng, Rupprecht,
  Vedaldi, and Wu]{zhou2026articraft}
Matt Zhou, Ruining Li, Xiaoyang Lyu, Zhaomou Song, Zhening Huang, Chuanxia
  Zheng, Christian Rupprecht, Andrea Vedaldi, and Shangzhe Wu.
\newblock {Articraft}: An agentic system for scalable articulated {3D} asset
  generation.
\newblock \emph{arXiv preprint arXiv:2605.15187}, 2026.

\bibitem[Zhu et~al.(2024{\natexlab{a}})Zhu, Zhan, Theobalt, and
  Habermann]{heminggaberman2024trihuman}
Heming Zhu, Fangneng Zhan, Christian Theobalt, and Marc Habermann.
\newblock {TriHuman}: A real-time and controllable tri-plane representation for
  detailed human geometry and appearance synthesis.
\newblock \emph{ACM Transactions on Graphics}, 2024{\natexlab{a}}.

\bibitem[Zhu et~al.(2024{\natexlab{b}})Zhu, Chen, Dai, Dong, Xu, Cao, Yao, Zhu,
  and Zhu]{zhu2024champ}
Shenhao Zhu, Junming~Leo Chen, Zuozhuo Dai, Zilong Dong, Yinghui Xu, Xun Cao,
  Yao Yao, Hao Zhu, and Siyu Zhu.
\newblock Champ: Controllable and consistent human image animation with {3D}
  parametric guidance.
\newblock In \emph{European Conference on Computer Vision (ECCV)},
  2024{\natexlab{b}}.

\bibitem[Zhuang et~al.(2025)Zhuang, Lv, Wen, Shuai, Zeng, Zhu, Chen, Yang, Cao,
  and Liu]{zhuang2025idol}
Yiyu Zhuang, Jiaxi Lv, Hao Wen, Qing Shuai, Ailing Zeng, Hao Zhu, Shifeng Chen,
  Yujiu Yang, Xun Cao, and Wei Liu.
\newblock {IDOL}: Instant photorealistic {3D} human creation from a single
  image.
\newblock In \emph{Proceedings of the IEEE/CVF Conference on Computer Vision
  and Pattern Recognition (CVPR)}, 2025.

\bibitem[Zielonka et~al.(2025)Zielonka, Bagautdinov, Saito, Zollh{\"o}fer,
  Thies, and Romero]{zielonka2023drivable}
Wojciech Zielonka, Timur Bagautdinov, Shunsuke Saito, Michael Zollh{\"o}fer,
  Justus Thies, and Javier Romero.
\newblock Drivable {3D} {Gaussian} avatars.
\newblock In \emph{International Conference on 3D Vision (3DV)}, 2025.

\end{thebibliography}
